\documentclass[pdflatex,sn-mathphys-num]{sn-jnl}% Math and Physical Sciences Numbered Reference Style 
\usepackage{amsmath,amssymb,amsfonts}
\usepackage{amsthm}
\usepackage{mathrsfs}
\usepackage{graphicx}
\usepackage{subcaption}   % modern subfigures (do NOT use subfigure)
\usepackage{tikz}         % only if you really draw with TikZ

\usepackage{booktabs}
\usepackage{longtable}
\usepackage{multirow}

\usepackage{xcolor}
\usepackage[most]{tcolorbox}

\usepackage[ruled,vlined,linesnumbered]{algorithm2e}
\SetKwComment{Comment}{$\triangleright$\ }{}
\usepackage{listings}
\usepackage{bm}

\usepackage{calc}
\usepackage[title]{appendix}
\usepackage{textcomp}
\usepackage{manyfoot}
\usepackage{float} % avoid [H] unless required by the journal/template
\usepackage[colorinlistoftodos]{todonotes}
\theoremstyle{thmstyleone}%
\theoremstyle{thmstyletwo}%

\theoremstyle{thmstylethree}%

\begin{document}

\title[Article Title]{Resolving the bias–precision paradox with stochastic causal representation learning for personalized medicine}

%%=============================================================%%
%% GivenName	-> \fnm{Joergen W.}
%% Particle	-> \spfx{van der} -> surname prefix
%% FamilyName	-> \sur{Ploeg}
%% Suffix	-> \sfx{IV}
%% \author*[1,2]{\fnm{Joergen W.} \spfx{van der} \sur{Ploeg} 
%%  \sfx{IV}}\email{iauthor@gmail.com}
%%=============================================================%%

% \author{\sur{authors}}
% \author[1]{\fnm{more authors } \sur{.....}}\email{}
%\equalcont{These authors contributed equally to this work.}

\author[1, 2]{\fnm{Peisong} \sur{Zhang}} \email{peisongzhang@u.nus.edu}

\author[1, 3]{\fnm{Manqiang} \sur{Peng}} \email{peng762550@sina.com}

\author[1]{\fnm{Yuxuan} \sur{Wu}} \email{saku2696983188@gmail.com}

\author[1, 4]{\fnm{Pawit} \sur{Phadungsaksawasdi}} \email{pawit.p@tu.ac.th}

\author[5]{\fnm{Wesley} \sur{Yeung}} \email{wesleyyeung123@gmail.com}

\author[1, 6]{\fnm{Ye} \sur{Zhang}} \email{zhangye860525@sina.com}

\author[1, 7]{\fnm{Trang} \sur{Nguyen}} \email{trangn@stanford.edu}

\author[8]{\fnm{Qiang} \sur{Zhang}} \email{qiang.zhang.cs@zju.edu.cn}

\author[9]{\fnm{Nan} \sur{Liu}} \email{liu.nan@duke-nus.edu.sg}

\author[1]{\fnm{Meng} \sur{Wang}} \email{wangm.nus@gmail.com}

\author[1]{\fnm{Kee Yuan} \sur{Ngiam}} \email{surnky@nus.edu.sg}

\author[1]{\fnm{Yih-Chung} \sur{Tham}} \email{thamyc@nus.edu.sg}

\author[1]{\fnm{Ching-Yu} \sur{Cheng}} \email{chingyu.cheng@nus.edu.sg}

\author[10]{\fnm{Tianfan} \sur{Fu}} \email{futianfan@gmail.com}

\author[11]{\fnm{Qingyu} \sur{Chen}} \email{qingyu.chen@yale.edu}

\author[12]{\fnm{Rosemary} \sur{Ke}}\email{rosemary.nan.ke@gmail.com}

\author[13, 14]{\fnm{Chang} \sur{Li}}\email{lichang1@sysucc.org.cn}

\author[13, 14]{\fnm{Wenzhuo} \sur{Yang}}\email{yangwz7017@163.com}

\author[15]{\fnm{Zhenghao} \sur{Lu}}\email{luzhenghao2007@126.com}

\author[16]{\fnm{Chunyou} \sur{Lai}}\email{laichunyou2010@163.com}

\author[7]{\fnm{Yu} \sur{Zhang}}\email{yzhangsu@stanford.edu}

\author[13, 14]{\fnm{Sheng} \sur{Zhong}}\email{zhongsheng@sysucc.org.cn}

\author[17]{\fnm{Hao} \sur{Deng}}\email{
hdeng1@mgh.harvard.edu}

\author*[1, 2]{\fnm{Dianbo} \sur{Liu}}\email{dianbo@nus.edu.sg}

\affil*[1]{\orgdiv{School of Medicine}, \orgname{NUS}, }

\affil[2]{\orgdiv{College of Design and Engineering}, \orgname{NUS}, }

\affil[3]{\orgdiv{Aier Academy of Ophthalmology}, \orgname{Central South University}}

\affil[4]{\orgdiv{Division of Dermatology}, \orgname{Thammasat University}}

\affil[5]{\orgdiv{National University Heart Center}, \orgname{National University Health System}}

\affil[6]{\orgdiv{Beijing Tongren Eye Center}, \orgname{Beijing Tongren Hospital, Capital Medical University}}

\affil[7]{\orgdiv{School of Medicine}, \orgname{Stanford University}}

\affil[8]{\orgdiv{College of Computer Science}, \orgname{Zhejiang University}}

\affil[9]{\orgdiv{Duke-NUS Medical School}, \orgname{NUS}}

\affil[10]{\orgdiv{Department of Computer Science}, \orgname{Nanjing University}}

\affil[11]{\orgdiv{School of Medicine}, \orgname{Yale University}}

\affil[12]{\orgdiv{DeepMind}, \orgname{Google}}

\affil[13]{\orgdiv{Department of Neurosurgery}, \orgname{State Key Laboratory of Oncology in South China}}

\affil[14]{\orgdiv{Collaborative Innovation Center for Cancer Medicine}, \orgname{Sun Yat-sen University Cancer Center}}

\affil[15]{\orgname{Chengdu OrganoidMed Medical Laboratory}}

\affil[16]{\orgdiv{Department of Geriatric Comprehensive Surgery and International Medicine}, \orgname{Sichuan Provincial People's Hospital}}

\affil[16]{\orgdiv{Department of Geriatric Comprehensive Surgery and International Medicine}, \orgname{Sichuan Provincial People's Hospital}}

\affil[17]{\orgdiv{Harvard Medical School}, \orgname{Harvard University}}

\abstract{
Estimating individualized treatment effects from longitudinal observational data is central to data-driven medicine, yet existing methods face a fundamental limitation: reducing confounding bias often suppresses clinically informative heterogeneity, degrading patient-specific predictions. Here, we identify this tension as a bias–precision paradox in causal representation learning and introduce sampling-based maximum mean discrepancy (sMMD), a stochastic alignment strategy that replaces global adversarial balancing with subset-level matching. We instantiate this approach in a framework for counterfactual outcome prediction with attribution-grounded interpretability. Across two large-scale ICU cohorts (n = 27,783), our framework improves accuracy under distribution shift, reducing error by up to 11.5\% and substantially increasing recall in high-risk tasks. Mechanistic analyses show that sMMD selectively preserves clinically decisive variables. In human–AI evaluation, our method outperforms clinicians-in-training and large language models, and improves clinician accuracy by 14.7\% while reducing decision time, enabling interpretable, real-time clinical decision support.
}


\maketitle
\section{Introduction}
\label{sec:introduction}

\begin{figure*}[p]
    \centering
        \includegraphics[width=\linewidth]{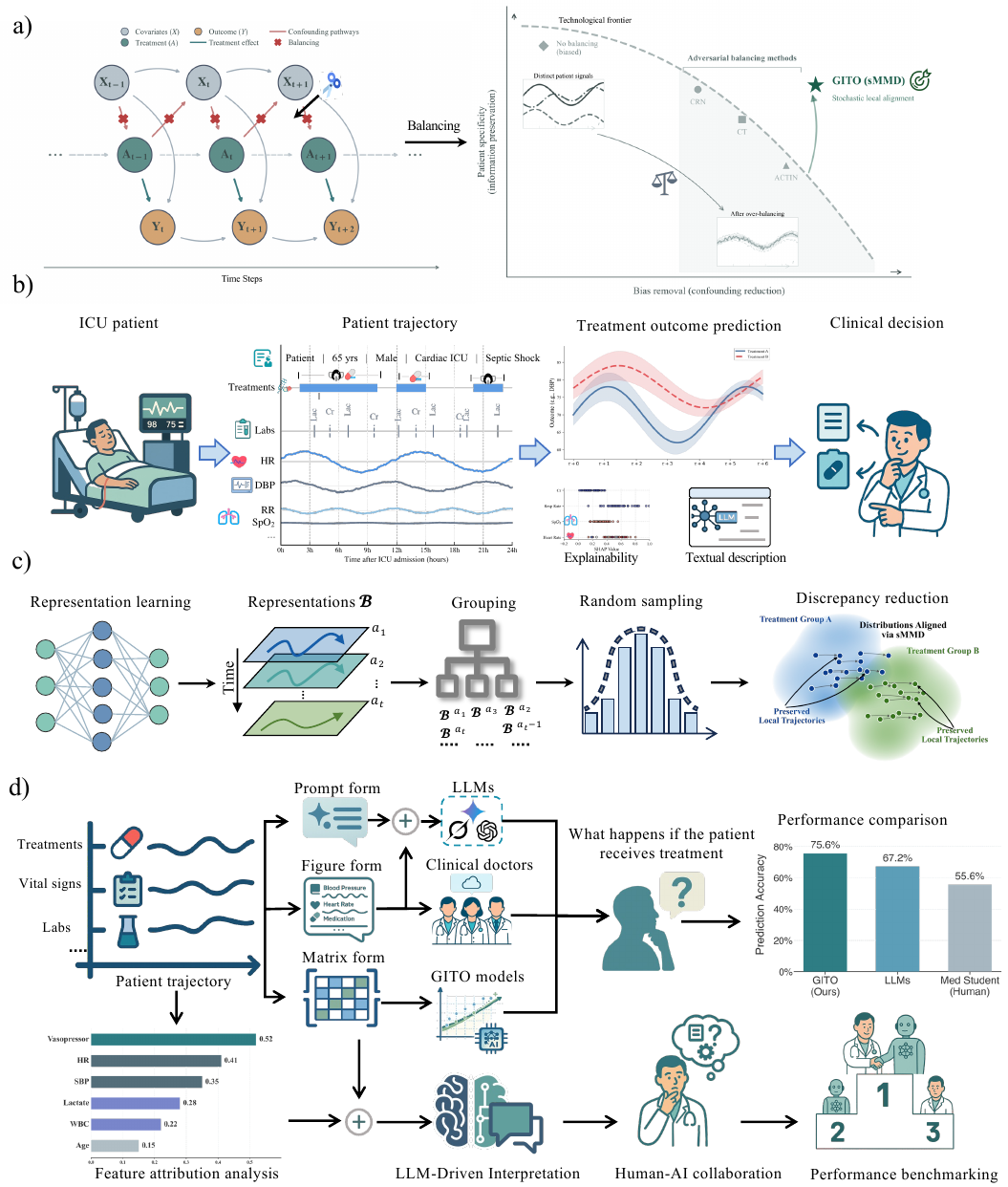}
    \caption{
 Resolving the precision-bias paradox in causal AI for critical care.
(a) AI integration within the ICU clinical workflow. The system aggregates multimodal patient data—including demographics, vital signs, treatments, and laboratory results—to provide decision support for treatment selection.
(b) The bias-precision dilemma and its causal origin. Left: the causal directed acyclic graph (DAG) illustrates how time-varying confounding arises in longitudinal treatment settings—co-variates ($X$) simultaneously influence treatment assignment ($A$) and outcomes ($Y$), creating confounding pathways (red arrows) that balancing methods aim to block (red crosses). Right: the bias-precision scatter plot shows that existing adversarial balancing methods are constrained by a technological frontier (dashed line), where aggressive bias removal ($x$-axis) sacrifices patient specificity ($y$-axis). In the over-balancing regime, generic representations fail to capture individual physiological dynamics. GITO (green star) transcends this frontier via sampling-based Maximum Mean Discrepancy (sMMD), simultaneously mitigating confounding while preserving information essential for individualized prediction.
(c) The sampling-based MMD (sMMD) balancing strategy. Representations $\bm{\mathcal{B}}$ are grouped according to treatment assignments. Random sampling is performed across groups to align sample distributions. The Maximum Mean Discrepancy (MMD) is then applied to minimize distributional differences between sampled treatment groups, achieving balanced representations without global homogenization.
(d) Human-AI comparison and collaboration. Patient trajectories are presented in modality-appropriate formats: text prompts for LLMs, vital sign charts for medical students, and structured matrices for GITO. GITO achieved 75.6\% prediction accuracy, outperforming LLMs (best: 67.2\%) and unassisted medical students (55.6\%). Bottom: feature attribution scores are combined 
with LLM-driven interpretation to support human-AI collaboration, where clinician performance improved when assisted by GITO's explanations (right).}
\label{fig:framework}
\end{figure*}

\begin{figure*}[htbp]
  \centering
  \begin{subfigure}[t]{\textwidth}
    \centering
    \includegraphics[width=\linewidth]{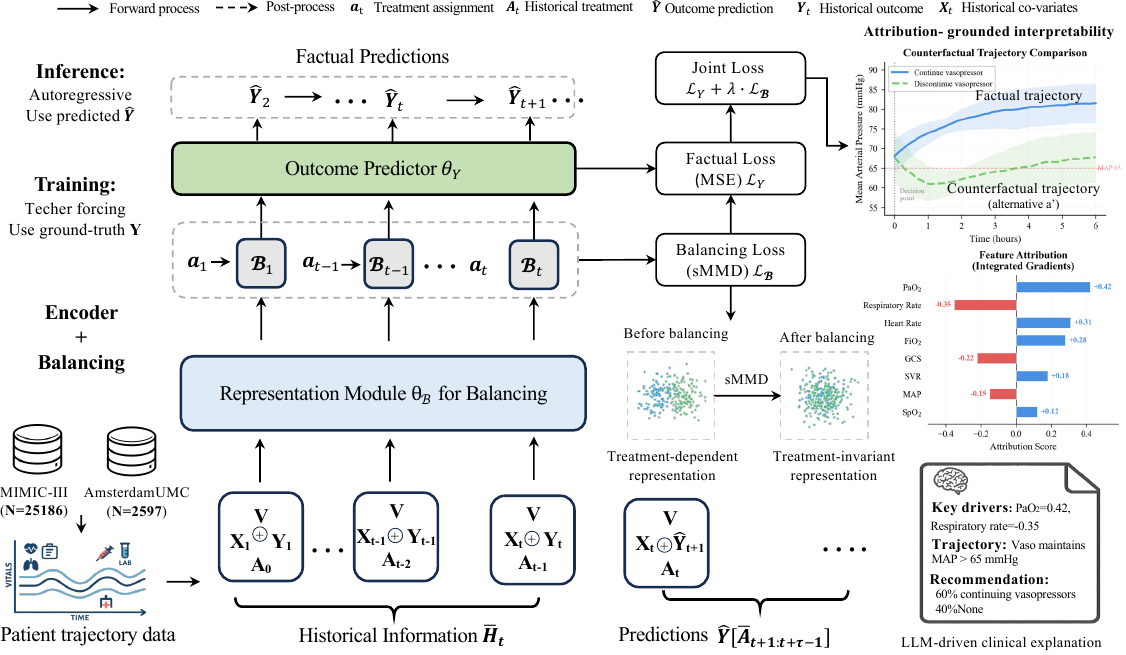}
  \end{subfigure}

  \begin{subfigure}[t]{\textwidth}
    \centering
    \includegraphics[width=\linewidth]{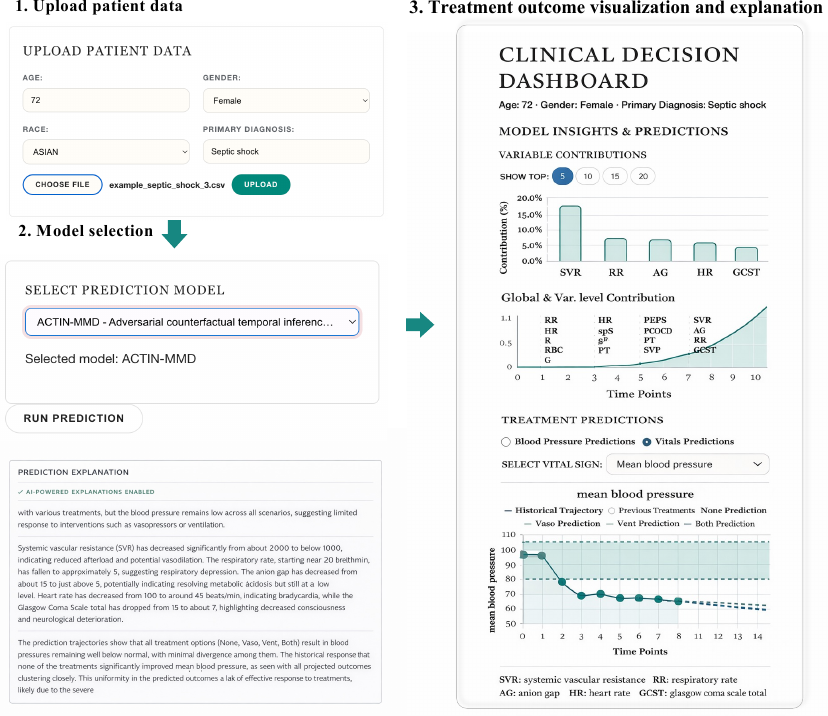}
  \end{subfigure}

  \caption{\textbf{The GITO framework and clinical 
  decision-support interface.}
  \textbf{a,}~Patient longitudinal data (vitals $\bm{V}$, co-variates $\bm{X}_t$, outcomes $\bm{Y}_t$, and treatments $\bm{A}_t$) are encoded into balanced representations $\bm{\mathcal{B}}_t$ by the representation module $\Theta_{\mathcal{B}}$. The outcome predictor $\Theta_Y$ generates multi-step counterfactual predictions $\hat{\bm{Y}}$ under alternative treatment scenarios. Sampling-based MMD (sMMD) aligns treatment group distributions while 
  preserving patient-level heterogeneity (right: distribution matching), in contrast to adversarial methods that enforce global invariance (left: scatter plots before vs.\ after balancing). An attribution module computes per-variable contributions, which are translated into natural-language clinical rationales by an LLM-based explanation module (upper right).
  \textbf{b,}~Web-based clinical interface. Clinicians upload patient data and select a prediction model (top). The dashboard displays variable-level attribution scores, temporal contribution patterns, and counterfactual treatment trajectories (right-bottom). An LLM-generated explanation provides a structured clinical rationale with treatment preference distribution (left-bottom). Data sources: MIMIC-III ($N{=}25{,}186$, United 
  States) and AmsterdamUMCdb ($N{=}2{,}597$, the Netherlands).
  }
  \label{fig:framework_and_user_interface}
\end{figure*}

% \textcolor{red}{Ching-Yu Cheng feedback: 1) clearly show generalization of model trained in one hospital performance on another is important 2) the storyline needs to be clear and focused, anything else need to be moved to appendix}

% ============================================================
% Structure: 6 paragraphs (General → Specific → General)
% ============================================================

Data-driven personalized medicine aims to tailor interventions to individual patient characteristics by estimating treatment effects from longitudinal observational data~\cite{journal/aim1997/757Rubin, journal/naturemed2024/30Feuerriegel, journal/epidemiology2000/Robins, journal/npjdm2020/17Sutton}.
A critical challenge, however, limits this endeavour: observational data are inherently confounded, because treatment decisions reflect patient severity rather than random allocation~\cite{journal/biometrika1983/41Rosenbaum, journal/jasa1984/516Rosenbaum}.
To address this, state-of-the-art frameworks enforce distributional alignment between treated and untreated populations in the learned representation space (Figure~\ref{fig:framework}a)~\cite{conference/nips2018/31Lim, conference/iclr2020/Bica, conference/icml2022/Melnychuk, conference/icml2024/wang}.
This deconfounding process introduces an under-examined trade-off: the patient features that drive treatment assignment, and thus differ most between groups, are often the most clinically informative.
Aggressive alignment can neutralize the clinically informative heterogeneity (severity indicators, disease subtypes, comorbidity profiles, and temporal trajectories) that is essential for individualized prediction~\cite{journal/ije2016/45Dahabreh, journal/biopsy2020/88Eric, journal/lancetDH/1Forte}.
We term this the bias-precision dilemma: global deconfounding improves average causal estimates at the expense of individual-level specificity.
This dilemma is not confined to a single clinical domain; it arises wherever observational data guide treatment, from oncology dose optimization to chronic disease management~\cite{journal/bmj2018/363Kent,journal/medrxiv2026/Soltanifar}.

Current representation-learning approaches for treatment effect estimation, including Counterfactual Recurrent Networks (CRN)~\cite{conference/iclr2020/Bica}, Causal Transformers (CT)~\cite{conference/icml2022/Melnychuk}, Adversarial Counterfactual Temporal Inference Network (ACTIN)~\cite{conference/icml2024/wang}, and their variants~\cite{conference/nips2018/31Lim, conference/icdm2022/1053Li, conference/kdd2024/12Wu, conference/nips2024/37Bouchattaoui}, adopt a shared encoder that maps patient co-variates into a latent space, from which treatment-specific outcome heads generate predictions. To remove confounding, these methods converge on a common paradigm: adversarial balancing. The encoder is trained to produce representations from which treatment assignment is unrecoverable, while an adversarial objective simultaneously attempts to identify it. This competition drives the learned representations toward global distributional invariance across treatment groups, effectively simulating the balance of a randomized trial. The consequence, however, is indiscriminate: the adversarial objective neutralizes any distributional difference between treatment groups, regardless of clinical relevance (Figure~\ref{fig:framework}a).
Consider vasopressor therapy in sepsis. Sicker patients are more likely to receive vasopressors, so blood pressure trajectories (the signals that determine vasopressor need) differ systematically between treated and untreated groups. Adversarial balancing suppresses precisely these trajectories, rendering the model unable to distinguish patients who require intervention from those who do not~\cite{conference/nips2022/Moayeri, conference/icml2024/Huang}. The result is ``over-balancing'': representations that are deconfounded on average but uninformative at the individual level, leading to poor generalization across clinically distinct subpopulations~\cite{journal/crt2024/4Curth, journal/cer2017/288Li}. A second barrier compounds the first: existing causal inference frameworks remain opaque. Standard explainability techniques yield feature-level importance scores but not the contextual, clinically grounded rationale that physicians need to trust and act on a recommendation~\cite{journal/naturemi/421Liu}. Together, over-balanced representations and opaque predictions constrain the real-world utility of current methods (Figure~\ref{fig:framework}a).

The bias-precision dilemma is particularly consequential in intensive care units (ICUs), where interventions are time-critical and the margin for error is narrow~\cite{journal/jbiei2016/3Roughead, journal/cricare2008/12Moyen, journal/bmj2019/366Panagioti}. Therapies such as vasopressor administration and mechanical ventilation demand continuous recalibration to a patient's evolving physiology~\cite{journal/ccm2024/1633Bauer, journal/aim2019/285Cox}, and treatment assignment is strongly confounded by illness severity. ICU environments also generate rich, high-resolution longitudinal data while carrying substantial adverse-event burden; unsafe care contributes to an estimated 3~million deaths annually worldwide~\cite{journal/bmj2016/353Makary, who2023patientsafety}. High stakes, strong confounding, and data availability make the ICU an ideal proving ground for resolving this dilemma. Data-driven models for personalized ICU interventions have shown the potential to reduce mortality by up to 20\%~\cite{conference/prmlhc2017/68Raghu, journal/eswa2021/169Akash, journal/aim2019/285Cox}, yet realizing this potential demands models that are both accurate and trustworthy.

We propose GITO (Generalized and Interpretable Treatment Outcome), a framework that replaces adversarial balancing with a fundamentally different de-confounding strategy (Figure~\ref{fig:framework}b,c). In longitudinal treatment settings, the encoder produces representations across many time steps, each associated with a treatment assignment. Rather than training a discriminator to enforce global invariance over all these representations simultaneously, GITO employs sampling-based maximum mean discrepancy (sMMD): at each training iteration, small random subsets are drawn from each treatment group and aligned via MMD. This stochastic, sample-level alignment provides a softer distributional constraint that mitigates confounding without forcing the entire representation space into a single homogenized distribution. As a result, the model retains the patient-level heterogeneity (severity indicators, disease subtypes, comorbidity profiles, and temporal trajectories) that adversarial methods inadvertently discard through their pursuit of global invariance.
Accuracy alone, however, does not earn clinical trust; physicians must understand why a model recommends a course of action before they act on it~\cite{journal/mlhc2019/106Tonekaboni}. Existing explainability methods such as SHAP values or attention-weight visualization yield numerical feature attributions but lack the clinical context that supports bedside reasoning~\cite{journal/lancetDH2021/3Ghassemi}. To bridge this gap, GITO incorporates an attribution-grounded interpretability pipeline that translates the model's per-feature contributions into natural-language clinical narratives via a Large Language Model, constrained to reason over model-derived evidence to mitigate hallucination risk (Figure~\ref{fig:framework}b).

We evaluate GITO on two real-world ICU databases, MIMIC-III (25,186~patients, United~States) and AmsterdamUMCdb (2,597~patients, the Netherlands), spanning populations with distinct demographic and ethnic compositions. GITO maintains robust performance when transferring from the training population (White, $N{=}3{,}560$) to held-out Asian ($N{=}119$), African ($N{=}383$), and Latino ($N{=}143$) descendant subgroups, achieving a 13\% average RMSE reduction relative to unbalanced baselines.
We validate the framework on two clinically distinct tasks: blood pressure trajectory prediction and ventilator re-intubation risk assessment. For re-intubation, GITO-augmented predictions reduced high-risk false negatives by 42\% (recall 0.506 to 0.719) with a concurrent AUC improvement from 0.711 to 0.756.
In a human-AI benchmark (Figure~\ref{fig:framework}d and \ref{fig:ai_human_comparison_cooperation}b), GITO achieved 75.6\% accuracy, outperforming all four tested LLMs by 8.4 to 19.0 percentage points and unassisted medical students ($n = 4$).
In a separate cooperation study with practicing clinicians ($n = 3$, Figure~\ref{fig:ai_human_comparison_cooperation}c), GITO's explanations improved accuracy by 14.7 percentage points, reduced decision-making time by 74\%, and raised the safety rate from 82.4\% to 89.8\%.
To our knowledge, GITO is the first framework to jointly resolve the bias-precision trade-off through sample-level distributional alignment and to close the interpretability gap with attribution-grounded LLM reasoning in longitudinal causal inference.
We release GITO as a freely accessible, open-source, web-based clinical tool (Figure~\ref{fig:framework_and_user_interface}) with sub-50\,ms inference latency on standard CPU hardware, enabling deployment within secure hospital intranets.
Both the sMMD alignment strategy and the interpretability pipeline are \emph{domain-agnostic}, applicable wherever individualized treatment effects must be estimated from observational data.

\begin{table*}[htbp]
\centering
\caption{\textbf{Demographic and clinical characteristics of the study cohort.} A random subset of 5{,}000 patients with ICU stays $\geq$\,30 hours was drawn from MIMIC-III ($N = 25{,}186$). Values are presented as median (IQR) or $n$ (\%).}
\label{tab:table1_all}
\resizebox{\textwidth}{!}{
\begin{tabular}{lcc}

\toprule
\textbf{Characteristic} &\textbf{All Patients (N=25186)} & \textbf{Subset Patients (N=5000)} \\
\midrule
\textbf{Age} &\\
- Age (for patients $\leq$ 89), mean (SD) & 62.90 (16.84) & 63.19 (16.83) \\
- Age $>$ 89, n (\%) &1347 (5.34\%) & 270 (5.4\%) \\
\textbf{Gender} & \\
- Male, n (\%) &14186 (56.3\%)& 2825 (56.5\%) \\
- Female, n (\%) &11000 (43.7\%) & 2175 (43.5\%) \\
\textbf{Ethnicity} & \\
- European descent, n (\%) &17919 (71.11\%) & 3560 (71.2\%)  \\
- African descent, n (\%) &1928 (7.66\%) & 383 (7.66\%) \\
- Latino descent, n (\%) &809 (3.21\%) & 143 (2.86\%) \\
- Asian descent, n (\%) &630 (2.50\%) & 119 (2.38\%) \\
- Native American, n (\%) &1799 (7.14\%) & 359 (7.18\%)  \\
- Other, n (\%) &2101 (8.34\%) &436 (8.72\%) \\
\textbf{Vitals}$^{\textit{a}}$ &\\
- Heart rate (bpm), mean (SD) &84.96 (15.11) & 85.07 (15.26) \\
- Red blood cells (M/$\mu$L), mean (SD) &3.65 (0.63)& 3.66 (0.62) \\
- Sodium (mEq/L), mean (SD) & 138.55 (4.22)& 138.52 (4.29) \\
- SVR (dyn$\cdot$/cm$^5$), mean (SD) &1502.48 (694.39)& 1496.52 (694.55) \\
- Glucose (mg/dL), mean (SD) &137.98 (38.63)& 138.39 (38.96) \\
- Chloride urine (mEq/L), mean (SD) &65.74 (47.67) & 67.43 (48.84) \\
- GCS score, mean (SD) & 13.54 (2.60) & 13.50 (2.65) \\
- Hematocrit (\%), mean (SD) &32.43 (5.09) & 32.48 (5.13) \\
% -Positive end-expiratory pressure(cmH$_2$O), mean (SD) & 5.16 (2.32) \\
- PEEP (cmH$_2$O), mean (SD) &5.17 (2.21) & 5.16 (2.32) \\
- Respiratory rate (bpm), mean (SD) & 18.56 (3.96) &18.48 (3.95) \\
- Prothrombin time (sec), mean (SD) & 15.13 (4.94)&15.02 (4.78) \\
- Cholesterol (mg/dL), mean (SD) &162.85 (47.35) & 162.68 (47.40) \\
- Hemoglobin (g/dL), mean (SD) &11.01 (1.82) & 11.03 (1.82) \\
- Creatinine (mg/dL), mean (SD) & &1.30 (1.31) \\
- Blood urea nitrogen (mg/dL), mean (SD)&24.68 (19.47)  & 24.23 (19.04) \\
- Bicarbonate (mEq/L), mean (SD) &23.84 (3.98) & 23.93 (4.01) \\
- Calcium ionized (mmol/L), mean (SD) &1.48 (7.57) & 1.54 (8.04) \\
- Partial pressure of CO$_2$ (mmHg), mean (SD) &40.85 (8.78) & 40.86 (8.63) \\
- Magnesium (mg/dL), mean (SD) &2.01 (0.32) & 2.01 (0.34) \\
- Anion gap (mEq/L), mean (SD) &13.87 (3.16) & 13.82 (3.18) \\
- Phosphorous (mg/dL), mean (SD) &3.49 (1.12) & 3.50 (1.12) \\
- Venous pvo2 (mmHg), mean (SD) &50.92 (13.69) & 50.68 (13.60) \\
- Platelets (K/$\mu$L), mean (SD) &220.07 (104.81) & 220.12 (104.52) \\
- Calcium urine (mg/dL), mean (SD) &5.11 (8.88) &5.42 (9.71) \\
- FiO$_{2}$ (\%), mean (SD) &51.96 (18.18) &52.15 (18.38) \\
- Diastolic blood pressure (mmHg), mean (SD) &60.50 (10.41)  & 60.55 (10.45) \\
- Mean blood pressure (mmHg), mean (SD) &78.20 (10.81)& 78.25 (10.81) \\
- Systolic blood pressure (mmHg), mean (SD) &78.20 (10.81)& 78.25 (10.81) \\
\textbf{Disease category}  \\
- Cardiovascular Disease, n (\%) & 5234 (20.78\%) &1044 (20.88\%)  \\
- Neurological disorders, n (\%) & 2760 (10.96\%) & 532 (10.64\%) \\
- Infectious and inflammatory  disease, n (\%) & 2682 (10.64\%) & 555 (11.10\%)\\
\textbf{Treatments}$^{\textit{b}}$ \\
- Vasopressor (h), mean (SD) &7.74 (15.02) & 7.39 (14.50)\\
- Ventilation (h), mean (SD)& 10.39 (17.49)& 10.28 (17.33)\\
\bottomrule
\end{tabular}
}
\vspace{2mm}
\parbox{\textwidth}{
\footnotesize
\textbf{Abbreviations:} SD, standard division; SVR, Systemic vascular resistance; GCS score, Glasgow coma scale total; PEEP, Positive end-expiratory pressure. FiO$_{2}$, Fraction inspired oxygen
(\textbf{\textit{a}}) For time-varying vital signs, mean values were computed over the first 24 hours following ICU admission.
(\textbf{\textit{b}}) For treatments, the average number of hours of continuous or intermittent interventions was computed across all patients.
}
\end{table*}

\section{Results}
\label{sec:results}

% \textcolor{red}{1.dataset 2. good performance and generalization 3. reduce risk of ventilator weaning.  4. interpretation of model behavior in septic shock 5. compare with human 6. can imporve human 7. interface 8. robust to confounding factor 9. balance accuracy and bias removal}

We developed GITO, a framework for individualized treatment outcome prediction in the ICU that integrates a sampling-based maximum mean discrepancy (sMMD) alignment strategy with attribution-grounded interpretability (Figure~\ref{fig:framework}). To facilitate clinical adoption, GITO is implemented as an open-source, web-based decision-support tool (Figure~\ref{fig:framework_and_user_interface}).
%briefly introduce each experiment we did in very simple words 

GITO was developed and validated on two large-scale ICU cohorts from the United States (MIMIC-III; $n = 25{,}186$) and the Netherlands (AmsterdamUMCdb; $n = 2{,}597$), comprising 27{,}783 individuals in total.
First, we evaluated generalization across geographic and demographic distribution shifts
(Table~\ref{tab:real_world_results}, Table~\ref{tab:ethnicity_results}, Figure~\ref{fig:experiment_results_diseases}, Figure~\ref{fig:delta_r2}). 
Next, we assessed downstream clinical utility through ventilator weaning prediction
(Figure~\ref{fig:gito_ventilator_weaning_evaluation}).
Then, we evaluated model interpretability via attribution analysis and LLM-based explanations in a septic shock case study
(Figure~\ref{fig:case_study}).
Moreover, we benchmarked GITO against ICU clinicians (Figure~\ref{fig:ai_human_comparison_cooperation}).
In addition, we conducted human-AI collaboration experiments to quantify the effect of explanation-enhanced outputs on clinician performance (Figure~\ref{fig:ai_human_comparison_cooperation}).
We further validated robustness to confounding on a fully synthetic dataset
(Table~\ref{tab:tg_one_step_prediction}, Figure~\ref{fig:tg_multi_step_prediction}, Figure~\ref{fig:recon_loss}).
Finally, we quantified bias-accuracy trade-offs across representation-balancing strategies
(Figure~\ref{fig:representation_visualization}).

\subsection{Patient cohort}
\label{subsec:patient_cohort}

We analyze the effectiveness and robustness of GITO on three patient cohorts that span varying levels of complexity and real-world variability: AmsterdamUMCdb, MIMIC-III ICU cohort, and a fully synthetic tumor growth dataset.

\textbf{MIMIC-III electronic medical record data.} 
MIMIC-III database~\cite{journal/scidata2016/Johnson} is a large and widely used ICU patient cohort comprising detailed electronic health records. In this study, we included patients whose ICU stays lasted between 30 and 60 hours to ensure sufficient temporal coverage for treatment-outcome modeling. A total of 25,186 patients met these criteria, comprising 56.3\% males and 43.7\% females, with a mean age of 62.9 years. The cohort included patients from 41 ethnic groups, with an average stay in the ICU of 44.93 hours.
Among the included patients, vasopressor therapy was administered for an average of 7.74 $\pm$ 15.02 hours and mechanical ventilation for 10.39 $\pm$ 17.49 hours. The baseline demographic and treatment characteristics are summarized in \ref{tab:table1_all}.

\textbf{AmsterdamUMCdb electronic medical record data.} 
The AmsterdamUMCdb database~\cite{journal/ccm2021/49Thoral} is a large, openly accessible intensive care dataset containing detailed electronic health records from two university medical centers in the Netherlands. In this study, we included adult patients whose ICU stays lasted between 30 and 60 hours to ensure sufficient temporal coverage for treatment-outcome modeling. A total of 2,597 patients met these criteria, comprising 1,614 (62.2\%) males and 983 (37.8\%) females, with the largest age group being 70-79 years (25.6\%).
The mean ICU stay was 42.9 $\pm$ 7.1 hours. Among the included patients, vasopressor therapy was administered for an average of 12.91 $\pm$ 15.76 hours, and mechanical ventilation for 12.34 $\pm$ 15.40 hours. Baseline demographic and treatment characteristics are summarized in Appendix~\ref{tab:patient_cohort_amsterdam}.

\textbf{Synthetic patient cohort for controlled confounding evaluation.} 
To enable controlled evaluation of counterfactual prediction, we simulated a synthetic patient cohort ($n = 10{,}000$) using a pharmacokinetic-pharmacodynamic (PKPD) tumor growth model~\cite{journal/scirep2017/7Geng}. This model simulates individualized treatment responses with known ground-truth counterfactual outcomes, allowing precise quantification of prediction accuracy under varying degrees of treatment selection bias~\cite{conference/iclr2020/Bica, conference/icml2022/Melnychuk, conference/icml2024/wang}. The synthetic cohort includes patients with diverse baseline tumor characteristics (volume and growth rate) and treatment scenarios spanning 30-day observation periods. Confounding strength ($\gamma$) was systematically varied from 0 (randomized treatment) to 7.0 (strong selection bias) to evaluate model robustness across realistic clinical scenarios where treatment assignment depends on patient severity and prognosis. While tumor growth differs from acute ICU conditions, the underlying mathematical framework for treatment effect estimation and confounding control directly translates to ICU trajectory prediction tasks.
Full simulation parameters, including treatment assignment mechanisms and validation protocols, are detailed in Appendix~\ref{subsec:details_synthetic_dataset}.

\subsection{GITO enables robust generalization across geographic and demographic shifts}

\newcommand{\sigstar}{\textcolor{red}{\textsuperscript{*}}}
\begin{table*}[t]
\centering
\caption{Multi-step-ahead prediction results on MIMIC-III and AmsterdamUMCdb cohorts. 
The IID setting refers to standard train-test splits within the same population distribution.
In contrast, the OOD setting is designed to assess the generalization ability of models to previously unseen patient subgroups. Specifically, for the MIMIC-III cohorts, the OOD evaluation is conducted on non-European decent patients, while models are trained only on European-decent patients. For AmsterdamUMCdb, OOD evaluation corresponds to a cross-dataset generalization setting, where models are trained exclusively on MIMIC-III and evaluated directly on AmsterdamUMCdb without any fine-tuning. This setting reflects a more realistic deployment scenario, where a model trained in one hospital system is applied to a different clinical environment with distinct patient characteristics and data distributions. \sigstar denotes statistically significant improvement ($p \leq 0.05$).}
\label{tab:real_world_results}
\resizebox{\textwidth}{!}{
\begin{tabular}{@{}llcccccc@{}}
\toprule
 & & $\tau = 1$ & $\tau = 2$ & $\tau = 3$ & $\tau = 4$ & $\tau = 5$ & $\tau = 6$ \\
\midrule
\multicolumn{8}{c}{\textbf{MIMIC-III from Boston, U.S.A}} \\
\midrule
\multirow{6}{*}{IID}
 & CRN         & 4.84$\pm$0.08  & 9.13$\pm$0.16  & 9.77$\pm$0.16  & 10.11$\pm$0.17 & 10.36$\pm$0.20 & 10.58$\pm$0.22 \\
 & CRN-sMMD    & 4.70$\pm$0.07  & 9.16$\pm$0.18  & 9.81$\pm$0.19  & 10.15$\pm$0.17 & 10.41$\pm$0.21 & 10.64$\pm$0.24 \\
 & CT          & 4.60$\pm$0.08  & 8.99$\pm$0.21  & 9.59$\pm$0.22  & 9.91$\pm$0.25  & 10.14$\pm$0.29 & 10.34$\pm$0.32 \\
 & CT-sMMD     & 4.58$\pm$0.07  & 8.97$\pm$0.18  & 9.56$\pm$0.18  & 9.87$\pm$0.20  & 10.10$\pm$0.23 & 10.28$\pm$0.26 \\
 & ACTIN       & 4.57$\pm$0.07  & 4.87$\pm$0.06  & 4.98$\pm$0.07  & 5.08$\pm$0.10  & 5.15$\pm$0.13  & 5.22$\pm$0.17 \\
 & ACTIN-sMMD  & 4.57$\pm$0.08  & 4.87$\pm$0.07  & 4.99$\pm$0.07  & 5.08$\pm$0.08  & 5.15$\pm$0.10  & 5.21$\pm$0.12 \\
\midrule
\multirow{6}{*}{OOD}
 & CRN         & 5.66$\pm$0.15  & 9.79$\pm$0.18  &10.84$\pm$0.30  & 11.63$\pm$0.41  & 12.42$\pm$0.48  & 13.20$\pm$0.52 \\
 & CRN-sMMD    & 5.69$\pm$0.13  & 9.72$\pm$0.14  &10.76$\pm$0.21  & 11.43$\pm$0.27  & 12.16$\pm$0.34  & 12.89$\pm$0.40 \\
 & CT          & 5.64$\pm$0.14  & 9.65$\pm$0.11  &10.34$\pm$0.16  & 10.72$\pm$0.20  & 11.05$\pm$0.22  & 11.35$\pm$0.24 \\
 & CT-sMMD     & 5.64$\pm$0.15  & 9.65$\pm$0.13  &10.34$\pm$0.18  & 10.73$\pm$0.22  & 11.05$\pm$0.26  & 11.36$\pm$0.29 \\
 & ACTIN       & 4.80$\pm$0.21  & 5.22$\pm$0.30  & 5.47$\pm$0.38  & 5.66$\pm$0.46  & 5.84$\pm$0.54  & 6.02$\pm$0.61 \\
 & ACTIN-sMMD  & 4.63$\pm$0.07\sigstar & 4.98$\pm$0.08\sigstar & 5.16$\pm$0.09\sigstar & 5.28$\pm$0.12\sigstar & 5.39$\pm$0.14\sigstar & 5.49$\pm$0.16\sigstar \\
\midrule
\multicolumn{8}{c}{\textbf{AmsterdamUMCdb from Amsterdam, the Netherlands}} \\
\midrule
\multirow{6}{*}{IID}
 & CRN         & 17.90$\pm$1.35 & 8.92$\pm$0.48 & 9.98$\pm$0.69 & 10.59$\pm$0.62 & 11.21$\pm$0.59 & 11.97$\pm$0.59 \\
 & CRN-sMMD    & 18.09$\pm$1.56 & 8.79$\pm$0.52\sigstar & 9.62$\pm$0.40\sigstar & 10.20$\pm$0.38\sigstar & 10.84$\pm$0.38\sigstar & 11.71$\pm$0.49\sigstar \\
 & CT          & 18.31$\pm$1.61 & 9.14$\pm$0.92 & 10.08$\pm$0.98 & 10.75$\pm$1.03 & 11.41$\pm$1.01 & 12.26$\pm$0.98 \\
 & CT-sMMD     & 18.26$\pm$1.60 & 9.02$\pm$0.66 & 9.95$\pm$0.66 & 10.62$\pm$0.66 & 11.31$\pm$0.62 & 12.18$\pm$0.63 \\
 & ACTIN       & 18.85$\pm$4.21 & 8.95$\pm$0.60 & 9.89$\pm$0.55 & 10.61$\pm$0.66 & 11.28$\pm$0.66 & 12.18$\pm$0.83 \\
 & ACTIN-sMMD  & 17.93$\pm$1.74 & 8.89$\pm$0.58 & 9.89$\pm$0.51 & 10.43$\pm$0.56 & 11.11$\pm$0.58 & 11.98$\pm$0.65 \\
\midrule
\multirow{6}{*}{OOD}
 & CRN         & 20.32$\pm$0.42 & 17.45$\pm$4.90 & 21.41$\pm$6.84 & 25.36$\pm$7.82 & 27.39$\pm$8.13 & 28.74$\pm$7.59 \\
 & CRN-sMMD    & 20.03$\pm$0.16 & 15.97$\pm$2.14\sigstar & 19.88$\pm$3.19\sigstar & 22.80$\pm$2.99\sigstar & 24.78$\pm$3.15\sigstar & 26.32$\pm$3.20\sigstar \\
 & CT          & 18.83$\pm$0.56 & 13.32$\pm$0.83 & 14.94$\pm$1.53 & 15.99$\pm$1.66 & 17.17$\pm$1.95 & 18.10$\pm$2.08 \\
 & CT-sMMD     & 19.02$\pm$0.86 & 12.79$\pm$0.63\sigstar & 14.38$\pm$1.56\sigstar & 15.18$\pm$1.37\sigstar & 16.30$\pm$1.46\sigstar & 17.16$\pm$1.37\sigstar \\
 & ACTIN       & 18.82$\pm$0.12 & 10.28$\pm$0.30 & 11.27$\pm$0.42 & 12.05$\pm$0.41 & 12.77$\pm$0.36 & 13.65$\pm$0.30 \\
 & ACTIN-sMMD  & 18.70$\pm$0.25 & 9.96$\pm$0.69 & 10.86$\pm$0.76\sigstar & 11.47$\pm$0.81\sigstar & 12.10$\pm$0.86\sigstar & 12.87$\pm$0.88\sigstar \\
\bottomrule
\end{tabular}
}
\normalsize
\end{table*}

\begin{table*}[htbp]
\centering
\caption{Multi-step-ahead prediction results on the MIMIC-III dataset in out of distribution settings with 3 ethnicity. In this case, White patient were set as training set, and Asian, African and  Latino descendant patients were set as test set, respectively. Shown: RMSE as mean $\pm$ standard deviation over ten runs.}
\label{tab:ethnicity_results}
\begin{minipage}{\textwidth}
\resizebox{\textwidth}{!}{
\begin{tabular}{@{}llcccccc@{}}
\toprule
      &       & $\tau = 1$ & $\tau = 2$ & $\tau = 3$ & $\tau = 4$ & $\tau = 5$ & $\tau = 6$ \\ \midrule
\multirow{7}{*}{ASIAN} 
    & CRN       & 5.21$\pm$0.34  & 9.24$\pm$0.35  & 10.08$\pm$0.50  &10.68$\pm$0.71 & 11.22$\pm$0.96 & 11.80$\pm$1.23 \\
    & CRN-sMMD   & 5.21$\pm$0.34  & 9.23$\pm$0.27  & 10.06$\pm$0.35  &10.66$\pm$0.51 & 11.19$\pm$0.70 & 11.72$\pm$0.92 \\
      & CT        & 4.98$\pm$0.44  & 9.14$\pm$0.46  &9.83$\pm$0.50  & 10.25$\pm$0.53  & 10.58$\pm$0.57 &10.89$\pm$0.60 \\
      & CT-sMMD    & 4.95$\pm$0.43  & 9.03$\pm$0.33  &9.72$\pm$0.37  & 10.14$\pm$0.42  & 10.46$\pm$0.46 &10.78$\pm$0.50 \\
      & ACTIN     & 4.91$\pm$0.33  & 5.38$\pm$0.40  & 5.66$\pm$0.46  & 5.88$\pm$0.52  & 6.10$\pm$0.61  &6.30$\pm$0.68 \\
      & ACTIN-sMMD & 4.67$\pm$0.27\sigstar\sigstar  & 5.05$\pm$0.29\sigstar\sigstar  & 5.23$\pm$0.30\sigstar\sigstar  & 5.36$\pm$0.31\sigstar\sigstar  & 5.50$\pm$0.31\sigstar\sigstar  &5.61$\pm$0.30\sigstar\sigstar \\ \midrule
\multirow{7}{*}{African} 
      & CRN       & 5.73$\pm$0.52  & 10.27$\pm$0.41  & 11.15$\pm$0.61  &11.78$\pm$0.84 & 12.37$\pm$1.13 & 12.94$\pm$1.40 \\
      & CRN-sMMD   & 5.72$\pm$0.51  & 10.25$\pm$0.40  & 11.09$\pm$0.52  &11.62$\pm$0.63 & 12.08$\pm$0.78 & 12.54$\pm$0.94 \\
      & CT        & 5.51$\pm$0.72  & 10.07$\pm$0.48  & 10.75$\pm$0.47  &11.11$\pm$0.47 & 11.38$\pm$0.49 & 11.64$\pm$0.52 \\
      & CT-sMMD    & 5.52$\pm$0.71  & 10.08$\pm$0.47  & 10.76$\pm$0.46  &11.13$\pm$0.46 & 11.41$\pm$0.47 & 11.66$\pm$0.50 \\
      & ACTIN     & 5.20$\pm$0.24  & 5.67$\pm$0.30  & 5.95$\pm$0.34  & 6.17$\pm$0.41  & 6.37$\pm$0.49  &6.55$\pm$0.57 \\
      & ACTIN-sMMD & 5.03$\pm$0.26  &5.42$\pm$0.29\sigstar &5.60$\pm$0.31\sigstar &5.76$\pm$0.36\sigstar  &5.89$\pm$0.41\sigstar  &6.00$\pm$0.42\sigstar \\ \midrule
\multirow{7}{*}{Latino} 
    & CRN       & 5.32$\pm$0.22  & 9.51$\pm$0.31  & 10.41$\pm$0.46  &11.03$\pm$0.68 & 11.65$\pm$0.94 & 12.24$\pm$1.18 \\
    & CRN-sMMD   & 5.29$\pm$0.23  & 9.49$\pm$0.27  & 10.31$\pm$0.35  &10.86$\pm$0.51 & 11.41$\pm$0.73 & 11.93$\pm$0.89 \\
    & CT       & 5.06$\pm$0.32  & 9.38$\pm$0.33  & 10.07$\pm$0.33  &10.44$\pm$0.33 & 10.78$\pm$0.38 & 11.09$\pm$0.43 \\
    & CT-sMMD    & 5.07$\pm$0.31  & 9.38$\pm$0.33  & 10.08$\pm$0.35  &10.46$\pm$0.37 & 10.81$\pm$0.42 & 11.12$\pm$0.48 \\
    & ACTIN     & 4.91$\pm$0.31  & 5.49$\pm$0.32  & 5.81$\pm$0.39  &6.04$\pm$0.45 & 6.28$\pm$0.50 & 6.49$\pm$0.58 \\
    & ACTIN-sMMD & 4.70$\pm$0.21\sigstar  & 5.14$\pm$0.21\sigstar\sigstar  & 5.34$\pm$0.25\sigstar\sigstar  &5.49$\pm$0.26\sigstar\sigstar &5.66$\pm$0.28\sigstar\sigstar & 5.79$\pm$0.32\sigstar\sigstar \\ 
\bottomrule
\end{tabular}
}
\vspace{4mm}
\includegraphics[width=1\textwidth]{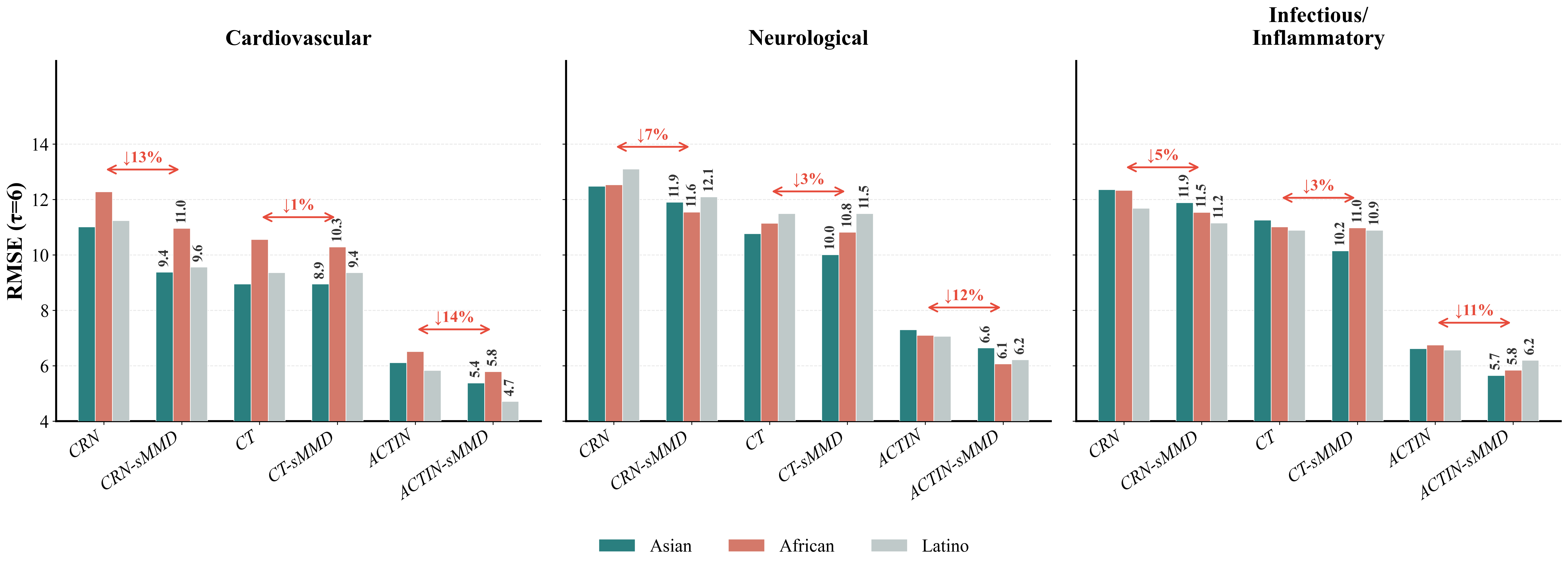} 
% \captionof{figure}{Multi-step-ahead prediction ($\tau=6$) results on MIMIC-III dataset in patients with 3 different diseases. In this case, White patient were set as training set, and the patients of Asian, African and Latino descent with 3 different diseases were set as test set, respectively. Shown: RMSE as mean ± standard deviation over ten runs.}
\captionof{figure}{\textbf{Disease-stratified out-of-distribution prediction performance 
($\boldsymbol{\tau = 6}$).}
Multi-step-ahead RMSE on MIMIC-III patients stratified by ethnicity and disease 
category. Models were trained exclusively on patients of European descent 
(N\,=\,3{,}560) and evaluated on held-out Asian (N\,=\,119), Black (N\,=\,383), 
and Hispanic (N\,=\,143) patients across three disease groups: cardiovascular 
(N\,=\,75), neurological (N\,=\,87), and infectious/inflammatory (N\,=\,85). 
Values are RMSE (mean $\pm$ s.d.; $n = 10$ independent runs). Statistical 
significance was assessed by two-sided paired \textit{t}-tests comparing each 
sMMD-enhanced model against its adversarial balancing counterpart; $^{*}p < 0.05$.}
\label{fig:experiment_results_diseases}
\end{minipage}
\end{table*}

\begin{figure}[t]
\centering
\includegraphics[width=\textwidth]{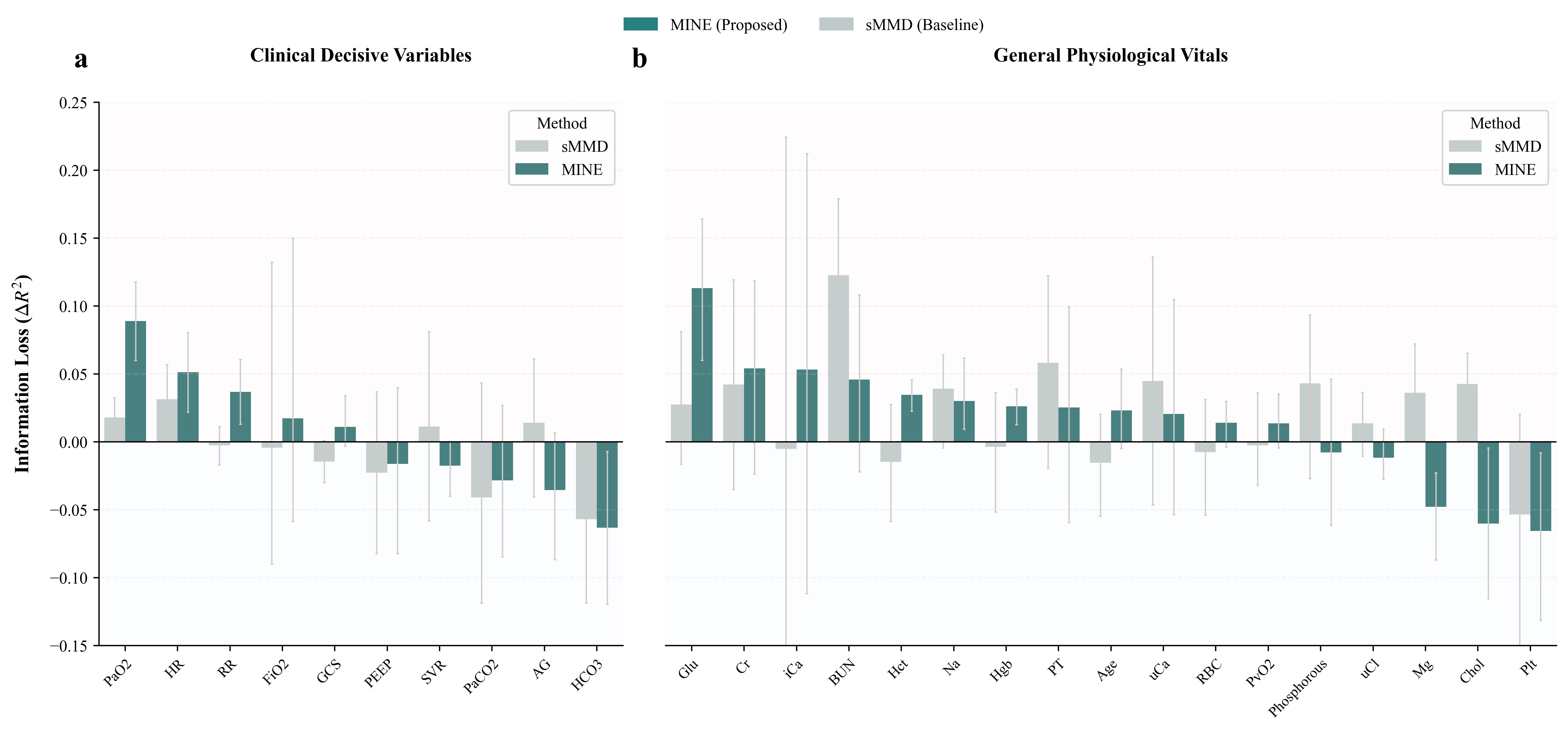}
\caption{Per-variable information loss ($\Delta R^{2}$) on the MIMIC-III cohort comparing sMMD and MINE (adversarial) balancing.
\textbf{(a)}~Clinically decisive variables directly involved in treatment decisions in ICU.
\textbf{(b)}~General physiological vitals and laboratory values.
Positive $\Delta R^{2}$ indicates information lost relative to an unbalanced encoder; values near zero indicate full preservation. Error bars denote 95\% confidence intervals over ten independent runs.}
\label{fig:delta_r2}
\end{figure}

% To understand if performance of predictive model constructed using our proposed sMMD can generalize across different hospitals and populations, we evaluated model performance using large-scale ICU cohorts from the United States (MIMIC-III) and the Netherlands (AmsterdamUMCdb). specifically, we examined two distinct types of distribution shift: geographic (cross-hospital) and demographic (cross-ethnicity). Our results indicate that the sMMD strategy effectively preserves patient-specific heterogeneity, achieving a reduction in Root Mean Squared Error (RMSE) of up to 19.2\% compared to state-of-the-art baselines in complex transfer scenarios.
To evaluate whether the sMMD balancing strategy enables predictive models to generalize across hospitals and patient populations, we assessed performance on large-scale ICU cohorts from the United States (MIMIC-III; $n = 25{,}186$) and the Netherlands (AmsterdamUMCdb; $n = 2{,}597$). We examined generalization along three axes, geographic (cross-hospital), demographic (cross-ethnicity), and disease-specific, and then investigated the mechanistic basis of the observed differences through per-variable information-preservation analysis.

\textbf{Institutional generalization (geographic shift).}
In the IID setting, GITO achieved parity with baseline models (CRN, CT, and ACTIN) on local test sets (Table~\ref{tab:real_world_results}). Performance divergence emerged under geographic shift: when models trained on the U.S.-based MIMIC-III cohort were deployed on the European AmsterdamUMCdb cohort without fine-tuning, baseline models showed increased error rates, whereas GITO maintained lower RMSE. Relative to each corresponding unbalanced baseline, sMMD-enhanced models reduced RMSE by 2.7-11.5\% across prediction horizons $\tau = 1$-$6$.

\textbf{Demographic generalization (subpopulation shift).}
We next examined whether the model generalizes across patient ethnicities within the MIMIC-III cohort. When models trained on patients of European descent were evaluated on Asian, African, and Latino populations, GITO consistently yielded lower error rates (Table~\ref{tab:ethnicity_results}). The model reduced RMSE by 3.5\% for single-step predictions and by up to 8.8\% for multi-step forecasts compared to baselines.
These gains were most pronounced in populations with higher baseline prediction errors: in patients of African descent, GITO reduced RMSE by 10.95\%, and for patients of Asian descent, by 8.8\% at $\tau = 6$.

\textbf{Disease-specific subgroup analysis.}
We stratified performance by disease type to assess reliability across clinical contexts (Figure~\ref{fig:experiment_results_diseases}). GITO yielded the lowest RMSE across cardiovascular, neurological, and infectious conditions. For cardiovascular disease, RMSE reductions reached 11.9\%, 11.06\%, and 19.21\% for patients of Asian, African, and Latino descent, respectively (paired \textit{t}-test, all $p < 0.05$; $n = 10$ splits). These disease-stratified results support the clinical applicability of GITO across heterogeneous ICU populations.

\textbf{Per-variable information preservation.}
To characterise how each balancing strategy affected individual clinical variables, we quantified per-variable information change. Any representation learning inherently involves information compression due to finite model capacity and dimensionality reduction; some degree of reconstruction error is therefore unavoidable regardless of the balancing strategy applied. To isolate the \textit{additional} effect of distribution alignment on each clinical variable, we computed per-variable $\Delta R^{2} = R^{2}_{\text{unbalanced}} - R^{2}_{\text{balanced}}$ on the MIMIC-III cohort (Figure~\ref{fig:delta_r2}). A positive $\Delta R^{2}$ indicates that balancing incurred additional information loss beyond baseline compression; values near zero indicate no added cost; negative values indicate that balancing improved reconstruction relative to the unbalanced encoder.
Both sMMD and the adversarial MINE objective showed positive $\Delta R^{2}$ on strongly treatment-correlated variables such as PaO$_2$ and heart rate (\ref{fig:delta_r2}~(a)). In contrast, variables with minimal direct influence on acute treatment selection,such as magnesium, cholesterol, and pH (\ref{fig:delta_r2}~(b)), remained near zero for both methods, suggesting that balancing predominantly affected treatment-correlated dimensions without disrupting less directly treatment-correlated physiological signals.
Despite this shared pattern, the two strategies differed substantially in both magnitude and consistency. Across all seven clinically decisive variables in \ref{fig:delta_r2}~(a), PaO$_2$, heart rate, respiratory rate, FiO$_2$, GCS, PEEP, and PaCO$_2$, sMMD achieved lower $\Delta R^{2}$ than MINE, indicating uniformly better information preservation. The gap was most pronounced for PaO$_2$ ($\Delta R^{2}$: 0.09 for MINE vs.\ 0.04 for sMMD), heart rate (0.09 vs.\ 0.06), and glucose (0.11 vs.\ 0.01 in \ref{fig:delta_r2}~(b)).
Beyond magnitude, the two methods diverged in direction on several clinically critical variables. For respiratory rate, FiO$_2$, and GCS, MINE showed positive $\Delta R^{2}$ (information lost relative to the unbalanced baseline), whereas sMMD achieved negative $\Delta R^{2}$ (information preserved better than the unbalanced encoder). Conversely, MINE produced a large negative $\Delta R^{2}$ for HCO$_3$ ($\approx -0.12$), yet incurred substantial positive $\Delta R^{2}$ on PaO$_2$ and glucose. Taken together, MINE exhibited high variance across variables, with both the largest gains and the largest losses, whereas sMMD produced a consistently near-zero $\Delta R^{2}$ profile across the full variable set, with selective improvements on variables directly involved in ventilator and consciousness assessment.

\subsection{Case study:improving ventilator intubation prediction by GITO}

\begin{figure}[H]
  \centering

  % ===================== (a) Forest =====================
  \begin{subfigure}[t]{\linewidth}
    \centering
    \includegraphics[width=\linewidth]{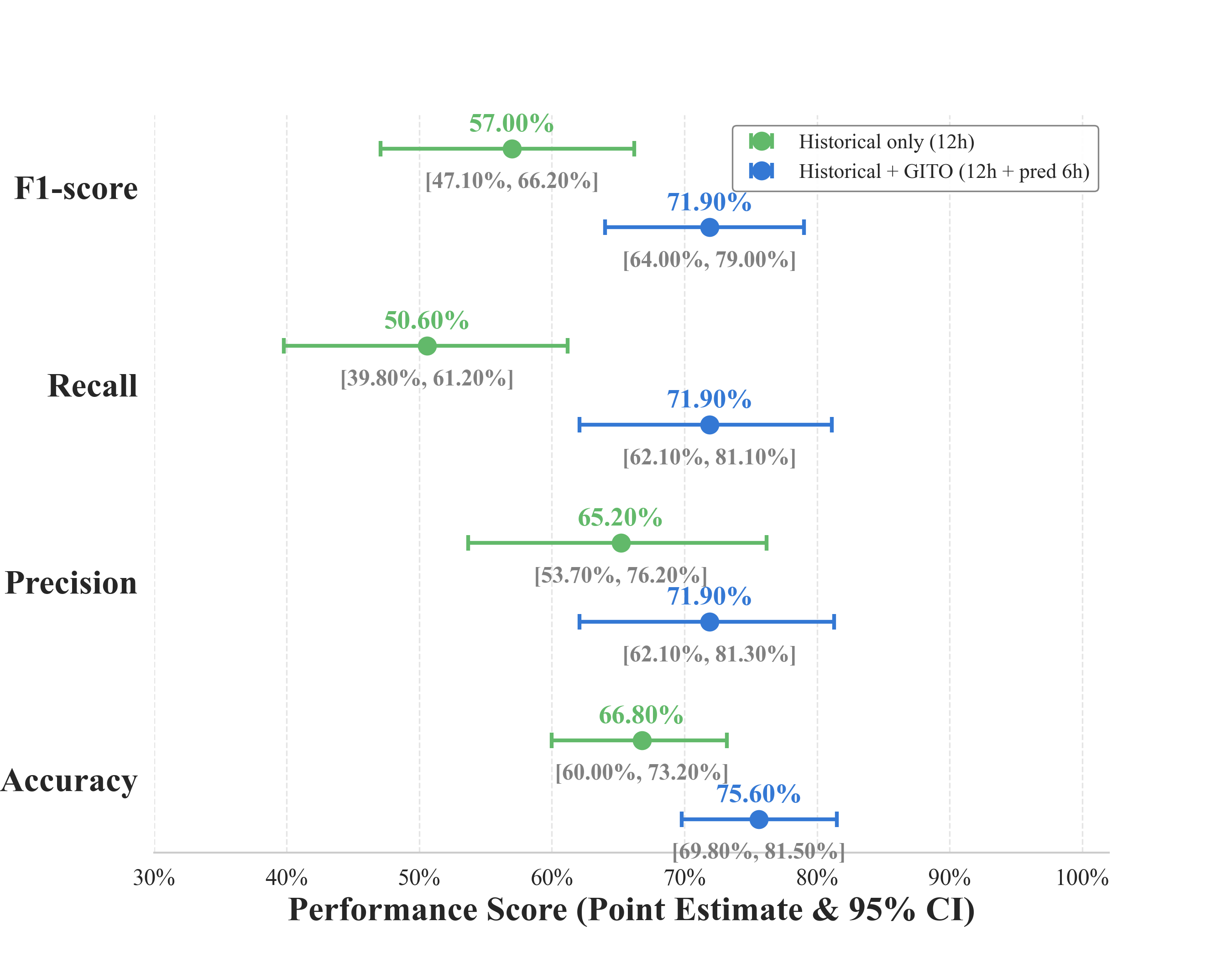}
    \caption{Forest plot comparing accuracy, precision, recall, and F1-score between the historical baseline (12-hour vitals) and the GITO-augmented model (6-hour predicted trajectories); point estimates with 95\% confidence intervals.}
    \label{fig:gito-forest}
  \end{subfigure}

  % ===================== (b) ROC =====================
  \begin{subfigure}[t]{0.35\linewidth}
    \centering 
    \includegraphics[width=\linewidth]{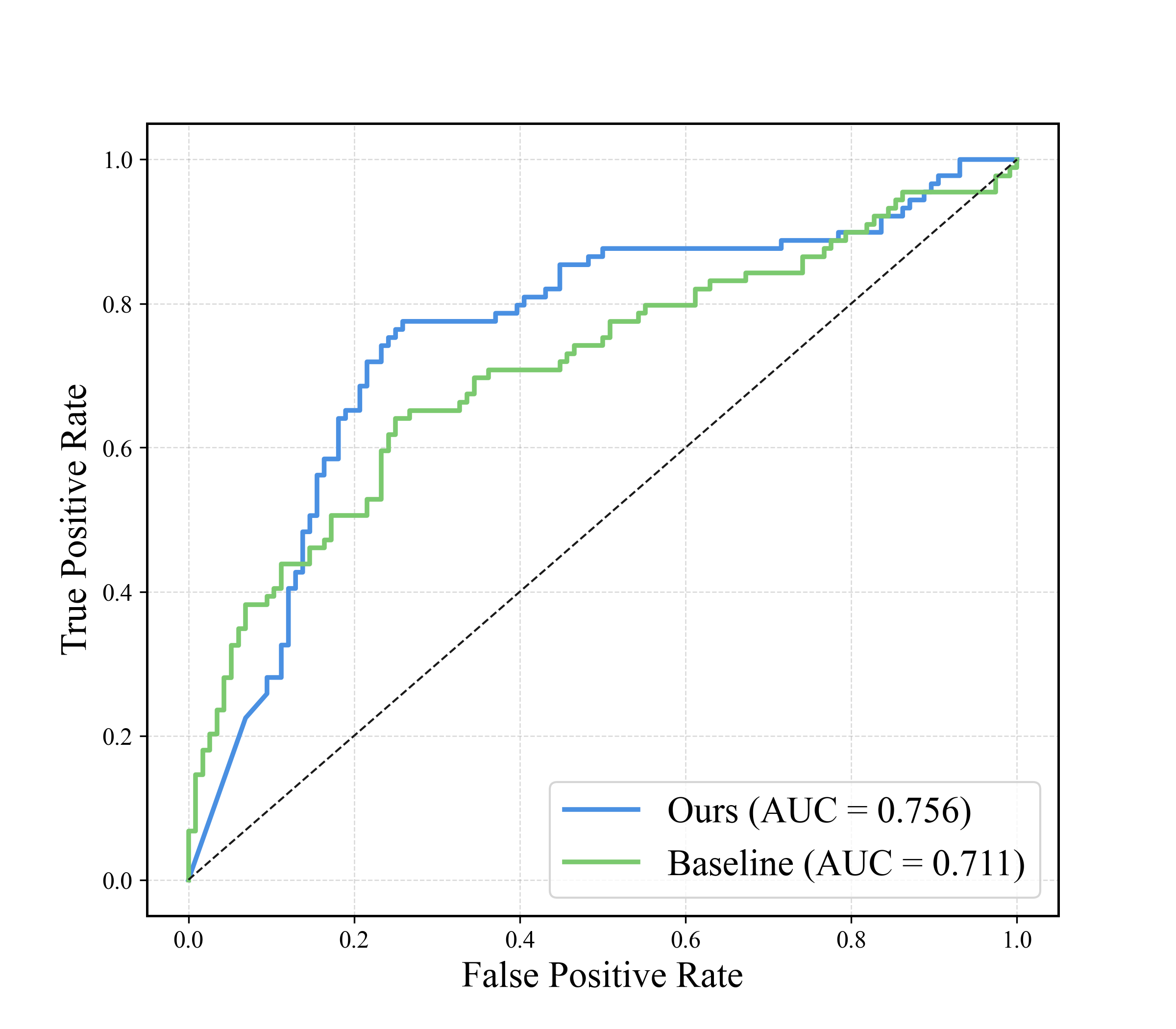}
    \caption{ROC curves; GITO augmentation (AUC = 0.756) versus baseline (AUC = 0.711)}
    \label{fig:gito-roc}
  \end{subfigure}\hfill
    \begin{subfigure}[t]{0.28\linewidth}
    \centering 
    \includegraphics[width=\linewidth]{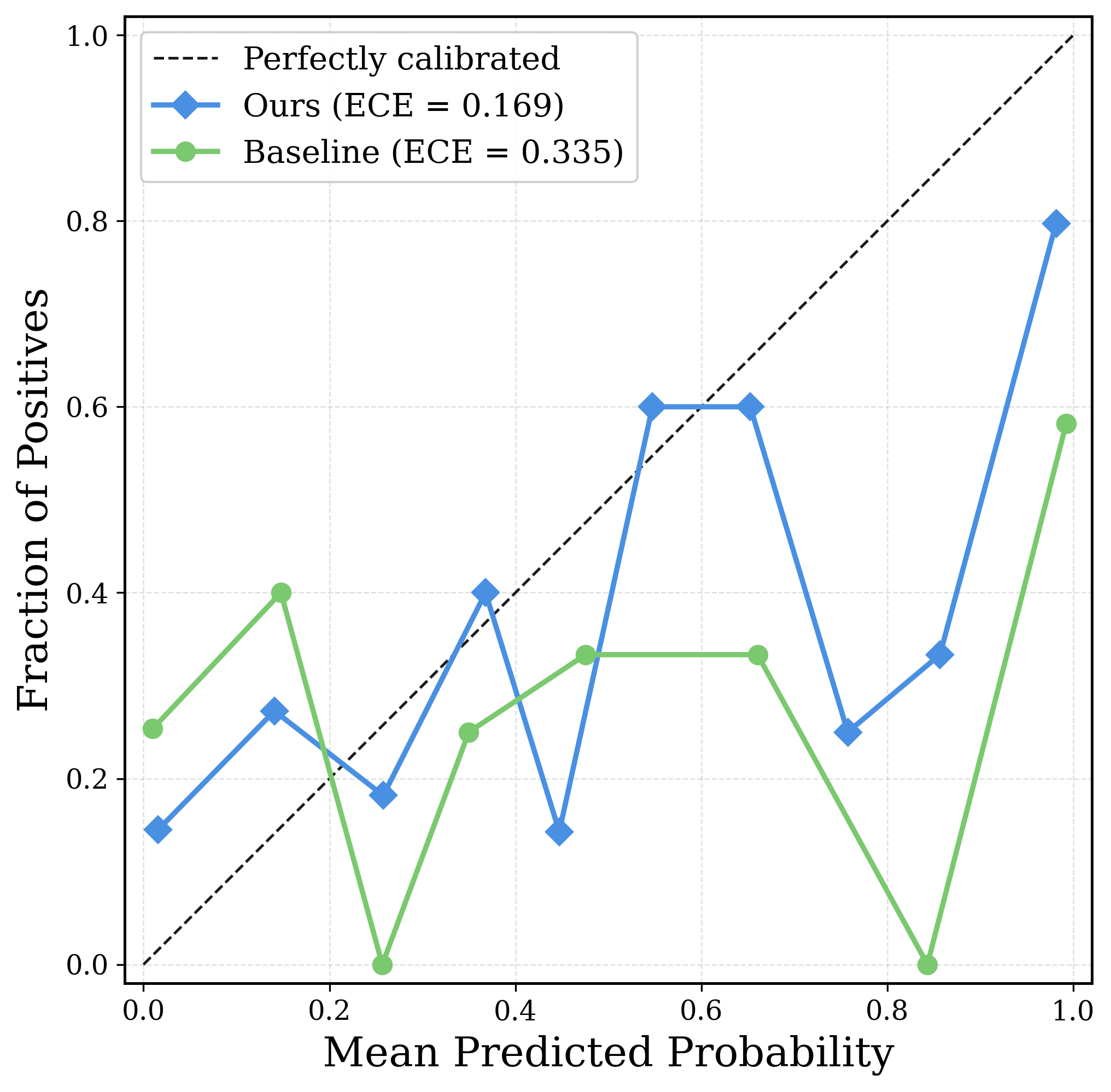}
    \caption{Calibration curves with expected calibration error 
    (GITO-augmented ECE\,=\,0.169 vs.\ baseline ECE\,=\,0.335).}
    \label{fig:gito-calibration}
  \end{subfigure}\hfill
  % ===================== (c) Error =====================
  \begin{subfigure}[t]{0.35\linewidth}
    \centering
    \includegraphics[width=\linewidth]{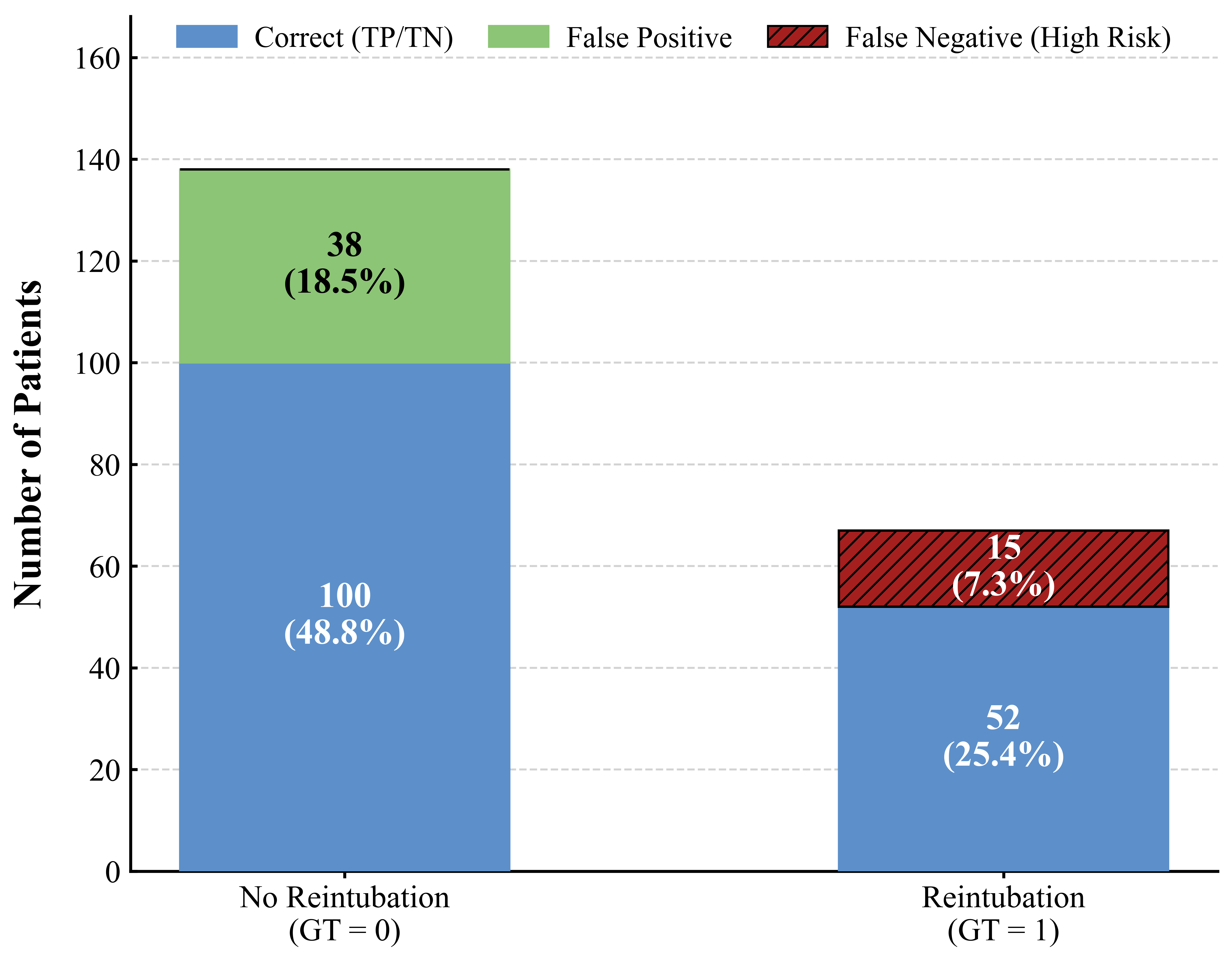}
    \caption{Confusion-matrix decomposition on the validation cohort 
    ($N = 205$), showing clinically relevant error patterns.}
    \label{fig:gito-error}
  \end{subfigure}

\caption{Performance evaluation of the GITO framework for ventilator 
  re-intubation prediction, demonstrating improved discriminative ability, 
  probability calibration, and clinically meaningful risk stratification.}
  
  \label{fig:gito_ventilator_weaning_evaluation}
\end{figure}

To evaluate clinical utility in a high-stakes scenario, we applied GITO to predict re-intubation risk within six hours of mechanical ventilation weaning in a cohort of 205 ICU patients (Figure~\ref{fig:gito_ventilator_weaning_evaluation}). Standard predictive models were augmented with GITO-projected six-hour physiological trajectories and compared against a historical-only baseline using 12 hours of retrospective data.

The GITO-augmented model improved performance across all four metrics (Figure~\ref{fig:gito-forest}): accuracy increased from 0.668 to 0.756, precision from 0.652 to 0.719, recall from 0.506 (95\% CI: 0.398-0.612) to 0.719 (95\% CI: 0.621-0.811), and the F1-score from 0.570 to 0.719. ROC analysis confirmed consistent improvement across decision thresholds, with the AUC increasing from 0.711 to 0.756 (Figure~\ref{fig:gito-roc}). The gains were most pronounced in the low-to-moderate false-positive region where clinical decisions typically operate.

Integration of GITO-generated trajectories addressed key limitations of historical-only models (Fig. \ref{fig:gito-forest}). The baseline model achieved a recall of only 0.506 (95\% CI: 0.398–0.612), missing nearly half of patients requiring re-intubation. Incorporating GITO projections increased recall to 0.719 (95\% CI: 0.621–0.811)—an improvement of 42\%. Precision improved from 0.652 to 0.719, accuracy increased from 0.668 to 0.756, and the F1-score rose from 0.570 to 0.719.

Calibration analysis further demonstrated that the GITO-augmented model 
produced more reliable risk estimates 
(Figure~\ref{fig:gito-calibration}). The expected calibration error 
decreased from 0.335 for the baseline to 0.169 for the GITO-augmented model. The baseline calibration curve exhibited pronounced non-monotonicity, with observed reintubation rates near zero in the 0.7-0.9 predicted-probability range, indicating severe overconfidence in its high-risk predictions. 
In contrast, the GITO-augmented model maintained a broadly monotonic relationship between predicted probabilities and observed event rates, with its calibration curve 
tracking closer to the diagonal across most risk strata.

Error decomposition analysis revealed a clinically favourable shift in 
the prediction error profile (Figure~\ref{fig:gito-error}). Of the 67 
patients who required re-intubation, the GITO-augmented model correctly 
identified 52 (25.4\% of the cohort) while misclassifying only 15 as 
low-risk false negatives (7.3\%). Among the 138 patients who did not 
require re-intubation, 100 (48.8\%) were correctly classified and 38 
(18.5\%) were false positives. Compared with the baseline model, which 
missed nearly half of re-intubation cases (recall 0.506), the 
GITO-augmented model reduced false negatives by 42\%, concentrating the 
majority of residual errors in the lower-acuity false-positive category.

\subsection{Case study: enhance AI model interpretability in a patient with septic shock}
\label{subsec:case_study}

\begin{figure}[H]
    \centering
    \includegraphics[width=\linewidth]{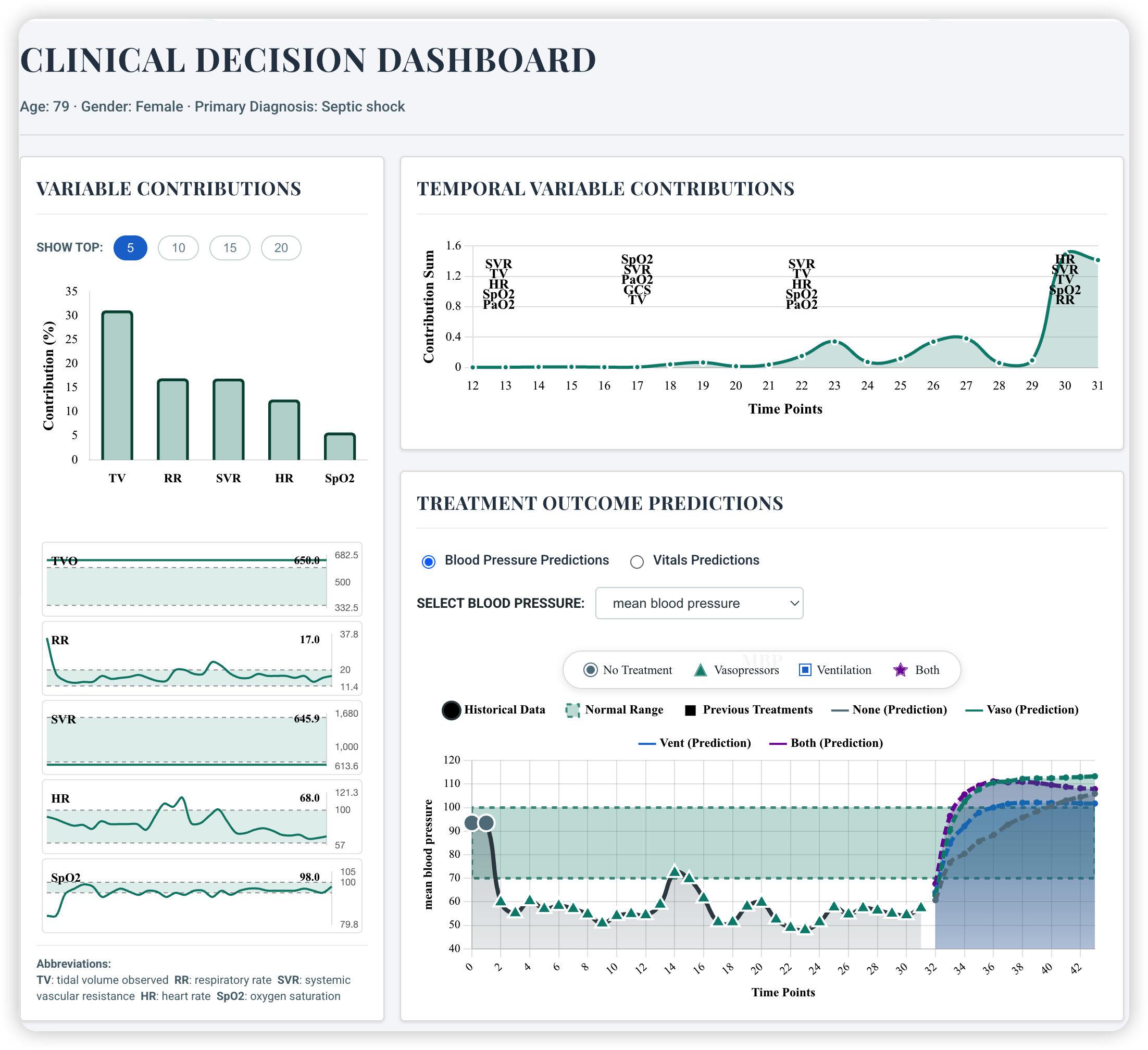}
    \caption{\textbf{Interpretable treatment outcome prediction for an 
individual ICU patient with septic shock.}
A single MIMIC-III patient (N\,=\,1) was analyzed using GITO 
(ACTIN-sMMD) to predict mean blood pressure (MBP) trajectories 
over 12 time steps ($\tau = 12$, corresponding to 12 hours) under 
four hypothetical treatment strategies (None, Vasopressor, 
Ventilation, Both).
\textbf{(Upper left)}~Top-five variables by integrated-gradient 
magnitude at the current time step.
\textbf{(Upper right)}~Temporal attribution: the five variables 
with the largest absolute integrated gradient at each historical 
time point, indicating their relative contribution to the 
predicted MBP.
\textbf{(Bottom)}~Observed MBP (black) and predicted MBP 
trajectories under the four treatment strategies. The 
LLM-generated clinical narrative is presented separately in 
Box~\ref{LLMreasoning}.}
\label{fig:case_study}
\end{figure}

We examined the interpretability of GITO through an individual-level case study of an ICU patient with septic shock,a life-threatening condition in which vasopressor therapy is administered to maintain mean blood pressure (MBP) within a target range that ensures adequate organ perfusion without overtreatment. We analyzed 12-hour MBP trajectory predictions under four alternative treatment strategies (Figure~\ref{fig:case_study}).
Variable attribution analysis identified tidal volume (contribution 0.20), respiratory rate (0.25), systemic vascular resistance (SVR; 0.18), heart rate (0.22), and oxygen saturation (0.15) as the five most influential variables at the current time point (Figure~\ref{fig:case_study}, upper left). Temporal attribution revealed that contribution magnitudes increased at later time points (upper right), with the largest effects concentrated in the final hours of the observation window.
Counterfactual trajectory analysis under four hypothetical strategies (no treatment, vasopressor only, ventilation only, both) revealed distinct MBP profiles (Figure~\ref{fig:case_study}, bottom). The no-treatment scenario projected gradual recovery toward the target range, indicating that the patient's underlying physiology was trending toward stabilization without additional intervention. Vasopressor administration accelerated this recovery but produced the most pronounced MBP rise, with the trajectory eventually exceeding the normal physiological range under sustained use. Ventilation alone yielded moderate improvement, and combined therapy projected the highest MBP, further above the target ceiling.
The framework's LLM-based explanation module (Box~\ref{LLMreasoning}) integrated historical MBP dynamics, variable attributions, and comparative trajectory analysis into a structured natural-language summary. Consistent with the trajectory patterns described above, the LLM assigned preference scores of 40\% to vasopressor use (rapid target attainment), 30\% to conservative management (gradual but sufficient recovery), 20\% to ventilation, and 10\% to combined therapy (risk of overshooting the target range). The treating clinicians chose not to escalate vasopressor therapy; the patient's blood pressure subsequently recovered to stable levels. Additional details about this case study are provided in Appendix~\ref{appendix:case_studies}.

\begin{tcolorbox}[
    title={\small\bfseries Example: LLM-generated 
    clinical rationale},
    colback=gray!5!white,
    colframe=gray!75!black,
    width=\columnwidth,
    left=4pt, right=4pt, top=3pt, bottom=3pt,
    boxsep=1.5pt
]
\label{LLMreasoning}
\small
\textit{Patient context:} MAP fluctuating below the 
65-85\,mmHg target range with transient dips consistent 
with septic shock. Top contributing variables: 
respiratory rate (0.25), heart rate (0.22), 
tidal volume (0.20), systemic vascular resistance (0.18), 
SpO\textsubscript{2} (0.15).

\textit{Counterfactual analysis:} Vasopressor-only 
(\textsc{Vaso}) predicts rapid MAP recovery into the 
target range; no-treatment (\textsc{None}) projects 
slower improvement; combined therapy (\textsc{Both}) 
risks overshooting.

\textbf{Treatment preference:} \textsc{Vaso}\,40\%\; 
\textsc{None}\,30\%\; \textsc{Vent}\,20\%\; 
\textsc{Both}\,10\%.

\hfill{\scriptsize Full output: 
Supplementary Box~\ref{appendix:full_LLMreasoning}}
\end{tcolorbox}

\subsection{Specialized GITO AI model outperform medical students and rule-guided LLMs in ventilator weaning prediction}

\begin{figure}[]
    \centering
    \includegraphics[width=\columnwidth]{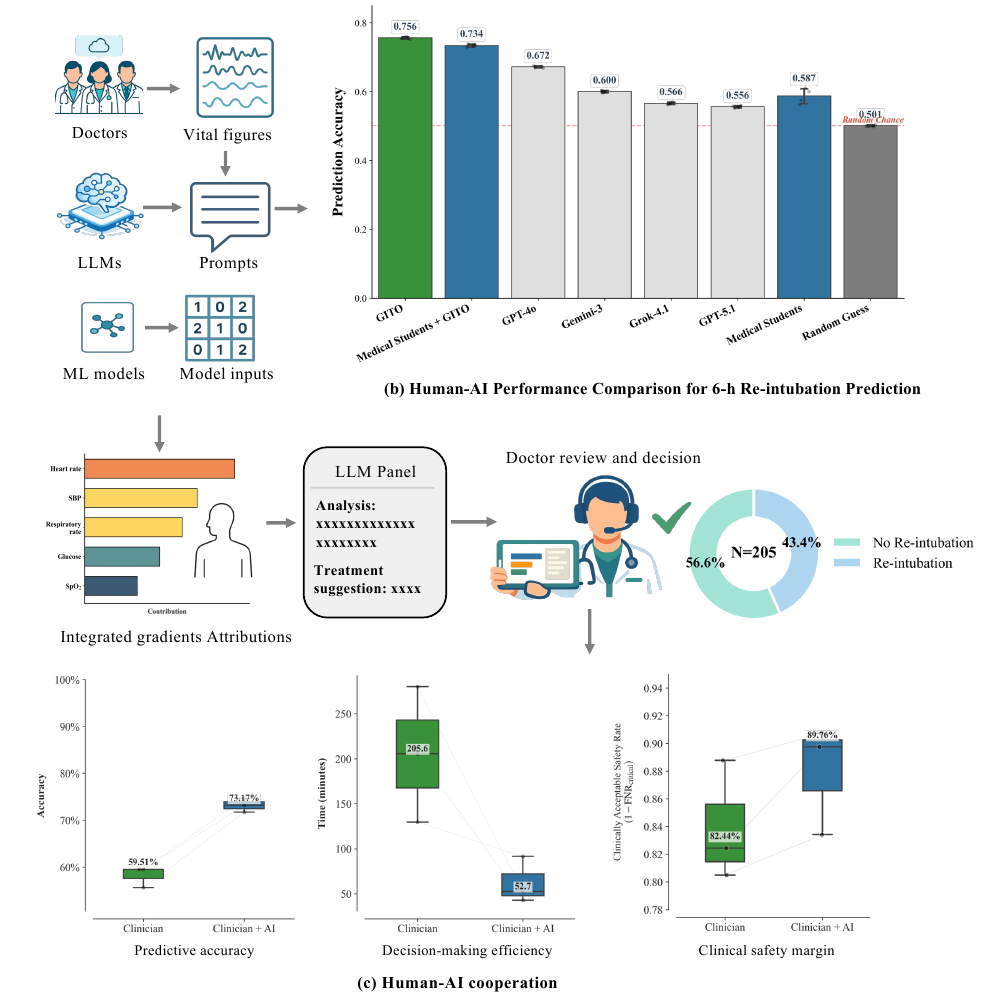}
    \caption{\textbf{Human-AI comparison and cooperation for ventilator weaning assessment based on predicted treatment outcomes.}
    \textbf{(a)}~Study design: medical students ($n = 3$), four large language models (GPT-4o, GPT-5.1, Gemini-3, Grok-4.1), and the GITO model each received patient vitals to predict 6-hour post-extubation trajectories, which were then used to assess re-intubation risk for $n = 205$ mechanically ventilated MIMIC-III patients (re-intubation prevalence, 43\%).
    \textbf{(b)}~Prediction accuracy (Top-1) across all agents; error bars denote standard deviation over $n = 4$ student participants or $n = 5$ independent model runs, as applicable.
    \textbf{(c)}~Human-AI cooperation outcomes from a two-period crossover study with $n = 3$ clinicians: predictive accuracy, decision-making time, and clinically acceptable safety rate ($1 - \text{FNR}$), compared between unassisted and GITO-assisted conditions (within-subject paired comparison).}
\label{fig:ai_human_comparison_cooperation}
\end{figure}

To benchmark GITO against human practitioners and general-purpose AI systems, we designed a prediction study in which medical students and junior clinicians ($n = \text{4}$) and four leading LLMs independently predicted re-intubation outcomes for the same 205-patient ventilator weaning cohort (Figure~\ref{fig:ai_human_comparison_cooperation}a; experimental details in Methods~\ref{subsec:human_ai_comparison}).
GITO achieved a prediction accuracy of 75.6\%, outperforming all LLMs and human participants (Figure~\ref{fig:ai_human_comparison_cooperation}b). Among LLMs equipped with expert clinical reasoning prompts, GPT-4o scored highest at 67.2\%, followed by Gemini-3 (60.0\%), Grok-4.1 (56.6\%), and GPT-5.1 (55.6\%). Unassisted medical students achieved 58.7\%, marginally above random chance (50.1\%).
When medical students were provided with GITO's predictions and attribution-based explanations, their accuracy increased from 58.7\% to 73.4\%, a 14.7-percentage-point improvement (Figure~\ref{fig:ai_human_comparison_cooperation}b). However, collaborative performance remained below GITO's standalone accuracy (75.6\%), with students occasionally overriding correct model predictions.

\subsection{Interpretable causal rationales of GITO models improved human clinicians' performance}

To assess whether GITO's interpretable outputs can enhance clinical decision-making, we conducted a controlled crossover experiment in which nine clinicians (three attending physicians, three residents, and three medical students) predicted re-intubation risk for all 205 mechanically ventilated patients in our test cohort (43\% requiring re-intubation; crossover design detailed in Methods~\ref{subsec:human_ai_comparison}).
GITO assistance improved clinician performance across all three evaluated dimensions (Figure~\ref{fig:ai_human_comparison_cooperation}c). Predictive accuracy increased from 59.5\% to 73.2\%, decision-making time decreased from 205.6 to 52.6 minutes per case batch, a 74\% reduction, and the clinically acceptable safety rate (1 $-$ FNR) rose from 82.4\% to 89.8\%. In the crossover analysis, clinicians who initially worked without AI assistance and then received GITO's predictions and attribution-based explanations revised a substantial proportion of their initial incorrect predictions in the AI-assisted round.

\subsection{Computational efficiency of GITO enables accessible and real-time clinical deployment}

To assess whether GITO's architectural simplification translates into practical deployment advantages, we compared the computational cost of GITO (ACTIN-sMMD) against the adversarial baseline (ACTIN) on identical hardware (Table~\ref{tab:computational_cost}). Replacing the discriminator with the sMMD module eliminated an auxiliary network, reducing the total parameter count by 3.25\% (120.0K to 116.1K) and converting the min-max Optimization into a single-objective problem. Per-epoch training time decreased by 21.1\% (5.12\,s to 4.04\,s), and the resulting stability improvement yielded a 9.0\% reduction in total training time (23.44 to 21.34\,min). At inference, prediction latency was 32.81\,ms per patient on CPU hardware.
This low-latency, CPU-compatible inference enabled us to deploy GITO as an open-source, web-based clinical interface (Figure~\ref{fig:framework_and_user_interface}) that integrates longitudinal vital-sign visualisation, counterfactual treatment simulation, and attribution-based explanation into a unified dashboard supporting real-time clinician interaction. The platform is publicly available\footnote{\url{https://huggingface.co/spaces/peisongzhang/TreatmentOutcomePredictionSystem}} and can be deployed within secure hospital intranets.

\begin{table*}[t]
\centering
\small
\caption{Computational cost comparison. GITO (ACTIN-sMMD) eliminates the discriminator network, reducing parameter count and inference latency. Although sMMD calculation adds slight per-epoch overhead, the model converges significantly faster, reducing total training time.}
\label{tab:computational_cost}

\resizebox{\textwidth}{!}{
\begin{tabular}{lccc}
\toprule
\textbf{Metric} & \textbf{ACTIN} & \textbf{ACTIN-sMMD} & \textbf{$\Delta$ (\%)} \\
\midrule
Params (K)                & 120.00 & 116.10 & $\downarrow$\,3.25 \\
Train / epoch (s)         & 5.12  & 4.04  & $\downarrow$\,21.10 \\
Total train (min)         & 23.44 & 21.34 & $\downarrow$\,8.96 \\
Latency (ms)              & 34.67 & 32.81 & $\downarrow$\,5.36 \\
Aux. networks             & Enc. + Disc. & Enc. only & Disc. removed \\
Stability                 & Min–max opt. & Single obj. & Improved \\
\bottomrule
\end{tabular}
}
\end{table*}

\subsection{sMMD effectively disentangles treatment bias from physiological heterogeneity in predicting treatment outcome}

\begin{table*}[t]
\centering
\caption{One-step-ahead prediction results ($\tau = 1$) on the synthetic tumor growth dataset under varying levels of time-varying confounding ($\gamma$). Values denote RMSE (mean $\pm$ standard deviation) over ten independent runs. Lower is better; best results are highlighted in bold.}
\label{tab:tg_one_step_prediction}
\begin{minipage}{\textwidth}
\centering
\resizebox{\textwidth}{!}{
\begin{tabular}{@{}lccccccccc@{}}
\toprule
      & $\gamma = 0$ & $\gamma = 1$ & $\gamma = 2$ & $\gamma = 3$ & $\gamma = 4$ & $\gamma = 5$ & $\gamma = 6$ & $\gamma = 7$ \\ \midrule
    CRN   &0.78$\pm$0.05 &0.82$\pm$0.05   &0.89$\pm$0.07  &1.13$\pm$0.17  &1.33$\pm$0.17  & 1.88$\pm$0.13 &2.43$\pm$0.16 &3.06$\pm$0.17 \\
    CRN-sMMD    &0.81$\pm$0.10 &0.85$\pm$0.07  &0.91$\pm$0.05  &1.15$\pm$0.11  &1.48$\pm$0.18  & 1.86$\pm$0.13 &2.46 $\pm$0.19 &3.14$\pm$0.24  \\ 
    CT    & 0.76$\pm$0.04 & 0.78$\pm$0.05 & 0.88$\pm$0.08 & 1.15 $\pm$0.17 & 1.62$\pm$0.29 & 2.06$\pm$0.26 & 2.46$\pm$0.29 &3.07$\pm$0.68 \\ 
    CT-sMMD &0.76$\pm$0.04  &0.78$\pm$0.05  &0.87$\pm$0.06  &1.16$\pm$0.15   &1.58$\pm$0.21& 1.95$\pm$0.25&2.39$\pm$0.17 &3.12$\pm$0.53 \\
    ACTIN & 0.75$\pm$0.06 & 0.78$\pm$0.04 & \textbf{0.85$\pm$0.07} & \textbf{1.01$\pm$0.13} & 1.29$\pm$0.20& 1.56$\pm$0.09 & 2.03$\pm$0.11 &2.76 $\pm$0.53\\
    ACTIN-sMMD &  \textbf{0.74$\pm$0.05} &\textbf{0.77$\pm$0.05} &\textbf{0.85$\pm$0.07} &1.02 $\pm$0.12 & \textbf{1.26$\pm$0.21} &\textbf{1.51$\pm$0.14} &\textbf{1.86$\pm$0.12} &\textbf{2.74$\pm$0.35}  \\ 
\bottomrule
\end{tabular}
}
\vspace{4mm}
\includegraphics[width=\textwidth]{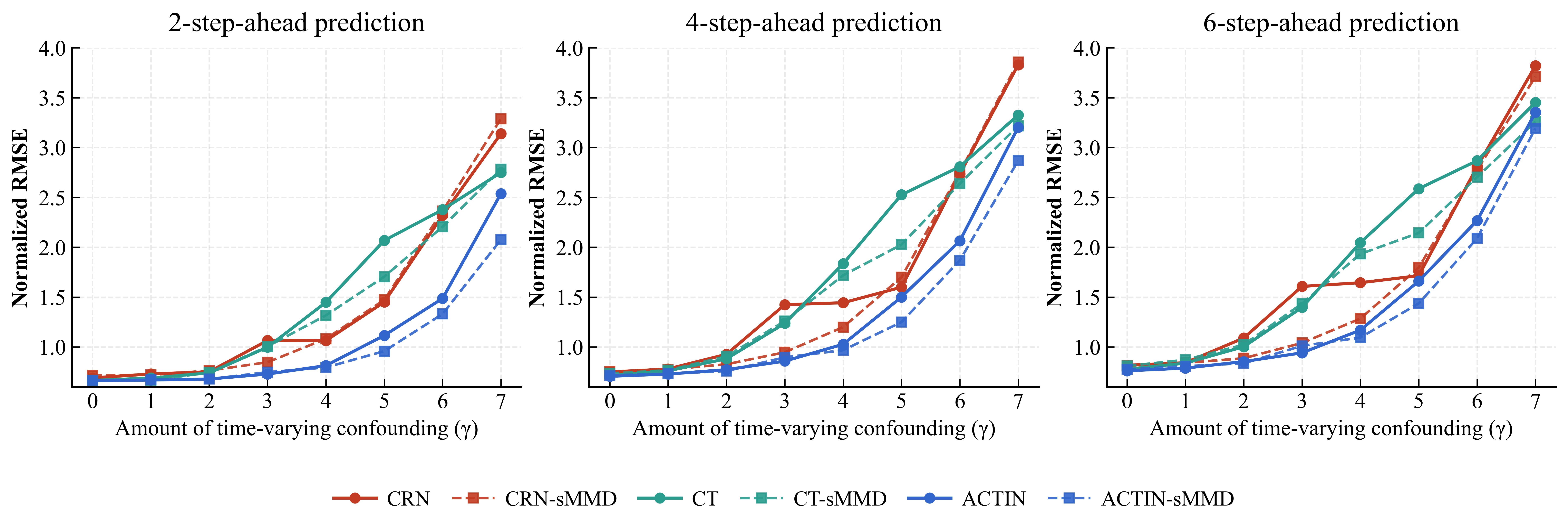} 
\captionof{figure}{Comparison of multi-step prediction performance (2-step, 4-step, and 6-step) across models on the synthetic tumor growth dataset under the single-treatment sliding window setting. Results are reported as the average RMSE over ten independent runs, across increasing levels of time-varying confounding strength $\gamma$.}
\label{fig:tg_multi_step_prediction}
\end{minipage}
\end{table*}

\begin{figure}[htbp]
\centering
\includegraphics[width=\textwidth]{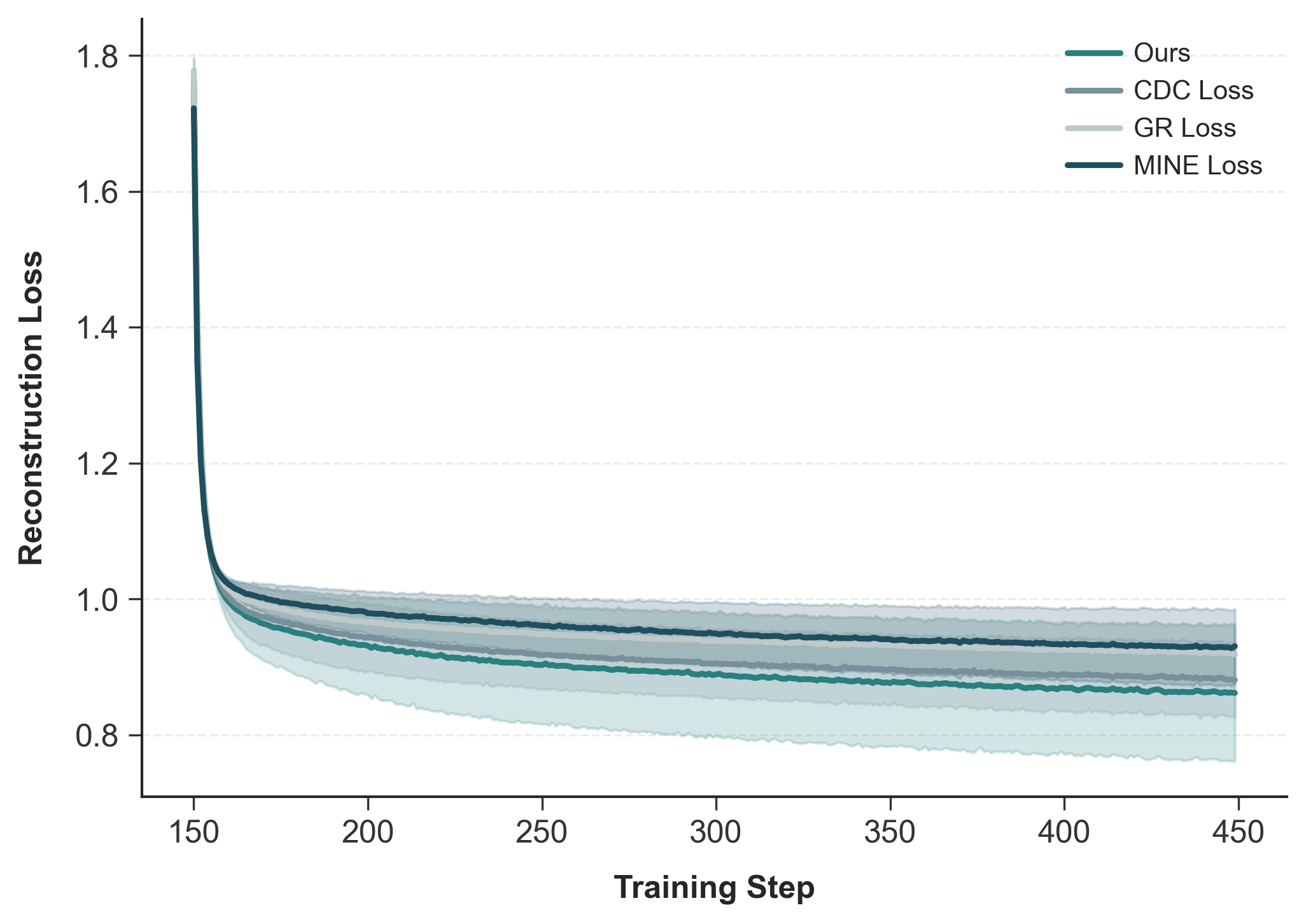}
\caption{Reconstruction loss during training for different balancing objectives. An independent decoder was trained to reconstruct the original patient co-variates from the balanced representations produced by each method. Lower reconstruction loss indicates greater preservation of patient-specific information. Shaded regions denote $\pm$1 standard deviation over ten independent runs.}
\label{fig:recon_loss}
\end{figure}

To evaluate sMMD under controlled confounding, we used a synthetic tumor growth dataset ($n = 10{,}000$) with an adjustable time-varying confounding parameter ($\gamma$). Baseline architectures (CRN, CT, ACTIN) were compared against their sMMD-enhanced counterparts across confounding levels $\gamma = 0$-$7$.
For one-step-ahead predictions ($\tau = 1$), sMMD conferred only marginal improvements at low confounding levels (Table~\ref{tab:tg_one_step_prediction}). At $\gamma \geq 4$, sMMD-enhanced models began to show consistent gains, with ACTIN-sMMD achieving the lowest RMSE at six of eight confounding levels. The advantage of sMMD became more pronounced at longer prediction horizons and higher confounding (Figure~\ref{fig:tg_multi_step_prediction}). At low confounding ($\gamma < 3$), all models performed comparably across all horizons. Beyond this threshold, baseline RMSE rose steeply with increasing $\gamma$ and $\tau$, whereas sMMD variants exhibited a flatter degradation trajectory. ACTIN-sMMD yielded the largest RMSE reductions relative to ACTIN at moderate-to-high confounding: at $\gamma = 5$, reductions were 14.1\% ($\tau = 2$), 16.7\% ($\tau = 4$), and 13.5\% ($\tau = 6$); at $\gamma = 7$, the short-horizon gain reached 18.2\% ($\tau = 2$), though the margin narrowed at longer horizons (10.5\% and 4.8\% at $\tau = 4$ and $\tau = 6$, respectively) as both models degraded under extreme confounding. The benefit of sMMD was consistent across architectures: CT-sMMD reduced RMSE relative to CT by 17.1\% at $\gamma = 5$, $\tau = 6$, though CRN-sMMD showed more variable gains across settings.

To quantify how much patient-specific information each balancing strategy retains, we trained an independent decoder to reconstruct the original co-variates from balanced representations (Figure~\ref{fig:recon_loss}). Among all balancing objectives tested,domain confusion (CT), gradient reversal (CRN), mutual-information-based loss (ACTIN/MINE), and sMMD,sMMD-balanced representations achieved the lowest reconstruction error throughout training, with the gap widening as training progressed.

\subsection{GITO balances predictive accuracy with fairness and bias mitigation}

\begin{figure}[htbp]
  \centering
  \begin{subfigure}[t]{0.32\linewidth}
    \centering
    \includegraphics[width=\linewidth]{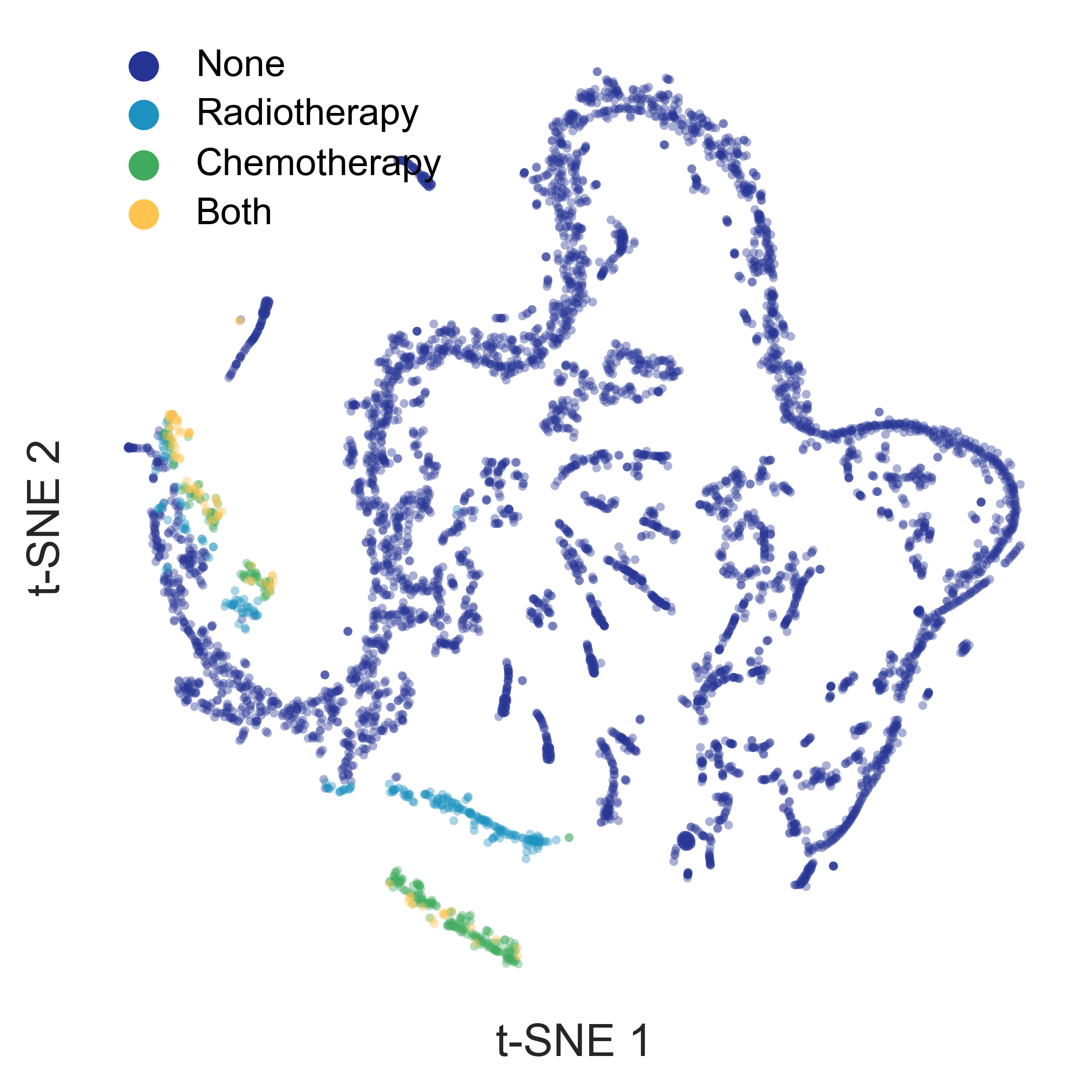}
    \caption{Without balancing: treatment-specific clustering.}
    \label{fig:representation_woBRM_visualization}
  \end{subfigure}\hfill
  \begin{subfigure}[t]{0.32\linewidth}
    \centering
    \includegraphics[width=\linewidth]{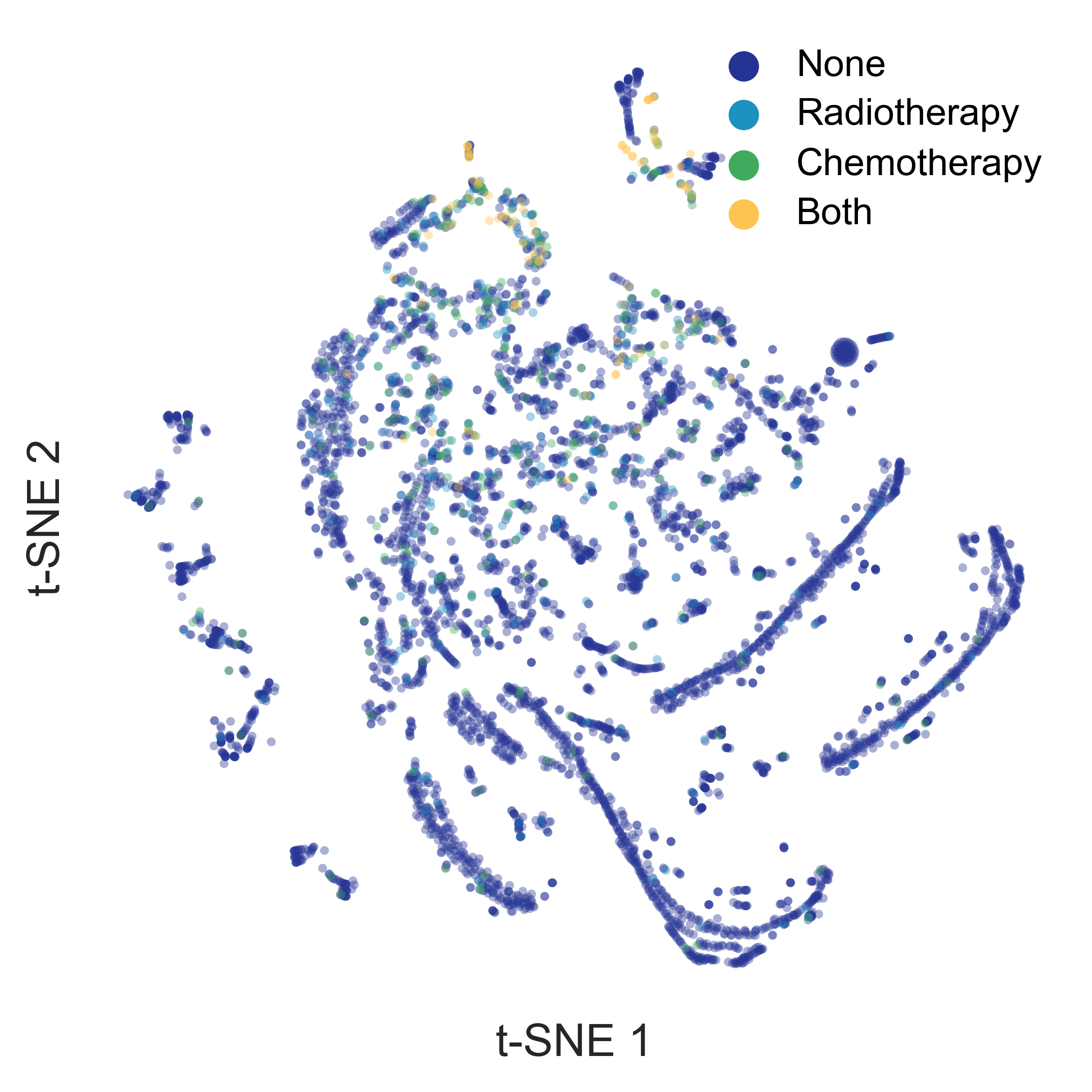}
    \caption{With sMMD: well-mixed treatment groups.}
    \label{fig:mmd_visualization}
  \end{subfigure}\hfill
  \begin{subfigure}[t]{0.32\linewidth}
    \centering
    \includegraphics[width=\linewidth]{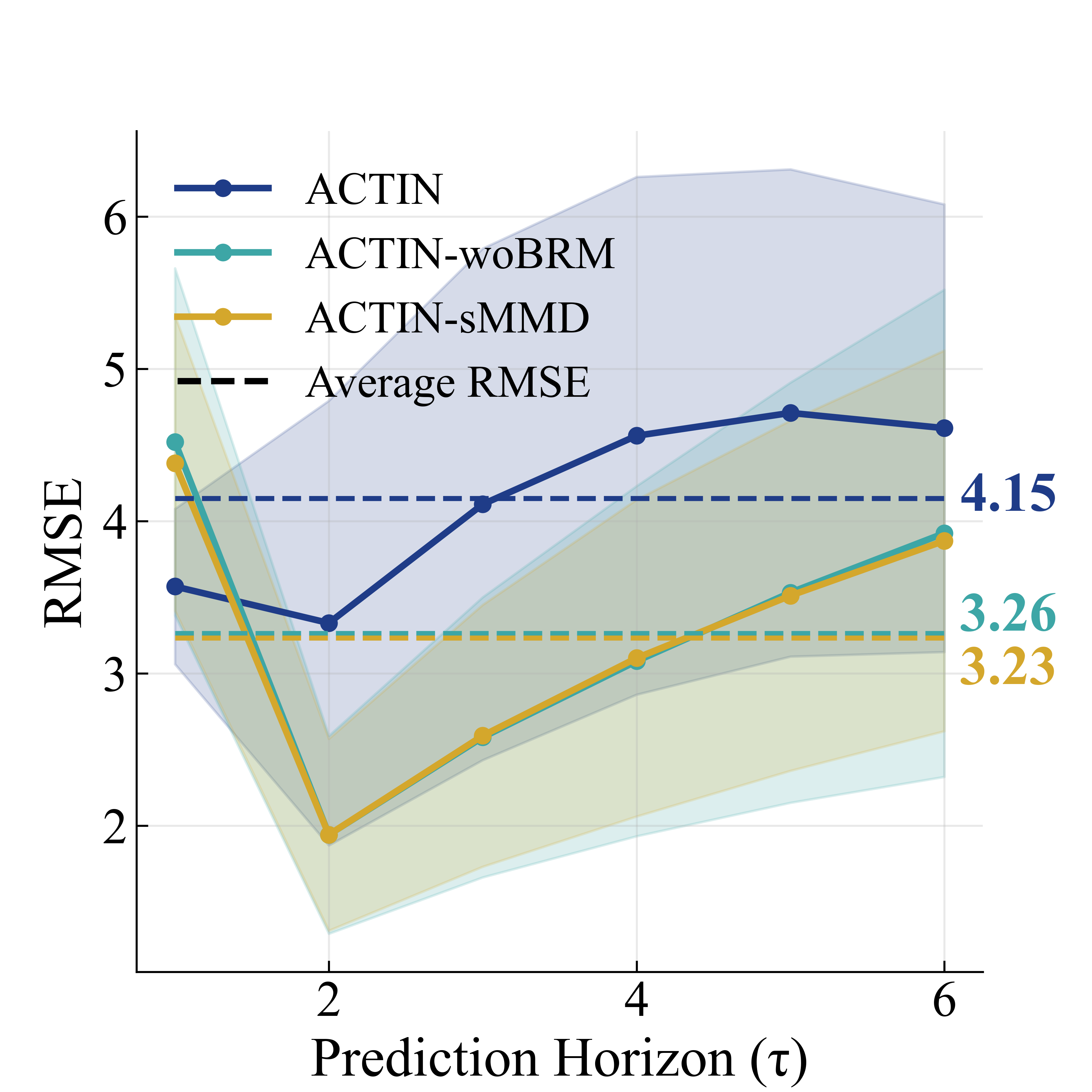}
    \caption{Multi-step RMSE comparison ($\gamma = 10$).}
    \label{fig:balancing_rmse_comparison}
  \end{subfigure}

  \vspace{0.8em}

  \begin{subfigure}[t]{\linewidth}
    \centering
    \includegraphics[width=\linewidth]{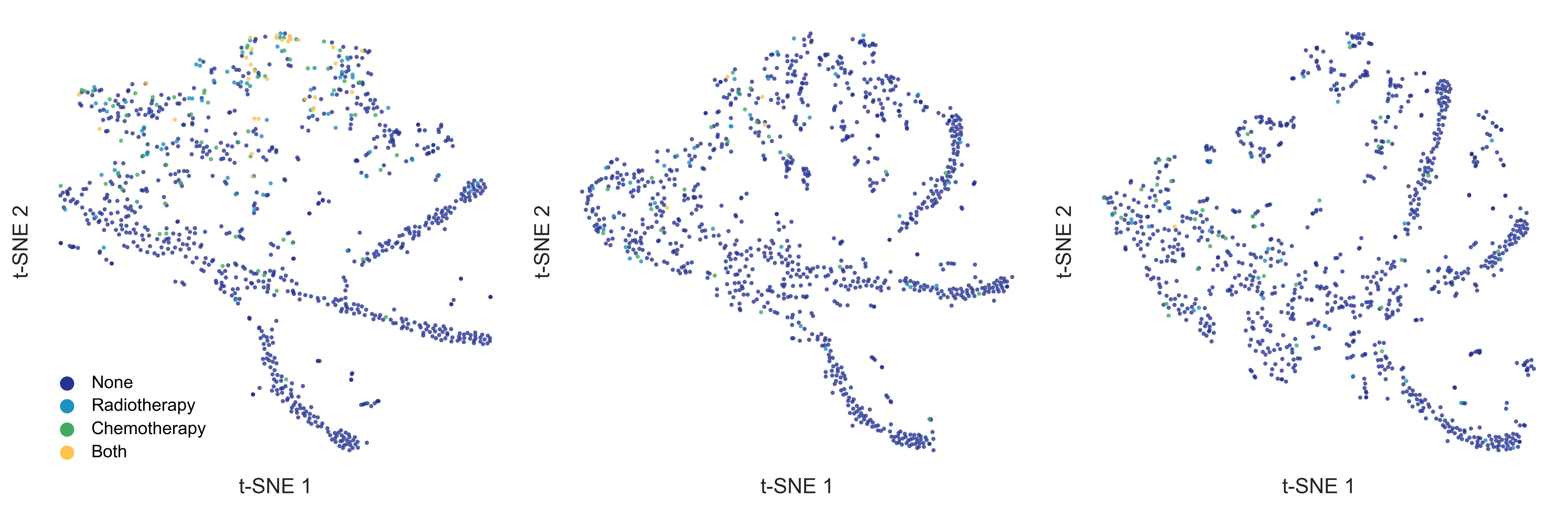}
    \caption{Temporal evolution of sMMD-balanced representations across time steps, showing that the learned representations remains balanced and consistent throughout the sequence.}
    \label{fig:mmd_time_step_visualization}
  \end{subfigure}

  \vspace{0.8em}

  \begin{subfigure}[t]{0.32\linewidth}
    \centering
    \includegraphics[width=\linewidth]{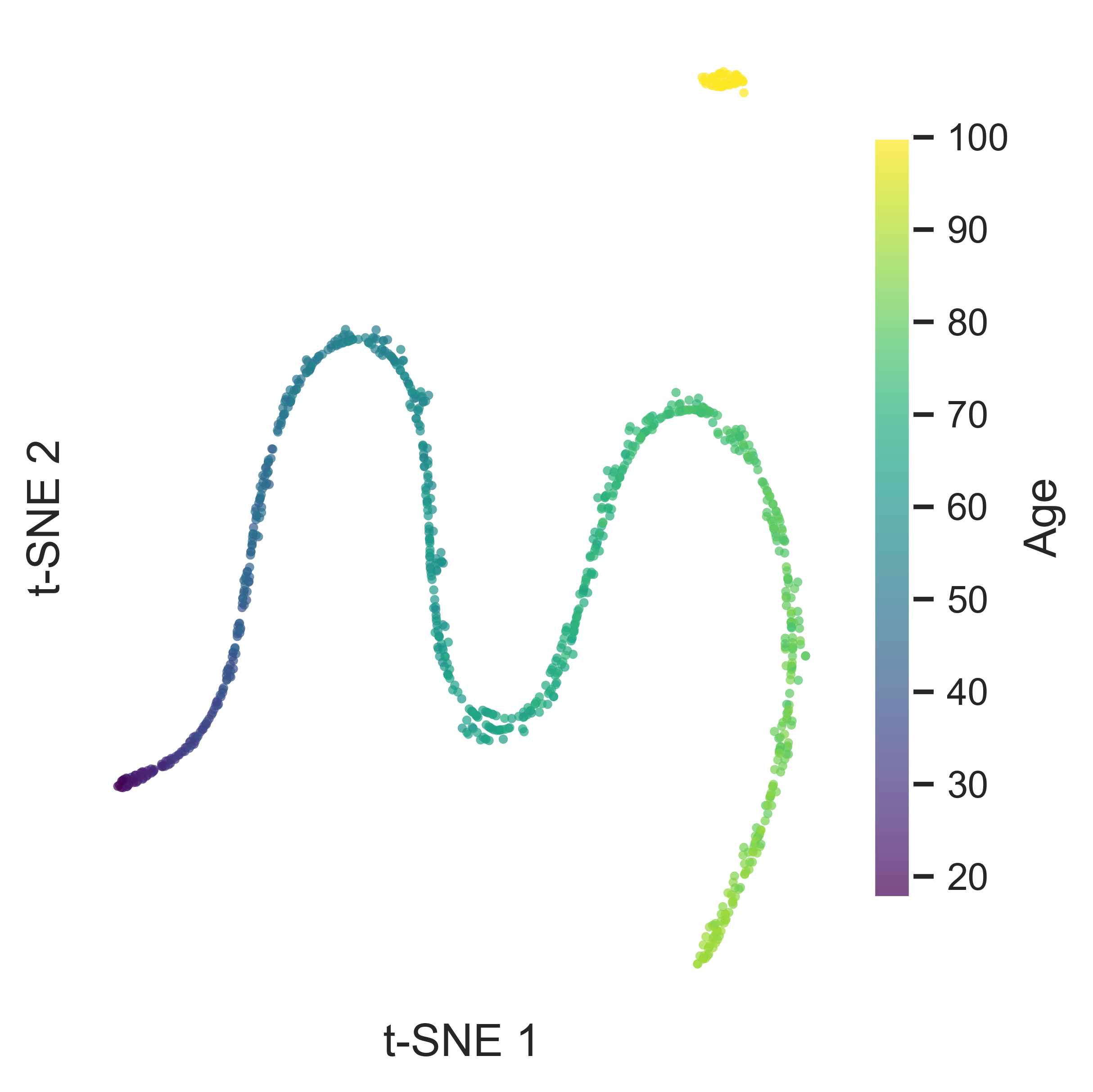}
    \caption{Colored by age. A mild gradient structure is observed, suggesting that age may influence the physiological trajectories and potentially affect treatment outcomes.}
    \label{fig:tsne_by_Age}
  \end{subfigure}\hfill
  \begin{subfigure}[t]{0.32\linewidth}
    \centering
    \includegraphics[width=\linewidth]{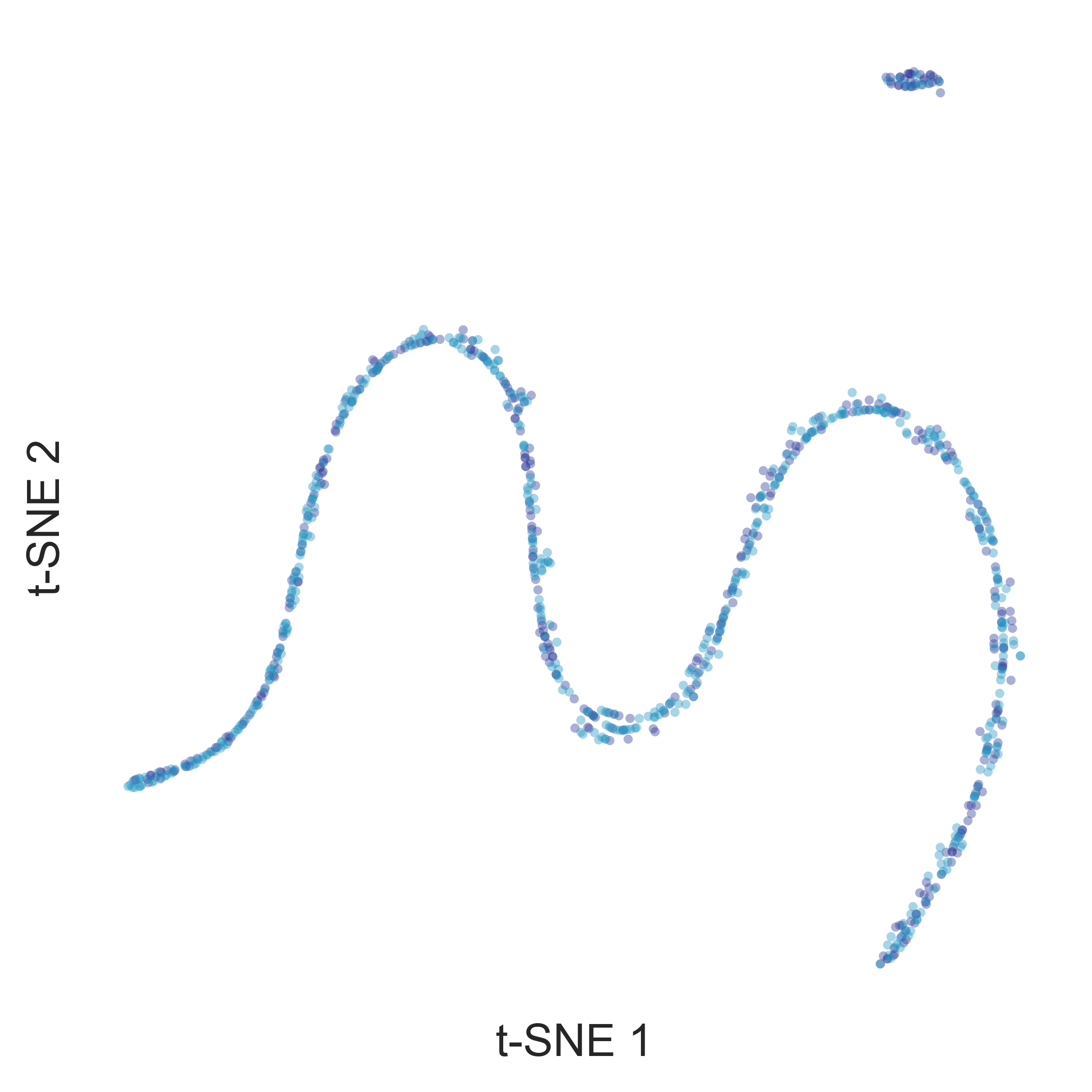}
    \caption{Colored by gender. Male and female patients are well mixed in the embedding space, indicating no observable gender bias in the learned representations.}
    \label{fig:tsne_by_Gender}
  \end{subfigure}\hfill
  \begin{subfigure}[t]{0.32\linewidth}
    \centering
    \includegraphics[width=\linewidth]{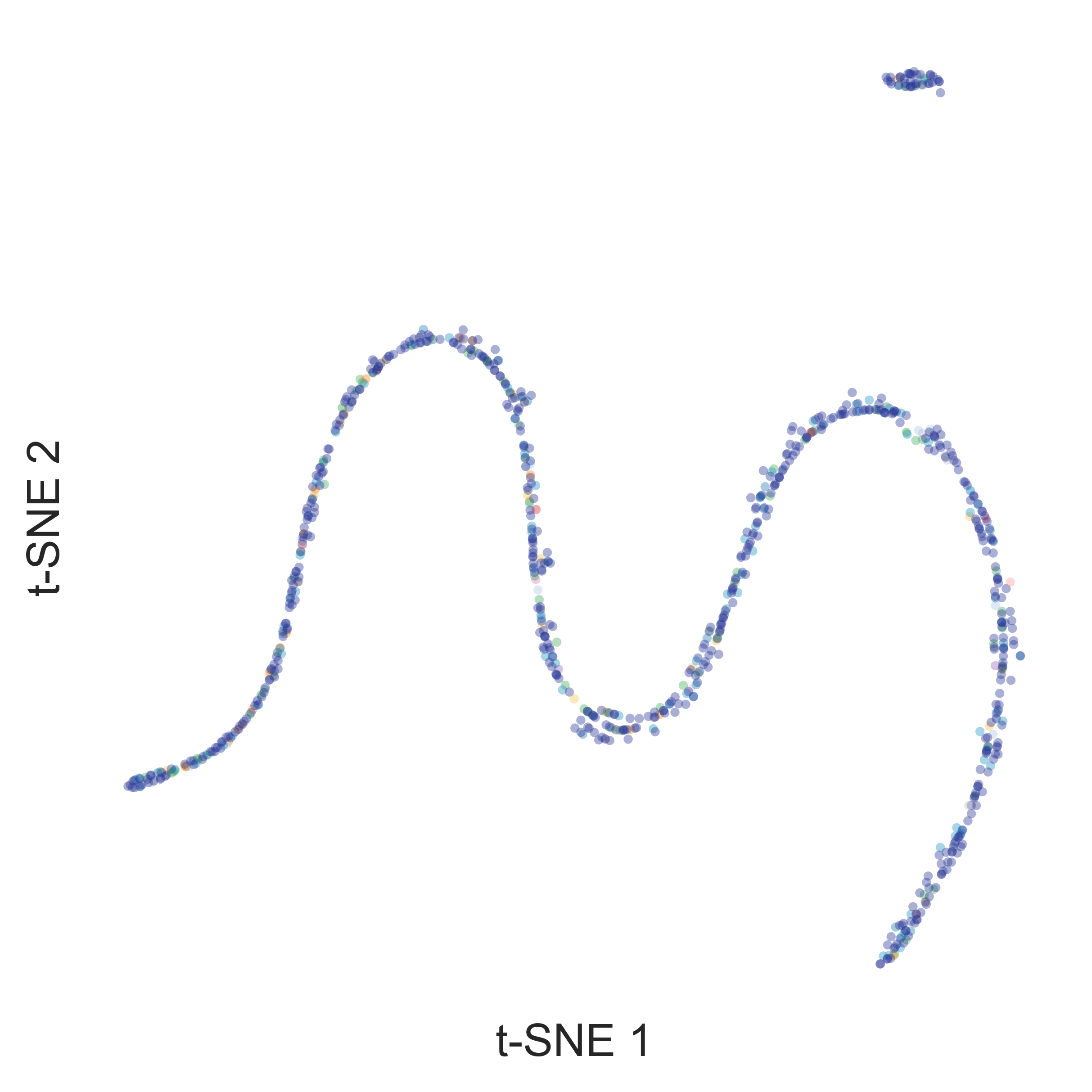}
    \caption{Colored by ethnicity. No distinct clustering is observed among ethnic groups, suggesting that the model captures clinically relevant features rather than demographic-specific patterns.}
    \label{fig:tsne_by_Ethnicity}
  \end{subfigure}

  \caption{\textbf{Visualization of learned representations and performance trade-offs.} (a-b) t-SNE embeddings of the synthetic dataset showing treatment-specific clustering without balancing (a) versus well-mixed distributions with sMMD balancing (b). (c) Multi-step RMSE comparison ($\gamma=10$) showing sMMD achieves the best trade-off between bias removal and accuracy. (d) Temporal stability of sMMD-balanced representations. (e-g) Embeddings of MIMIC-III data colored by Age, Gender, and Ethnicity. The structured gradient in Age (e) contrasts with the uniform mixing in Gender (f) and Ethnicity (g), indicating the preservation of clinically relevant physiology over demographic bias.}
  \label{fig:representation_visualization}
\end{figure}

To examine whether sMMD removes treatment-assignment bias while preserving clinically relevant structure, we visualized learned representations using t-SNE on both synthetic and real-world data (Figure~\ref{fig:representation_visualization}).
Under severe confounding ($\gamma = 10$), we compared three model variants: adversarial balancing (ACTIN), no balancing (ACTIN-woBRM), and sMMD balancing (ACTIN-sMMD) across prediction horizons $\tau = 1$-$6$ (Figure~\ref{fig:balancing_rmse_comparison}). Adversarial balancing yielded the highest average multi-step RMSE (4.15) and exhibited substantial instability: RMSE spiked at $\tau = 2$ before partially recovering, and the variance across runs was markedly wider than for either alternative. Removing balancing entirely produced a lower average RMSE (3.26) with a smooth, monotonically increasing trajectory, but left treatment-group distributions clearly separated in the latent space (Figure~\ref{fig:representation_woBRM_visualization}). ACTIN-sMMD achieved the lowest average RMSE (3.23) with the narrowest cross-run variance. Its per-horizon trajectory closely paralleled the unbalanced model, indicating that sMMD corrected for treatment-assignment bias without incurring an accuracy penalty. The resulting representations showed well-mixed treatment groups (Figure~\ref{fig:mmd_visualization}). Temporal visualisation confirmed that this alignment was sustained across all sequential time steps (Figure~\ref{fig:mmd_time_step_visualization}), with no re-emergence of treatment-specific clustering at later horizons.
To assess whether the balanced representations encode demographic biases, we coloured the MIMIC-III embeddings by patient attributes (Figure~\ref{fig:tsne_by_Age}-\ref{fig:tsne_by_Ethnicity}). Gender and ethnicity showed uniform mixing across the representation space, with no observable clustering by demographic group. In contrast, age exhibited a structured gradient, with patients over 90~years forming a distinct cluster (Figure~\ref{fig:tsne_by_Age}, upper right). Taken together, the representations did not exhibit systematic demographic partitioning; the age-related structure was the sole axis of separation, consistent with preserved physiological heterogeneity rather than encoded demographic bias.

\section{Discussion}
\label{sec:discussion}

% \textcolor{red}{(DL: briefly state the problem again before describe the solution)}

A major barrier to deploying AI in intensive care is that current causal models often sacrifice patient-specific information to reduce confounding bias, limiting generalization across populations. This study introduced GITO, a framework that resolves this trade-off through a sampling-based MMD (sMMD) alignment strategy. Across synthetic and real-world ICU datasets, GITO demonstrated robust out-of-distribution performance, including cross-hospital and cross-ethnicity generalization, and achieved predictive accuracy comparable to or exceeding that of experienced clinicians. Its interpretable outputs enhanced clinical reasoning and improved the performance of less experienced physicians, underscoring its role as an augmentation tool for clinical expertise.

From a methodological perspective, GITO addresses the long-standing tension between confounding removal and information preservation in treatment outcome prediction. 
Adversarial balancing methods enforce global distributional invariance, which suppresses treatment-related signals indiscriminately, including the clinically informative heterogeneity essential for individualized prediction. The sMMD strategy circumvents this by performing stochastic, sample-level alignment: at each iteration, small random subsets are drawn from each treatment group and aligned via MMD, imposing a softer constraint that encourages the encoder to learn treatment-invariant features without overwriting covariate-level detail.
Three lines of evidence support this interpretation. First, synthetic experiments under strong confounding revealed that sMMD prevented the representation space from encoding treatment-assignment artefacts while preserving patient-level variation needed for accurate long-range prediction, whereas adversarial objectives progressively eroded this information. Second, reconstruction analyses confirmed that sMMD-balanced representations achieved the lowest reconstruction error across all balancing objectives, indicating that downstream outcome heads retained access to richer, patient-specific features. Third, per-variable $\Delta R^{2}$ analysis on MIMIC-III data showed that adversarial MINE balancing exhibited high variance across clinical variables, achieving large negative $\Delta R^{2}$ on some (e.g., HCO$_3$) while incurring substantial positive $\Delta R^{2}$ on others (e.g., PaO$_2$, glucose), consistent with an invariance objective that non-selectively reshapes the representation space. In contrast, sMMD produced a consistently near-zero $\Delta R^{2}$ profile and, notably, achieved negative $\Delta R^{2}$ on respiratory rate, FiO$_2$, and GCS, variables that the adversarial method eroded, suggesting that stochastic sub-sampling acts as an implicit regularizer that enhances reconstruction fidelity on these clinically decisive variables. This selective preservation of variables central to ventilator management (PaO$_2$, FiO$_2$, respiratory rate), haemodynamic monitoring (heart rate), and consciousness assessment (GCS) provides a mechanistic account for the observed generalization advantage across hospitals, ethnicities, and disease categories.

Clinically, the ventilator weaning experiment illustrates how GITO's predicted trajectories translate methodological gains into patient-level benefit. The 42\% reduction in high-risk false negatives,from a baseline recall of 0.506 to 0.719,indicates that forward-projected physiological trajectories captured deterioration signals absent from retrospective data alone. In safety-critical settings such as ventilator weaning, where missed re-intubation events may lead to delayed intervention and adverse outcomes, this shift toward fewer missed detections is of direct clinical consequence. The concurrent improvement in AUC (0.711 to 0.756) further confirms that these gains are not achieved at the expense of increased false alarms. Importantly, calibration analysis revealed that the GITO-augmented model not only improved discrimination but also produced substantially more reliable probability estimates (ECE = 0.169 vs.\ 0.335 for the baseline). The baseline's calibration curve exhibited pronounced non-monotonicity, with observed reintubation rates near zero in the 0.7-0.9 predicted-probability range, indicating that patients flagged as ``high-risk'' by the baseline were, in practice, rarely reintubated. This form of miscalibration is particularly hazardous in clinical settings, as it may drive unnecessary interventions based on inflated risk estimates. In contrast, the GITO model maintained a broadly monotonic relationship between predicted and observed event rates, enabling clinicians to interpret its probability outputs as meaningful risk estimates rather than ordinal rankings. This property is a prerequisite for shared decision-making, where the absolute magnitude of predicted risk, not merely relative ordering, directly informs the aggressiveness of subsequent management.

Beyond prediction accuracy, the interpretability framework enhanced clinical reasoning. The septic shock case study demonstrated how quantitative attribution, counterfactual trajectory analysis, and LLM-generated explanations jointly enabled clinicians to interpret treatment trade-offs, providing a coherent reasoning pathway that aligned model predictions with clinically meaningful narratives rather than isolated feature-importance scores. Critically, the counterfactual trajectories revealed that the patient's blood pressure was trending toward recovery even without intervention, while sustained vasopressor use would accelerate recovery but risk overshooting the target range. This multi-scenario view enables a clinically important inference that neither trajectory alone would support: short-term vasopressor administration to hasten stabilization, followed by timely de-escalation to avoid over-treatment, a nuanced strategy that goes beyond binary treat-or-not decisions. The fact that treating clinicians independently chose conservative management, with the patient subsequently recovering, corroborates the clinical relevance of GITO's trajectory-based reasoning.

The human-AI comparison study further contextualizes these advantages. GITO outperformed all four general-purpose LLMs by 8.4-19.0 percentage points despite these models being equipped with expert clinical reasoning prompts. This performance gap is consistent with a fundamental limitation of prompt-based approaches: while LLMs can apply rule-based logic to static clinical snapshots (e.g., checking RSBI thresholds), they lack the capacity to model the non-linear temporal dynamics,such as gradual haemo-dynamic drift or evolving respiratory patterns, that GITO's sMMD-balanced representations are specifically trained to encode. When medical students were provided with GITO's predictions and attribution-based explanations, their accuracy improved by 14.7 percentage points (from 58.7\% to 73.4\%), demonstrating that the framework's explanations are sufficiently interpretable to improve novice clinical judgment. However, collaborative accuracy remained below GITO's standalone performance (75.6\%), indicating imperfect trust calibration: students occasionally overrode correct model predictions based on their own assessment. This suggests that effective deployment requires not only transparent explanations but also calibration mechanisms that help users recognize when to defer to algorithmic judgment.

The cooperation study with practicing clinicians reinforced these findings while revealing additional benefits. GITO assistance reduced decision-making time by 74\% (from 205.6 to 52.6 minutes per case batch) and improved the clinically acceptable safety rate from 82.4\% to 89.8\%. The efficiency gain likely reflects the role of attribution-based explanations in directing clinicians' attention toward the most prognostically relevant variables, reducing the cognitive burden of manually reviewing high-dimensional temporal data. Importantly, the crossover design revealed that clinicians actively revised their initial incorrect predictions after reviewing GITO's explanations, rather than passively accepting the model's output. This distinction is clinically meaningful: it suggests that GITO's interpretable outputs engage clinicians in a corrective reasoning process, enabling them to identify errors in their own assessment rather than merely deferring to the algorithm. The simultaneous gains in accuracy, efficiency, and safety position GITO as a decision-support tool that strengthens human judgment by exposing the temporal dynamics and comparative consequences of alternative interventions.

Translating these clinical benefits into practice requires both computational feasibility and demographic fairness. On the engineering side, the replacement of adversarial min-max optimization with a single-objective sMMD loss yielded concrete deployment advantages: fewer trainable parameters, faster convergence, an,  critically, sub-50\,ms inference latency on standard CPU hardware without GPU acceleration. This low computational footprint enabled us to release GITO as an open-source, web-based platform that can be deployed within secure hospital intranets, lowering the barrier to adoption in resource-limited settings where access to both experienced intensivists and specialized computing infrastructure is constrained. On the fairness side, the representation analysis confirmed that sMMD-balanced embeddings showed no systematic partitioning by gender or ethnicity, demographic attributes that should not influence physiological predictions. The sole axis of structured separation was age, where patients over 90 years formed a distinct cluster; rather than indicating bias, this pattern is consistent with the well-established physiological distinctiveness of advanced age, including reduced organ reserve and altered pharmacokinetics, which legitimately influence treatment response. The contrast, mixing on demographics while preserving clinically meaningful age-related heterogeneity, indicates that sMMD's stochastic alignment selectively targets treatment-assignment confounding without collapsing the physiological variation that underlies individualized prediction. Together, the computational efficiency and demographic neutrality of the framework support its readiness for broader clinical adoption. The same methodological foundation, sMMD alignment and attribution-grounded interpretability, is domain-agnostic and may extend to non-ICU settings, including emergency triage, ward-level monitoring, and chronic disease management.

Several limitations of this study should be acknowledged, each pointing toward directions for future research. First, although GITO was evaluated on two geographically distinct cohorts (MIMIC-III and AmsterdamUMCdb), both are retrospective observational datasets; prospective validation in a randomised or pragmatic clinical trial setting remains necessary to confirm real-world benefit and to quantify the effect of GITO-assisted decision-making on patient outcomes. Second, the current sMMD framework is designed for binary treatment decisions. Extending GITO to continuous treatment variables, such as drug dosages or infusion rates, through kernel-based propensity matching would enable dosage optimization and graduated intervention strategies. Third, our hourly modeling intervals may provide insufficient temporal granularity for time-sensitive interventions such as vasopressor administration, where clinical effects manifest within minutes. Multi-resolution temporal modeling could address this limitation while expanding applicability to additional high-impact interventions, including antibiotic selection and renal replacement therapy timing. Fourth, while the attribution-grounded LLM explanations improved clinician performance in our benchmark, they were not evaluated for factual accuracy against established clinical guidelines; the risk of LLM hallucination, even when constrained by model attributions, cannot be fully eliminated, and structured evaluation against clinical knowledge bases represents an important next step. Fifth, ensuring temporal validity poses a fundamental challenge: clinical best practices evolve continuously as new evidence emerges, yet the current framework relies on historical data. Future work will explore mechanisms for safely incorporating newly generated clinical evidence while enabling privacy-preserving model updates, transforming GITO from a static prediction tool into a continuously learning clinical decision-support system.

By bridging methodological innovation with clinical relevance, GITO represents a step toward trustworthy, globally accessible AI for personalized treatment optimization in critical care and beyond.

\section{Methods}
\label{sec:methods}

\subsection{Study design}
\label{subsec:study_design}
We evaluated GITO on three patient cohorts spanning varying levels of complexity and real-world variability: two real-world ICU databases (MIMIC-III and AmsterdamUMCdb) and one synthetic tumor growth dataset.

\textbf{MIMIC-III electronic medical record data.}
The MIMIC-III database~\cite{journal/scidata2016/Johnson} is a large, publicly available ICU dataset comprising detailed electronic health records from Beth Israel Deaconess Medical Center (Boston, U.S.A.). We included patients whose ICU stays lasted between 30 and 60 hours to ensure sufficient temporal coverage for treatment-outcome modeling. A total of 25,186 patients met these criteria, comprising 56.3\% males and 43.7\% females, with a mean age of 62.9 years. The cohort included patients from 41 self-reported ethnic groups, with a mean ICU stay of 44.93 hours.
Among the included patients, vasopressor therapy was administered for an average of 7.74 $\pm$ 15.02 hours and mechanical ventilation for 10.39 $\pm$ 17.49 hours. All clinical variables were aggregated at hourly resolution. The complete list of vital signs, laboratory values, and treatment variables used in the model is provided in Table~\ref{tab:table1_all}.

\textbf{Out-of-distribution evaluation partitions.}
To assess cross-ethnicity generalization, we partitioned the MIMIC-III cohort by self-reported ethnicity: patients of European descent formed the training set, while Asian, African-descent, and Latino patients served as three independent out-of-distribution (OOD) test sets. These three groups were selected as the largest non-European subpopulations with sufficient sample sizes for robust evaluation; remaining ethnic groups were excluded due to small cohort sizes.
For cross-hospital evaluation, models trained on the MIMIC-III cohort were deployed on AmsterdamUMCdb without fine-tuning.

\textbf{Disease-category stratification.}
To evaluate disease-specific robustness, patients were stratified into four clinical categories based on primary diagnosis ICD-9 codes: cardiovascular and circulatory disorders (e.g., acute myocardial infarction, congestive heart failure, coronary artery disease), neurological disorders (e.g., stroke, intracranial haemorrhage, seizure), infectious and inflammatory diseases (e.g., pneumonia, sepsis, septic shock), and gastrointestinal, hepatobiliary, and metabolic disorders (e.g., gastrointestinal bleed, pancreatitis, diabetic ketoacidosis). The full mapping of ICD-9 codes to disease categories is provided in Supplementary Table~\ref{tab:disease_icd9_mapping}.

\textbf{Ventilator weaning sub-cohort.}
For the ventilator re-intubation prediction task, we identified a sub-cohort of 205 mechanically ventilated patients from the MIMIC-III dataset. Patients were selected based on ICD-9 codes for heart failure (428.x) and acute respiratory distress syndrome (ARDS; 518.82, 518.5), representing high-risk populations for extubation failure. Re-intubation was defined as the resumption of mechanical ventilation within six hours of extubation. Of the 205 patients, 67 (43\%) required re-intubation.

\textbf{Septic shock case study.}
The individual-level case study was selected from the MIMIC-III cohort based on ICD-9 code 785.52 (septic shock) to demonstrate the interpretability framework on a clinically representative scenario involving vasopressor therapy decisions.

\textbf{AmsterdamUMCdb electronic medical record data.}
The AmsterdamUMCdb database~\cite{journal/ccm2021/49Thoral} is a large, openly accessible intensive care dataset from two university medical centers in the Netherlands. We included adult patients whose ICU stays lasted between 30 and 60 hours. A total of 2,597 patients met these criteria, comprising 1,614 (62.2\%) males and 983 (37.8\%) females, with the largest age group being 70-79 years (25.6\%).
The mean ICU stay was 42.9 $\pm$ 7.1 hours. Among the included patients, vasopressor therapy was administered for an average of 12.91 $\pm$ 15.76 hours, and mechanical ventilation for 12.34 $\pm$ 15.40 hours. Baseline demographic and treatment characteristics are summarized in Appendix~\ref{tab:patient_cohort_amsterdam}.

\textbf{Synthetic patient cohort for controlled confounding evaluation.}
To enable controlled evaluation of counterfactual prediction, we simulated a synthetic patient cohort ($n = 10{,}000$) using a pharmacokinetic-pharmacodynamic (PKPD) tumor growth model~\cite{journal/scirep2017/7Geng}. This model simulates individualized treatment responses with known ground-truth counterfactual outcomes, allowing precise quantification of prediction accuracy under varying degrees of treatment selection bias~\cite{conference/iclr2020/Bica, conference/icml2022/Melnychuk, conference/icml2024/wang}. The synthetic cohort includes patients with diverse baseline tumor characteristics (volume and growth rate) and treatment scenarios spanning 30-day observation periods. Confounding strength ($\gamma$) was systematically varied from 0 (randomized treatment) to 7 (strong selection bias) for prediction experiments, and extended to $\gamma = 10$ for the representation balancing analysis following established protocol~\cite{conference/iclr2020/Bica}. Full simulation parameters, including treatment assignment mechanisms and validation protocols, are detailed in Appendix~\ref{subsec:details_synthetic_dataset}.

\subsection{Problem formulation and notations}
\label{subsec:problem_formulation}

The objective of individualized treatment outcome prediction is to estimate the potential evolution of a patient's physiological state under alternative treatment strategies, including counterfactual scenarios not observed in the historical data.

\textbf{Patient trajectories.}
For each patient $i$, we observe a longitudinal health trajectory spanning discrete time steps $t = 1, \ldots, T^i$. At each step $t$, let $\bm{X}_t^i \in \mathbb{R}^{d_x}$ denote the time-varying co-variates (e.g., vital signs and laboratory values), $\bm{A}_t^i \in \{a_1, \ldots, a_{d_a}\}$ denote the treatment administered, and $\bm{Y}_t^i \in \mathbb{R}^{d_y}$ denote the outcome of interest at the subsequent step. Static co-variates (e.g., age, gender, ethnicity, or comorbidities) are represented as $\bm{V}^i \in \mathbb{R}^{d_v}$. In the present study, treatments are binary ($d_a = 2$): presence or absence of vasopressor therapy, and presence or absence of mechanical ventilation. The observational dataset for $M$ patients is therefore
\begin{equation}
    \bm{H}_t = \{\{X^{i}_{t}, A^{i}_{t}, Y^{i}_{t}\}_{t=1}^{T^{i}} \cup \bm{V}^{i} \}_{1}^{M}. 
\end{equation}
\textbf{Patient history.}
Following prior work~\cite{journal/aim1997/757Rubin, journal/epidemiology2000/Robins,conference/iclr2020/Bica,conference/icml2022/Melnychuk,conference/icml2024/wang, conference/kdd2024/12Wu, conference/nips2024/37Bouchattaoui}, we define the history up to time $t$ as $\bar{\bm{H}}_t = \{\bar{\bm{X}}_t, \bar{\bm{A}}_{t-1}, \bar{\bm{Y}}_t, \bm{V}\}$,
where $\bar{\bm{X}}_t = (\bm{X}_1, \ldots, \bm{X}_t)$, 
$\bar{\bm{A}}_{t-1} = (\bm{A}_1, \ldots, \bm{A}_{t-1})$, and 
$\bar{\bm{Y}}_t = (\bm{Y}_1, \ldots, \bm{Y}_t)$.  
We condition on $\bar{\bm{A}}_{t-1}$ to ensure causal consistency, as the outcome at time $t$ is influenced by treatments administered before $t$.

\textbf{Representation learning objective.}
Instead of conditioning directly on high-dimensional raw trajectories, we employ a representation learning network $f_{\Theta_{\mathcal{B}}}(\cdot)$ to extract a compact, patient-specific latent state:
\begin{equation}
    \bm{\mathcal{B}}_t = f_{\Theta_{\mathcal{B}}}(\bm{H}_t), 
\end{equation}
where $\bm{\mathcal{B}}_t \in \mathbb{R}^{D}$ summaries the historical information available at time $t$.  
We denote the sequence of latent representations up to time $t$ as $\bar{\bm{\mathcal{B}}}_t = (\bm{\mathcal{B}}_1, \ldots, \bm{\mathcal{B}}_t).$
This representation trajectory serves as the input for estimating the expected counterfactual outcomes at future horizons.
$\bar{\bm{\mathcal{B}}}_t$ is then used to predict potential outcomes given assigned treatments $\bar{\bm{a}}_{t:t+\tau-1} = (a_t, \ldots, a_{t+\tau-1})$:
\begin{equation}
    \mathbb{E}[\hat{\bm{Y}}_{t+\tau}[\bar{\bm{A}}_{t:t+\tau-1}]|\bar{\bm{\mathcal{B}}}_t],
\end{equation}
where $\tau \geq 1$ denotes the prediction horizon, i.e., the number of future time steps ahead.
The key challenge is that treatment assignment in observational data is confounded: sicker patients may systematically receive more aggressive interventions. To mitigate this, we introduce a balancing regularizer into the representation learning objective. The total training loss comprises an outcome prediction term and a distribution alignment term:
\begin{equation}
    \mathcal{L} = \mathcal{L}_{\Theta_{\bm{Y}}} + \lambda \cdot \mathcal{L}_{\Theta_{\bm{\mathcal{B}}}},
\end{equation}
where $\mathcal{L}_{\Theta_{\bm{\mathcal{B}}}}$ encourages the learned representations $\bm{\mathcal{B}}_t$ to be distributionally similar across treatment groups. In GITO, we instantiate this term using a sampling-based Maximum Mean Discrepancy (sMMD) objective, described in detail in Section~\ref{subsec:gito_framework}.

\textbf{Causal assumptions.}
To establish the identifiability of treatment effects from observational data, we follow assumptions from previous studies~\cite{journal/aim1997/757Rubin, journal/epidemiology2000/Robins,books/crc2008/Robins, conference/icml2024/wang}, including consistency, sequential ignorability, and sequential overlap.

\textbf{Assumption 1: Consistency (aligning potential and observed outcomes).} At time step $t+1$, the observed outcome $\bm{Y}_{t+1}$ is assumed to be the potential outcome $\bm{Y}_{t+1}[a_t]$ that would have been realised under the assigned treatment $a_t$ at $t$, i.e., 
\begin{equation}
    \bm{Y}_{t+1} = \bm{Y}_{t+1}[a_t],
\end{equation}
this assumption ensures that the observed outcome aligns with the counterfactual outcome under the specific, well-defined treatment $a_t$. This requires the \textbf{Stable Unit Treatment Value Assumption (SUTVA)} to hold, specifically assuming no interference between subjects and a single, consistent version of the treatment.

\textbf{Assumption 2: Sequential overlap (positivity).}
For reliable estimation, we require that the probability of receiving any specific treatment $a_t$ is bounded away from zero for any patient history $\bar{h}_t$ that has a non-zero probability of occurrence.
\begin{equation}
    0 < P(\bm{A}_t = a_t \mid \bar{\bm{H}}_t = \bar{h}_t) < 1, \quad \text{if } P(\bar{\bm{H}}_t = \bar{h}_t) > 0, \quad \text{for all } a_t \in \bm{A}_t,
\end{equation}
this condition, often termed \textbf{positivity}, ensures that all treatment options remain possible given the observed clinical history.

\textbf{Assumption 3: Sequential ignorability (no unmeasured confounding).}
The treatment assigned at any time $t$ is assumed to be conditionally independent of the potential outcome, given the observed history. Formally, for all $a_t \in \bm{A}$,
\begin{equation}
    \bm{A}_t \perp \bm{Y}_{t+1}[a_t] \mid \bar{\bm{H}}_t,
\end{equation}
this is the \textbf{no unmeasured confounding (NUC)} assumption, which is critical for counterfactual outcome estimation. It implies that all variables influencing both $\bm{A}_t$ and $\bm{Y}_{t+1}$ have been adequately measured and included in $\bar{\bm{H}}_t$.
The key mathematical notation is summarized in Table~\ref{tab:notation}.

\begin{table}[htbp]
\centering
\begin{tabular}{l l}
\hline
\textbf{Symbol} & \textbf{Description} \\ \hline
    $i$ & Index for individual patients \\ 
    $t$ & Time step for each patient's health trajector \\ 
    $M$ & Total number of patients \\ 
    $T^{i}$ & Total number of time steps for patient \( i \)'s health trajectory \\ 
    $\bm{X}^{i} \in \mathbb{R}^{d_x}$ & Time-varying co-variates for patient \( i \) at time step \( t \) \\ 
    $\bm{A}^{i} \in \{a_1, ..., a_{d_a}\}$ & Types of treatments received by patient \( i \) at time step \( t \) \\ 
    $\bm{Y}^{i} \in \mathbb{R}^{d_y}$ & Treatment outcomes for patient \( i \) at time step \( t \)\\ 
    $\bm{V}^{i} \in \mathbb{R}^{d_v}$ & Static co-variates for patient \( i \) (e.g., age, gender, risky factors) \\ 
    $\bm{H}$ & Observational data for \( M \) patients \\ 
    $\bar{\bm{H}}_t$ & Trajectory information for patient \( i \) at time step \( t \) \\ 
    $\bar{\bm{X}}_t$ & Health trajectory co-variates up to time step \( t \), \( \bar{\bm{X}}_t = (\bm{X}_1, ..., \bm{X}_t) \) \\ 
    $\bar{\bm{A}}_{t-1}$ & Treatment history up to time step \( t-1 \), \( \bar{\bm{A}}_{t-1} = (\bm{A}_1, ..., \bm{A}_{t-1}) \) \\ 
    $\bar{\bm{Y}}_t$ & Outcome history up to time step \( t \), \( \bar{\bm{Y}}_t = (\bm{Y}_1, ..., \bm{Y}_t) \) \\ 
    $\hat{\bm{Y}}_{t+\tau}[\bar{\bm{A}}_{t:t+\tau}]$ & Estimated potential outcome at time step \( t+\tau \) \\
    $\mathbb{E}[\hat{\bm{Y}}_{t+\tau}[\bar{\bm{A}}_{t:t+\tau-1}] | \bar{\bm{H}}_t]$ & Expected potential outcome at time \( t+\tau \) \\ 
    $\bm{A}_t$ & Treatment at time step \( t \) \\ 
    $\bm{\mathcal{B}}_t$ & Learned balanced representations at time step \( t \) \\ 
    $\Theta_{\mathcal{B}}(\cdot)$ & Network for balanced representations learning \\ 
    $\Theta_{{Y}}(\cdot)$ & Network for treatment outcomes generation\\ \hline
\end{tabular}
\caption{Summary of mathematical notations used in the GITO framework.}
\label{tab:notation}
\end{table}

\begin{algorithm}[]
\caption{GITO training and inference procedure}
\textbf{Input:} Historical patient data $\bar{\bm{H}}_t = \{\bar{\bm{X}}_{t}, \bar{\bm{A}}_{t-1}, \bar{\bm{Y}}_{t}, \bm{V}\}$, Treatment assignments $\bm{A}_{t}$\;
\textbf{Output:} Predicted multi-step outcomes $\hat{\bm{Y}}_{t+1:t+\tau}$ under treatment sequence $\bm{A}_{t:t+\tau-1}$;

\vspace{0.5em}
\textbf{// Training Phase: One-step ahead prediction with balancing}\\
Initialize $\lambda \gets 0$ \Comment*[r]{Initial weight for balancing loss}

\For{$epoch = 1$ \KwTo $EPOCH$}{

    Compute progression factor: $\alpha_{epoch} = \frac{2}{1 + \exp\left( -10 \cdot \frac{epoch}{EPOCH} \right)} - 1$\;
    Update balancing weight: $\lambda \gets \alpha_{epoch}$ \Comment*[r]{Progressively increase $\lambda$ during training}

    Sample mini-batch $\mathcal{M}$ from training set\;
    \For{each time step $t$ in $\mathcal{M}$}{
        Encode input history $\bar{\bm{H}}_t = \{\bar{\bm{X}}_{t}, \bar{\bm{A}}_{t-1}, \bar{\bm{Y}}_{t}, \bm{V}\}$\;
        Learn representation $\bm{\mathcal{B}}_t = f_{\Theta_{\mathcal{B}}}(\bar{\bm{H}}_t)$\;
        Predict one-step outcome: $\hat{\bm{Y}}_{t+1} = g_{\Theta_Y}(\bm{\mathcal{B}}_t, \bm{A}_{t})$\;
    }

    Partition representations $\{\bm{\mathcal{B}}_t\}$ into treatment-specific subsets $\bm{\mathcal{D}}_k = \{ \bm{\mathcal{B}}_i \mid a_i = k \}$ for each treatment type $k \in \{1, \dots, d_a\}$\;
    \Comment*[r]{Sampling is done in representation space}

    Compute prediction loss using Equation~\ref{eq:prediction_loss}: $\mathcal{L}_{\Theta_Y} = \text{MSE}(\hat{\bm{Y}}_{t+1}, \bm{Y}_{t+1})$\;
    Compute aggregate sMMD loss via Eq.~\ref{equation:mmd}: $\mathcal{L}_{\Theta_{\mathcal{B}}} \leftarrow \sum_{1 \le i < j \le d_a} \text{MMD}^2_u(\bm{S}_i, \bm{S}_j)$\;
    Compute total loss: $\mathcal{L} = \mathcal{L}_{\Theta_Y} + \lambda \mathcal{L}_{\Theta_{\bm{B}}}$\;

    Update parameters $\Theta_{\mathcal{B}}, \Theta_Y$ via backpropagation\;
}

\vspace{0.5em}
\textbf{// Inference Phase: Multi-step prediction via expanding window}\\
Given initial history $\bar{\bm{H}}_t = \{\bar{\bm{X}}_{t}, \bar{\bm{A}}_{t-1}, \bar{\bm{Y}}_{t}, \bm{V}\}$ and treatment assignments $\bm{A}_{t:t+\tau-1}$:

Initialize cumulative contribution vector: $\bar{\bm{\omega}} \gets \bm{0}$\;

\For{$k = 1$ \KwTo $\tau$}{
    Encode history $\bar{\bm{H}}_{t+k-1}$ to get $\bm{\mathcal{B}}_{t+k-1} = f_{\Theta_{\mathcal{B}}}(\bar{\bm{H}}_{t+k-1})$\;
    Predict: $\hat{\bm{Y}}_{t+k} = f_{\Theta_Y}(\bm{\mathcal{B}}_{t+k-1}, \bm{A}_{t+k-1})$\;
    Compute variable contributions via Integrated Gradients: $\bm{\omega}_{t+k} = \text{IG}(\bm{H}_{t+k-1}, \bm{A}_{t+k-1})$\;
    Accumulate total contribution vector: $\bm{\omega}  \gets \bm{\omega} + \bm{\omega}_{t+k}$ \Comment*[r]{Each $\bm{\omega}$ is a vector over input variables}\;
    Update history: $\bm{H}_{t+k} \leftarrow \bm{H}_{t+k-1} \cup \{\hat{\bm{Y}}_{t+k}, \bm{A}_{t+k-1}\}$\;
}

Compute average contribution: $\bar{\bm{\omega}} \gets \bm{\omega} / \tau$\;
\label{alg:pseudocode}
\end{algorithm}

\subsection{GITO framework architecture}
\label{subsec:gito_framework}

The GITO framework integrates three synergistic components to achieve reliable counterfactual prediction: (1) a representation learning module (Encoder, $\Theta_{\mathcal{B}}$) that extracts patient state embeddings; (2) an outcome prediction network (Decoder, $\Theta_Y$) that generates trajectory forecasts under arbitrary treatment plans; and (3) a sampling-based distribution alignment mechanism (sMMD) that mitigates treatment-selection confounding. The overall architecture and training workflow are illustrated in Figure~\ref{fig:framework_and_user_interface}, with the procedural logic detailed in Algorithm~\ref{alg:pseudocode}.

\textbf{Representation Learning (Encoder).}
The encoder $\Theta_{\mathcal{B}}$ maps a patient's historical record $\bar{\bm{H}}_t$ to a compact latent representation. 
As defined in Section~\ref{subsec:problem_formulation}, the history at time $t$ comprises four components: time-varying co-variates $\bar{\bm{X}}_t$ (vital signs and 
laboratory values), previous outcomes $\bar{\bm{Y}}_t$, past treatments $\bar{\bm{A}}_{t-1}$, and static co-variates $\bm{V}$ (demographics). At each time step, the temporal inputs $[\bm{X}_t \oplus \bm{Y}_t \oplus \bm{V}]$ are 
concatenated and processed jointly with the treatment history:
\begin{equation}
\bm{\mathcal{B}}_t = f_{\Theta_{\mathcal{B}}}(
  \bar{\bm{X}}_t \oplus \bar{\bm{Y}}_t \oplus \bm{V}, \;
  \bar{\bm{A}}_{t-1}),
\end{equation}
where $\oplus$ denotes feature-level concatenation. The resulting latent state $\bm{\mathcal{B}}_t \in \mathbb{R}^D$ serves as an informationally sufficient summary of the patient's physiological history up to time $t$. Our framework is 
architecture-agnostic: the encoder can leverage various sequential 
modeling backbones, such as Transformers~\cite{conference/icml2022/Melnychuk}, 
LSTMs~\cite{conference/iclr2020/Bica}, or 1D-CNNs~\cite{conference/icml2024/wang}. We empirically validate this portability across all three architectures in Section~\ref{sec:results}.

\textbf{Outcome Prediction (Decoder) and Counterfactual Inference.}
The prediction network $\Theta_Y$ takes the learned representation $\bm{\mathcal{B}}_t$ and a candidate treatment action $\bm{a}_t$ (encoded as a one-hot vector over $d_a$ possible treatments) to forecast the next outcome:
\begin{equation}
\hat{\bm{Y}}_{t+1} = g_{\Theta_Y}(\bm{\mathcal{B}}_t, \bm{a}_t).
\end{equation}
For multi-step prediction over a horizon $\tau$, the model operates autoregressively: each predicted outcome $\hat{\bm{Y}}_{t+k}$ is fed back to update the latent state, generating a continuous trajectory:
\begin{equation}
\hat{\bm{Y}}_{t+k+1} = g_{\Theta_Y}\!\bigl(
    f_{\Theta_{\mathcal{B}}}(\bar{\bm{H}}_t, \hat{\bm{Y}}_{t+1:t+k}),\;
    \bm{a}_{t+k}
\bigr), \quad k = 1, \dots, \tau - 1.
\label{eq:multistep}
\end{equation}
During training, we employ teacher forcing~\cite{journal/neco1989/1Williams}: 
the ground-truth outcomes $\bm{Y}_{t+1:t+k}$ are supplied as inputs 
at each recursive step. At inference, teacher forcing is switched off 
and the model autoregressively consumes its own predictions 
$\hat{\bm{Y}}_{t+1:t+k}$, enabling multi-step trajectory generation 
without access to future observations. This autoregressive mechanism 
also enables counterfactual trajectory generation: by substituting alternative treatment sequences $\{\bm{a}'_{t}, \bm{a}'_{t+1}, \dots\}$ into Eq.~\eqref{eq:multistep}, clinicians can explore hypothetical physiological responses to different treatment plans from the same patient state $\bm{\mathcal{B}}_t$.

\textbf{Bias Mitigation via sMMD.} To ensure that the learned representations $\bm{\mathcal{B}}_t$ capture true physiological states rather than treatment-assignment artifacts, we regularize the encoder with a sampling-based Maximum Mean Discrepancy (sMMD) loss that minimizes the distributional distance between treatment groups in the latent space. The formulation and computational details of this balancing objective are presented in Section~\ref{subsec:training_objective_and _ptimization}.

\subsection{Training objective and optimization}
\label{subsec:training_objective_and _ptimization}

\textbf{Factual Prediction Loss.}
To ensure the model accurately captures physiological dynamics, we minimize the mean squared error (MSE) between predicted and observed outcomes. For a batch of training samples, the prediction loss is:
\begin{equation}
\mathcal{L}_{\Theta_Y} = \frac{1}{M'} \sum_{i, t} ||\bm{Y}_{t+1}^i - g_{\Theta_{Y}}(\bm{\mathcal{B}}_t^i, \bm{A}_t^i)||^2,
\label{eq:prediction_loss}
\end{equation}
where $g_{\Theta_{Y}}$ denotes the outcome prediction network and $M'$ is the total number of transition tuples in the batch.

\textbf{Balancing loss.} A central challenge in counterfactual outcome estimation is learning latent representations that are predictive of outcomes yet invariant to treatment assignment. While adversarial training (e.g., GAN-based discriminators) can enforce such invariance, it frequently suffers from optimization instability (min–max gaming) and mode collapse. To overcome these limitations, we adopt a discrepancy-based regularization strategy using Maximum Mean Discrepancy (MMD)~\cite{conference/nips2006/Gretton}. Unlike adversarial discriminators, MMD provides a closed-form, kernel-based distance metric that directly penalizes distributional mismatch.

\textbf{Balancing Loss via sMMD.}
A central challenge in counterfactual outcome estimation is learning latent representations that are predictive of outcomes yet invariant to treatment assignment. While adversarial training (e.g., GAN-based discriminators) can enforce such invariance, it frequently suffers from optimization instability and mode collapse~\cite{conference/icml2017/214arjovsky}. To overcome these limitations, we adopt Maximum Mean Discrepancy (MMD)~\cite{conference/nips2006/Gretton}, a kernel-based distributional distance that provides a stable, closed-form regularization signal without min-max optimization.

For a set of $d_a$ possible treatments, we minimize the average pairwise discrepancy across all treatment groups:
\begin{equation}
\mathcal{L}_{\Theta{\mathcal{B}}} = \frac{1}{\binom{d_a}{2}} \sum_{1 \le i < j \le d_a} \text{MMD}^2(\bm{\mathcal{D}}_i, \bm{\mathcal{D}}_j),
\end{equation}
where $\bm{\mathcal{D}}_k = \{ \bm{\mathcal{B}}_t \mid \bm{A}_t = k \}$ denotes the set of latent representations associated with treatment $k$.

Computing MMD over the entire dataset scales quadratically ($O(N^2)$) and is therefore computationally prohibitive. We adopt a sampling-based approximation (sMMD), formally instantiated as an unbiased U-statistic estimator. At each iteration, fixed-size random subsets $\bm{S}_i \subset \mathcal{D}_i$ and $\bm{S}_j \subset \mathcal{D}_j$ (with $|\bm{S}|=N_s$) are drawn to compute:
\begin{equation}
\begin{aligned}
\text{MMD}^2_u(\bm{S}_i, \bm{S}_j) &= \frac{1}{N_s(N_s-1)} \sum_{x \in \bm{S}_i} \sum_{x' \in \bm{S}_i, x' \neq x} k(x, x') + \frac{1}{N_s(N_s-1)} \sum_{y \in \bm{S}_j} \sum_{y' \in \bm{S}_j, y' \neq y} k(y, y') \\
&- \frac{2}{N_s^2} \sum_{x \in \bm{S}_i} \sum_{y \in \bm{S}_j} k(x, y).
\end{aligned}
\label{equation:mmd}
\end{equation}
The expected value of this estimator over random subsets is identical to the population squared MMD, ensuring an unbiased gradient signal. We set $N_s = 200$, which provides a stable variance-computation trade-off; an ablation over $N_s \in \{50, 100, 200, 500\}$ showed negligible performance variation ($<$\,0.5\% RMSE), confirming the estimator's robustness to this choice.

We employ a Radial Basis Function (RBF) kernel $k(\bm{x},\bm{y})=\exp(-\|\bm{x}-\bm{y}\|^2 / 2\sigma^2)$. Because the RBF kernel is characteristic, $\text{MMD}=0$ uniquely implies distributional equivalence. The bandwidth $\sigma$ is dynamically determined at each step using the median heuristic, setting $\sigma$ to the square root of the median pairwise distance within the pooled subsets $\bm{S}_i \cup \bm{S}_j$, yielding a data-adaptive kernel scale that adjusts to the optimization trajectory without manual tuning. We optimize $\text{MMD}^2$ rather than MMD because $\text{MMD}^2$ is directly proportional to the variance in kernel embeddings, avoids numerical instability of the square root near zero, and thus ensures smoother gradient-based alignment. The procedure is summarized in Algorithm~\ref{alg:mmd_loss}

\textbf{Joint Optimization.}
The entire framework is trained end-to-end by jointly optimizing the factual prediction error and the distribution balancing loss:
\begin{equation}
(\hat{\Theta}_Y, \hat{\Theta}_{\mathcal{B}}) = \arg \min_{\Theta_Y, \Theta_{\mathcal{B}}}
\mathcal{L}_{\Theta_Y}(\Theta_Y, \Theta_{\mathcal{B}}) + \lambda \mathcal{L}_{\Theta_{\mathcal{B}}}(\Theta_{\mathcal{B}}),
\label{eq:objective_function}
\end{equation}
where $\lambda$ governs the trade-off between predictive accuracy and confounding removal.
\textbf{Sigmoidal Annealing Schedule.}
Applying strong balancing regularization during early training can suppress physiologically meaningful variability before the encoder has learned useful representations. To ensure stable convergence, we adopt a smooth sigmoidal schedule for $\lambda$. At training epoch $e$:
\begin{equation}
    \lambda_e = \frac{2}{1 + \exp\!\bigl(-10 \cdot \tfrac{e}{E}\bigr)} - 1,
\label{eq:lambda_schedule}
\end{equation}
where $E$ denotes the total number of epochs. This schedule initializes $\lambda$ near zero, allowing the encoder to first learn physiologically relevant features, then smoothly increases toward full regularization during mid-to-late training. This progressive strategy yielded consistently stable training behavior across all three backbone architectures evaluated (Appendix~\ref{appendix:optimization_properties_and_convergence_discussion}).

\begin{algorithm}
\caption{Computation of Sampling-based MMD Loss}
\label{alg:mmd_loss}
\SetAlgoLined
\KwIn{Batch representations $\bm{\mathcal{B}} \in 
\mathbb{R}^{N \times D}$, treatment labels 
$\bm{A} \in \{1, \dots, d_a\}^N$, sample size $N_s$, 
kernel function $k(\cdot,\cdot)$}
\KwOut{Balancing loss $\mathcal{L}_{\Theta_\mathcal{B}}$}

\tcp{Group representations by treatment}
Partition $\bm{\mathcal{B}}$ into subsets 
$\{\bm{\mathcal{D}}_1, \dots, \bm{\mathcal{D}}_{d_a}\}$ 
where 
$\mathcal{D}_k = \{\bm{b} \in \bm{\mathcal{B}} 
\mid a = k\}$\;
$\mathcal{L}_{\Theta_\mathcal{B}} \gets 0$\;

\ForEach{treatment pair $(i, j)$ with 
$1 \leq i < j \leq d_a$}{
  \If{$|\bm{\mathcal{D}}_i| < N_s$ \textbf{or} 
  $|\bm{\mathcal{D}}_j| < N_s$}{
    \textbf{continue}\tcp*[r]{Skip if insufficient}
  }
  Sample $\bm{S}_i \sim \mathcal{D}_i$, 
  $\bm{S}_j \sim \mathcal{D}_j$, each of size $N_s$\;

  \tcp{Unbiased kernel estimates (Eq.~\ref{equation:mmd})}
  $\hat{\mu}_{ii} \gets 
  \frac{1}{N_s(N_s{-}1)}\!\sum_{p \neq q} 
  k(\bm{s}_p^i, \bm{s}_q^i)$\;
  $\hat{\mu}_{jj} \gets 
  \frac{1}{N_s(N_s{-}1)}\!\sum_{p \neq q} 
  k(\bm{s}_p^j, \bm{s}_q^j)$\;
  $\hat{\mu}_{ij} \gets 
  \frac{1}{N_s^2}\sum_{p,q} 
  k(\bm{s}_p^i, \bm{s}_q^j)$\;

  $\widehat{\mathrm{MMD}}^2_{ij} \gets 
  \hat{\mu}_{ii} + \hat{\mu}_{jj} 
  - 2\hat{\mu}_{ij}$\;
  $\mathcal{L}_{\Theta_\mathcal{B}} \gets 
  \mathcal{L}_{\Theta_\mathcal{B}} 
  + \widehat{\mathrm{MMD}}^2_{ij}$\;
}

$\mathcal{L}_{\Theta_\mathcal{B}} \gets 
\mathcal{L}_{\Theta_\mathcal{B}} 
\big/ \binom{d_a}{2}$
\tcp*[r]{Average over all pairs}

\Return $\mathcal{L}_{\Theta_\mathcal{B}}$\;
\end{algorithm}

\subsection{Downstream clinical classifier for ventilator re-intubation}
To evaluate the clinical utility of GITO-generated counterfactuals, we developed a downstream predictive model tasked with assessing the risk of re-intubation within six hours post-extubation.
The classifier takes as input a concatenated multivariate time series consisting of $T_{\text{hist}}=12$ hours of observed history and $T_{\text{pred}}=6$ hours of predicted trajectories from GITO.

\textbf{Model Architecture.}
We adopted a 1D Residual Convolutional Neural Network (ResNet-1D) architecture~\cite{journal/artmed2021/117Jia}, which is effective at capturing local temporal dependencies in physiological signals. The backbone comprised three residual blocks with progressively increasing channel dimensions ($[64, 128, 256]$), each containing two 1D convolutional layers (kernel size $k = 3$, stride $= 1$, same padding) with batch normalization and ReLU activation, followed by an identity shortcut connection. The output of the final residual block was passed through global average pooling and a fully connected layer ($256 \to 1$) with sigmoid activation.
The dataset was split at the patient level into training (70\%), validation (15\%), and test (15\%) subsets. The model parameters were optimized using the AdamW algorithm (learning rate $\eta = 10^{-3}$, weight decay $\lambda = 10^{-4}$). Given the inherent class imbalance in ventilator weaning outcomes (where re-intubation events are the minority), we employed two complementary loss functions to ensure robust sensitivity:

\textbf{Weighted Binary Cross-Entropy (BCE):} To penalize false negatives more heavily, we applied class weights inversely proportional to class frequencies ($w_{+} = N / (2 N_{+}),\; w_{-} = N / (2 N_{-})$, yielding $w_{+} \approx 1.53,\; w_{-} \approx 0.74$ for the 43\% re-intubation prevalence):
\begin{equation}\mathcal{L}_{\mathrm{BCE}} = -\frac{1}{N} \sum_{i=1}^{N}\Big[ w_{+} y_i \log(\hat{p}_i) + w_{-} (1-y_i)\log(1-\hat{p}_i) \Big],
\end{equation}
where $y_i \in \{0,1\}$ is the ground-truth label, $\hat{p}_i = \sigma(\hat{z}_i)$ is the predicted probability.

\textbf{Focal Loss:} In sensitivity analyses, we further addressed the ``easy-negative'' dominance problem using Focal Loss, which dynamically down-weights well-classified examples to focus training on hard samples:
\begin{equation}\mathcal{L}_{\mathrm{focal}} = -\frac{1}{N} \sum_{i=1}^{N}\alpha (1-p_{t,i})^{\gamma} \log(p_{t,i}),
\end{equation}
where $p_{t,i} = \hat{p}_i$ if $y_i=1$ and $p_{t,i}=1-\hat{p}_i$ otherwise. We set the balancing factor $\alpha=0.25$ and the focusing parameter $\gamma=2.0$ following standard practices for dense object detection and rare event prediction.

\subsection{Human-AI comparison and collaboration study}
\label{subsec:human_ai_comparison}
\textbf{Study design and participants.}
To evaluate GITO's impact on clinical decision-making, we employed a two-period, two-treatment crossover design involving nine healthcare professionals stratified by clinical experience: three attending physicians (senior clinicians), three residents (junior clinicians), and three medical students. All participants provided written informed consent.
Participants predicted the risk of re-intubation within six hours post-extubation for the full cohort of 205 MIMIC-III ventilator weaning cases (see Section~\ref{subsec:study_design}). The prediction task was binary: whether the patient would require re-intubation (label $= 1$) or not (label $= 0$).

\textbf{Crossover procedure.}
Participants were randomly assigned to one of two sequences to control for learning and order effects. Group~1 ($n = 6$) first made predictions without AI assistance (control period), then with GITO assistance (treatment period). Group~2 ($n = 3$) followed the reverse order.
In all experimental phases, clinicians were provided with the patient's baseline clinical information, including demographic data (age, gender), admission diagnosis, and key vital signs from the preceding 12~hours (features consistent with those defined in Table~\ref{tab:table1_all}).
In the GITO assistance phase (the treatment period), this baseline information was augmented. Clinicians additionally received GITO's output: a quantitative re-intubation risk prediction and a corresponding attribution-based interpretable explanation generated from the patient's time-series data. Participants were required to review this AI output before finalizing their clinical decision, enabling the assessment of the effect of human-AI collaboration on accuracy and process. \textbf{Outcome measures.} Three primary endpoints were evaluated:
\begin{enumerate}
    \item \textbf{Prediction accuracy}: the proportion of cases in which the clinician's prediction matched the ground-truth re-intubation outcome (Top-1 accuracy), analyzed using a linear mixed-effects model to correct for the crossover design and control for clinical experience tier.
    \item \textbf{Decision-making time}: the elapsed time from case presentation to final prediction submission, recorded per batch of 205 cases.
    \item \textbf{Clinically acceptable safety rate} ($1 - \text{FNR}$): In this prediction task, the two types of error carry asymmetric clinical consequences. A false negative, predicting label $= 0$ when the patient actually requires re-intubation, represents the most dangerous error, as it may lead to premature extubation and subsequent respiratory failure requiring emergency re-intubation. Conversely, a false positive,predicting label $= 1$ when the patient would not require re-intubation, results in a conservative decision to delay extubation, an outcome that, while suboptimal, does not pose an immediate safety risk to the patient. The clinically acceptable safety rate was therefore defined as $1 - \text{FNR}$ (equivalently, recall for the positive class), reflecting the proportion of true high-risk cases correctly identified.
\end{enumerate}
Accuracy and safety rate were compared between control and treatment periods using within-subject paired comparisons. The crossover design enabled each participant to serve as their own control, reducing inter-individual variability.

\textbf{Foundation model benchmarking.}
To establish a rigorous AI baseline, we evaluated four state-of-the-art general-purpose large language models,GPT-4o, GPT-5.1, Gemini-3, and Grok-4.1,against GITO on the same 205-patient cohort.
To ensure these models operated at their peak potential, we did not rely on zero-shot inference. Instead, we implemented a Structured Clinical Reasoning Pipeline that encoded expert ICU knowledge into the system prompt.
The prompt instructed the LLMs to follow a three-step reasoning process:
\begin{enumerate}
    \item \textbf{Feature extraction and rule application:}
    The models first evaluated key physiological indicators against standard weaning thresholds derived from clinical literature. Specific criteria included:
    \begin{itemize}
        \item \textbf{Respiratory mechanics and gas exchange:}
        Rapid Shallow Breathing Index (RSBI) $< 105$~breaths/min/L~\cite{journal/nejm1991/21yang, journal/atm2016/11Karthika};
        Tidal Volume $> 5$~mL/kg~\cite{journal/nejm1995/332Estebon, tobin2006principles};
        Respiratory Rate $8 \leq \text{RR} \leq 30$~\cite{journal/erj2007/29Boles};
        PaO$_2$/FiO$_2$ ratio $\geq 200$~mmHg~\cite{journal/nejm1995/332Estebon, journal/erj2007/29Boles};
        PaCO$_2 < 50$~mmHg or within baseline range~\cite{journal/erj2007/29Boles}.
        \item \textbf{Haemodynamic stability:}
        Mean Arterial Pressure (MAP) $\geq 65$~mmHg~\cite{journal/erj2007/29Boles, journal/ccm2021/49Evans};
        Heart Rate $60 \leq \text{HR} \leq 140$~beats/min~\cite{journal/erj2007/29Boles}.
        \item \textbf{Acid-base and metabolic status:}
        pH $7.35$-$7.45$~\cite{journal/erj2007/29Boles, tobin2006principles};
        Lactate $< 2$~mmol/L~\cite{journal/ccm2015/43Thille};
        Bicarbonate (HCO$_3^-$) $22$-$30$~mEq/L;
        Potassium (K$^+$) $3.5$-$5.0$~mEq/L~\cite{journal/ec2018/7Kardalas, journal/statpearls2025/Castro};
        Sodium (Na$^+$) $135$-$145$~mEq/L~\cite{journal/afp2015/5Braun, journal/statpearls2025/Castro}.
        \item \textbf{Renal function:}
        BUN/Creatinine evaluated for acute deterioration; renal failure-associated fluid overload can precipitate pulmonary oedema~\cite{journal/ccm2015/43Thille}.
    \end{itemize}
    \item \textbf{Composite scoring:} Based on these checks, the models synthesized a ``Spontaneous Breathing Trial (SBT) Likelihood'' (High/Moderate/Low).
    \item \textbf{Probabilistic prediction:} Finally, models predicted the probability of re-intubation, which was binarized using a decision threshold of 0.5.
\end{enumerate}
Each model received the same patient data in an identical prompt format. All predictions were deterministic (temperature $= 0$) to ensure reproducibility. This prompt design ensures that any observed performance gap reflects an inherent limitation of LLMs in processing temporal physiological dynamics, rather than a lack of domain knowledge.
The medical students in the comparison analysis ($n = 3$) were the same three students enrolled in the crossover study; their unassisted predictions from the control period served as the human baseline.

\subsection{Evaluation metrics}
\textbf{General protocol.}
For multi-step treatment outcome prediction experiments on MIMIC-III, AmsterdamUMCdb, and the synthetic tumor growth dataset, all models were trained and evaluated over $n = 10$ independent runs with fixed random seeds (seeds 10, 101, 1010, 10101, 101010, 50, 505, 5050, 50505, 505050) controlling data splitting, model initialization, and sMMD sampling; results are reported as mean $\pm$ standard deviation. For downstream clinical tasks (e.g., ventilator re-intubation prediction), $n = 5$ independent runs were used unless otherwise stated. Statistical significance between paired model comparisons was assessed using two-sided paired $t$-tests, with significance thresholds indicated by $^{*}$ ($p < 0.05$) and $^{**}$ ($p < 0.01$) in the corresponding tables.

\textbf{Model comparison (regression).}
For multi-step treatment outcome prediction experiments, we evaluated numerical accuracy using the root mean squared error (RMSE), a standard regression metric that quantifies the discrepancy between predicted and observed outcomes:
\begin{equation}
\mathrm{RMSE} = \sqrt{ \frac{1}{M} \sum_{i=1}^M (\hat{\bm{Y}}_i - \bm{Y}_i)^2 }.
\label{metrics:rmse}
\end{equation}
\textbf{Ventilator re-intubation prediction (classification).}
For the binary re-intubation prediction task, model performance was evaluated on an independent test cohort using accuracy, precision, recall, F1-score, and area under the receiver operating characteristic curve (AUROC) as primary metrics:
\begin{align}
\mathrm{Accuracy} &= \frac{\mathrm{TP} + \mathrm{TN}}{\mathrm{TP} + \mathrm{TN} + \mathrm{FP} + \mathrm{FN}}, \\
\mathrm{Precision} &= \frac{\mathrm{TP}}{\mathrm{TP} + \mathrm{FP}}, \\
\mathrm{Recall} &= \frac{\mathrm{TP}}{\mathrm{TP} + \mathrm{FN}}, \\
\mathrm{F1} &= 2 \cdot \frac{\mathrm{Precision} \cdot \mathrm{Recall}}{\mathrm{Precision} + \mathrm{Recall}},
\end{align}
where $\mathrm{TP}$, $\mathrm{TN}$, $\mathrm{FP}$, and $\mathrm{FN}$ denote true positive, true negative, false positive, and false negative counts, respectively.
The AUROC was computed by integrating the receiver operating characteristic (ROC) curve over all classification thresholds:
\begin{equation}
\mathrm{AUROC} = \int_{0}^{1} \mathrm{TPR}(\mathrm{FPR}) \, d(\mathrm{FPR}),
\end{equation}
where $\mathrm{TPR}$ and $\mathrm{FPR}$ represent the true positive and false positive rates, respectively.
95\% confidence intervals for all classification metrics were estimated using bootstrap resampling (10,000 iterations) on the test cohort.

\textbf{Per-variable information preservation ($\Delta R^{2}$).}
To quantify how each balancing strategy affects the retention of individual clinical variables in the learned representations, we computed per-variable $\Delta R^{2}$:
\begin{equation}
\Delta R^{2}_j = R^{2}_{j,\text{unbalanced}} - R^{2}_{j,\text{balanced}},
\end{equation}
where $R^{2}_{j,\text{unbalanced}}$ and $R^{2}_{j,\text{balanced}}$ denote the coefficient of determination for reconstructing variable $j$ from the unbalanced and balanced representations, respectively. A positive $\Delta R^{2}$ indicates that balancing incurred additional information loss beyond baseline compression; values near zero indicate no added cost; negative values indicate that balancing improved reconstruction relative to the unbalanced encoder. Error bars denote 95\% confidence intervals over $n = 10$ runs.

\textbf{Reconstruction loss.}
To evaluate how much patient-specific information each balancing objective preserves, an independent decoder network was trained to reconstruct the original patient co-variates from the balanced representations. The reconstruction objective was mean squared error (MSE):
\begin{equation}
\mathcal{L}_{\text{recon}} = \frac{1}{M \cdot d_x} \sum_{i=1}^{M} \| \hat{\bm{X}}_i - \bm{X}_i \|^2,
\end{equation}
where $\hat{\bm{X}}_i$ is the reconstructed covariate vector and $\bm{X}_i$ is the original. The decoder was trained with the encoder weights frozen, ensuring that reconstruction quality reflects only the information content of the balanced representations, not the decoder's capacity. Both training and validation reconstruction losses are reported.

\textbf{Evaluation of representation quality.}
To validate the effectiveness of sMMD in mitigating confounding bias, we analyzed the structure of the learned latent representations using t-Distributed Stochastic Neighbor Embedding (t-SNE)~\cite{journal/jmlr2008/11Van} with perplexity $= 30$ and 1,000 iterations. High-dimensional patient embeddings were projected into a two-dimensional manifold to assess two properties:
\begin{enumerate}
    \item \textbf{Treatment invariance}: whether the distributions of treated and control groups are indistinguishable (well-mixed) in the latent space, rather than forming treatment-specific clusters.
    \item \textbf{Preservation of prognostic structure}: whether the embeddings retain clinically meaningful heterogeneity (e.g., age-related physiological variation) while removing spurious demographic correlations (e.g., gender, ethnicity).
\end{enumerate}

\subsection{Interpretability and explainability pipeline}

\textbf{Gradient-based feature attribution.}
To quantify the contribution of individual physiological variables to the model's predictions, we employed Integrated Gradients (IG)~\cite{conference/icml2017/70Mukund}, a widely adopted axiomatic attribution method. Given an input sequence $\bar{\bm{X}}_{t}$, we defined the baseline reference $\bar{\bm{X}}^0_{t}$ as the cohort mean for each variable.
This yields a raw attribution score $\phi^{(j)}_{i}$ for variable $i$ corresponding to prediction step $j$.
To summarize importance across the entire prediction window $\tau$, we averaged the contributions:
\begin{equation}
\omega_{i}^{\text{raw}} = \frac{1}{\tau} \sum_{j=1}^{\tau} \phi^{(j)}_{i}.
\label{eq:avg_ig}
\end{equation}
For comparative analysis, these raw scores were normalized using a softmax function to produce a relative importance distribution:
\begin{equation}
\omega_{i} = \frac{\exp(\omega_{i}^{\text{raw}})}{\sum_{k=1}^{d_x} \exp(\omega_{k}^{\text{raw}})},
\label{eq:softmax_ig}
\end{equation}
where $d_x$ is the number of input variables. This normalization highlights the dominant physiological signals driving the forecast. In addition to the aggregated scores $\omega_i$, the per-step attributions $\phi^{(j)}_{i}$ are visualized individually to reveal how each variable's contribution evolves over the prediction horizon (see Figure~\ref{fig:case_study}, upper right).

\textbf{Counterfactual trajectory generation.}
To enable clinicians to compare alternative treatment strategies, GITO generates multi-step predicted trajectories under each candidate treatment plan. Given the learned representation $\bar{\bm{\mathcal{B}}}_t$ at the current time step, the model autoregressively rolls out future predictions by conditioning on a specified treatment sequence $\bar{\bm{a}}_{t:t+\tau-1}$. At each roll-out step, the predicted outcome $\hat{\bm{Y}}_{t+j}$ is fed back as input for the next step. In the present study, four scenarios were evaluated: no treatment, vasopressor only, ventilation only, and both treatments simultaneously. The resulting trajectory set provides a comparative view of expected physiological evolution under each strategy.

\textbf{LLM-driven interpretability and multimodal reasoning.}
To bridge the gap between quantitative risk scores and clinical reasoning, we developed a structured multi-modal prompting pipeline that synthesizes predictions into interpretable narratives. The pipeline employs a large language model (LLM), GPT-4o (version \texttt{gpt-4o-2024-08-06}) by default, with temperature set to 0 for deterministic output and a maximum token limit of 4,096, though the platform supports user-selectable alternatives, acting under a strict ``critical care physician'' persona to produce a three-tiered clinical summary.

To mitigate the risk of generative hallucination, we implemented a two-stage Chain-of-Thought (CoT) framework that hybridizes explicit data extraction with scenario-based reasoning:

\begin{itemize}
    \item \textbf{Stage~I: Visual grounding and extraction.}
    Unlike standard ``black-box'' generation, the pipeline first enforces a grounding step. The model is supplied with pre-computed statistics (current value, moving average, linear trend) of the top-$k$ ($k = 5$) contributing features identified by Integrated Gradients. It is guided to cross-reference these structured inputs with the encoded visual charts (vital signs trend and patient history) to validate physiological states (e.g., verifying whether MAP is trending below the 65~mmHg threshold) before narrative construction begins.

    \item \textbf{Stage~II: Structured narrative synthesis.}
    Leveraging the grounded data, the model generates a structured explanation following a rigorous protocol:
    (1)~\textit{Primary metric analysis}: assessment of the target outcome's trajectory relative to historical interventions;
    (2)~\textit{Holistic vital status}: integration of secondary vital sign trends;
    (3)~\textit{Comparative scenario reasoning}: a disciplined comparison of the counterfactual prediction trajectories (None, Vaso, Vent, Both).
    The prompt enforces a ``quantification discipline,'' requiring the model to cite specific approximate deltas when comparing scenarios and prohibiting the inference of superiority when differences are clinically negligible ($< 2\%$ probability delta).
\end{itemize}

The LLM is additionally instructed to output a structured JSON response that includes, for each treatment scenario, a numerical preference score (as a percentage) reflecting the estimated clinical suitability based on trajectory analysis and grounded vital sign assessment. The preference scores across all scenarios are constrained to sum to 100\%, providing an interpretable ranking of treatment options. An abbreviated example of the generated output is shown in Box~\ref{LLMreasoning}; the full prompt and output schema are provided in Appendix~\ref{appendix:llm-full-output}.

% ============================================================
% Methods — Implementation Details and Experimental Setup
% ============================================================

\subsection{Implementation details and experimental setup}

\textbf{Data preprocessing.}
Given the irregular sampling frequency inherent in ICU electronic health records, handling missing data is critical. We applied a Last Observation Carried Forward (LOCF) strategy to impute missing values in time-varying co-variates, followed by Next Observation Carried Backward (NOCB) for any remaining initial gaps. This approach preserves the temporal continuity of physiological states.
To facilitate stable model convergence, continuous co-variates (both static and temporal) were standardized using Z-score normalization:
\begin{equation}
x'_{t,i} = \frac{x_{t,i} - \mu_i}{\sigma_i},
\end{equation}
where $\mu_i$ and $\sigma_i$ represent the global mean and standard deviation of feature $i$ calculated across the entire training corpus.

\textbf{Baseline comparisons.}
We benchmarked GITO against state-of-the-art treatment outcome estimation models:
\begin{itemize}
\item \textbf{CRN (Counterfactual Recurrent Network)}~\cite{conference/iclr2020/Bica}: Uses LSTMs with domain adversarial training to build balanced representations.
\item \textbf{CT (Causal Transformer)}~\cite{conference/icml2022/Melnychuk}: A Transformer-based architecture that uses distinct attention heads for processing treatment and covariate history.
\item \textbf{ACTIN (Adversarial Counterfactual Temporal Inference Network)}~\cite{conference/icml2024/wang}: The backbone architecture of our proposed method, which originally uses a GAN-based discriminator for balancing. We used ACTIN as the primary baseline to isolate the specific contribution of our sMMD module.
\end{itemize}
To construct the sMMD-enhanced variants (CRN-sMMD, CT-sMMD, and ACTIN-sMMD), we replaced each model's original adversarial balancing mechanism with our proposed sMMD loss while keeping all other architectural components and hyperparameters unchanged. This controlled substitution isolates the effect of the balancing strategy from other architectural differences.

\textbf{Computational environment.}
All models were implemented in Python~3.10 using the PyTorch~2.1 deep learning framework. Model training was performed on the NUS Vanda high-performance computing cluster equipped with $2\times$ NVIDIA Tesla A40 GPUs (48\,GB VRAM each) and $2\times$ 36-core Intel Xeon 8452Y CPUs. Inference latency benchmarks (Table~\ref{tab:computational_cost}) were measured on CPU only (Intel Xeon 8452Y) without GPU acceleration, to reflect deployment conditions in resource-constrained hospital environments.

\textbf{Training and evaluation protocol.}
To ensure rigorous evaluation and prevent data leakage, the dataset was randomly partitioned at the patient level into training (70\%), validation (15\%), and test (15\%) sets. For the out-of-distribution (OOD) evaluation on MIMIC-III, the training and validation sets comprised exclusively patients of European descent; the remaining non-European subpopulations (Asian, African-descent, and Latino) served as independent OOD test sets.

Model parameters were optimized using the Adam optimizer. Training was terminated via early stopping with a patience of 10 epochs based on validation-set RMSE; the checkpoint with the lowest validation loss was selected for evaluation. For all baselines and multi-step-ahead prediction, teacher forcing was used during training~\cite{journal/neco1989/1Williams}. During evaluation of multi-step-ahead prediction, teacher forcing was switched off and models autoregressively fed their own predictions.

Key hyperparameters for the primary MIMIC-III experiments are summarized in Table~\ref{tab:hyperparameters_main}; full per-dataset configurations are provided in Supplementary Tables~\ref{tab:crn_parameters}-\ref{tab:actin_parameters}.

\begin{table}[h]
\centering
\caption{Key hyperparameters for GITO and baselines on the MIMIC-III dataset.}
\label{tab:hyperparameters_main}
\begin{tabular}{lccc}
\toprule
\textbf{Parameter} & \textbf{CRN-sMMD} & \textbf{CT-sMMD} & \textbf{ACTIN-sMMD} \\
\midrule
Learning rate & 0.001 & 0.0005 & 0.001 \\
Batch size & 256 & 64 & 128 \\
Max epochs & 100 & 150 & 400 \\
Hidden dimension (encoder) & 72 & 48 & 24 \\
FC hidden units (predictor) & 72 & 48 & 48 \\
Dropout rate & 0.1 & 0.3 & , \\
Early stopping patience & \multicolumn{3}{c}{10 epochs} \\
sMMD sample size $N_s$ & \multicolumn{3}{c}{200} \\
RBF kernel bandwidth $\sigma$ & \multicolumn{3}{c}{Median heuristic (adaptive)} \\
Balancing weight $\lambda$ & \multicolumn{3}{c}{Sigmoidal schedule (Eq.~\ref{eq:lambda_schedule})} \\
optimizer & \multicolumn{3}{c}{Adam ($\beta_1 = 0.9$, $\beta_2 = 0.999$)} \\
\bottomrule
\end{tabular}
\end{table}

\backmatter

\section*{Data availability}
The MIMIC-III dataset is publicly available from PhysioNet (https:// mimic.physionet.org/). AmsterdamUMCdb is publicly available from the Amsterdam Medical Data Science website (https://amsterdammedicaldatascience.
nl/).

\section*{Code availability}
The implementation of GITO, together with preprocessing scripts and trained models, 
is available at https://github.com/peisong-zhang/COEOT.  

\section*{Supplementary information}

Supplementary information is available in the online version of the paper. It includes additional figures, tables, methods, and source data supporting the findings of this study.

% \bmhead{Acknowledgements}

% Acknowledgements are not compulsory. Where included they should be brief. Grant or contribution numbers may be acknowledged.

% Please refer to Journal-level guidance for any specific requirements.

% \section*{Declarations}

% Some journals require declarations to be submitted in a standardized format. Please check the Instructions for Authors of the journal to which you are submitting to see if you need to complete this section. If yes, your manuscript must contain the following sections under the heading `Declarations':

% \begin{itemize}
% \item Funding
% \item Conflict of interest/Competing interests (check journal-specific guidelines for which heading to use)
% \item Ethics approval and consent to participate
% \item Consent for publication
% \item Data availability 
% \item Materials availability
% \item Code availability 
% \item Author contribution
% \end{itemize}

% \noindent

\begin{appendices}
% \appendix

\section{Related works}
\label{sec:related_works}

Early methodologies for counterfactual estimation focus on static data; existing methods mainly fall into the following categories: propensity score-based approaches, covariate adjustment techniques, matching algorithms, and outcome modeling methods. 

Propensity score-based methods, such as propensity score matching and inverse probability weighting, estimate the probability of receiving treatment conditional on co-variates~\cite{journal/jasa1984/516Rosenbaum, journal/biometrika1983/41Rosenbaum, journal/aim1997/757Rubin}. These methods aim to balance the covariate distributions between treated and untreated groups to reduce confounding bias. However, they are sensitive to model misspecification and cannot address hidden confounding factors. Furthermore, matching approaches may reduce sample size and statistical power because some units cannot be matched.
Another class of methods involves directly adjusting for co-variates via regression models, such as linear regression and generalized linear models~\cite{journal/pads2008/1202Austin}. While these models are easy to implement and interpret, their performance relies heavily on correct model specification. 
Matching methods non-parametrically pair treated units with control units of similar covariate distributions. These approaches are attractive due to their simplicity~\cite{journal/tkdd2021/1Yao}. However, in high-dimensional settings, achieving good matches becomes increasingly difficult, leading to potential imbalance and loss of information due to discarded units.
Moreover, all these approaches may be limited in real-world applications, particularly in healthcare where patient conditions and treatment effects often evolve over time, and decisions are made sequentially based on time-varying information.

To address these limitations, numerous methods have been developed for estimating treatment effects in time-varying settings. Unlike static approaches, these methods explicitly account for temporal dependencies, where treatments, co-variates, and outcomes change over time. 
These methods, such as marginal structural models (MSMs)~\cite{journal/mathmodel1986/7Robins,journal/epidemiology2000/Robins,journal/jasa2001/454Hernán,books/crc2008/Robins}, attempt to mitigate confounding bias by reweighting or stratifying data based on estimated treatment probabilities. 
However, these methods rely on strong assumptions about model specification and often struggle to capture complex temporal dependencies inherent in longitudinal data.
To address this limitation, more recent research integrates sequential models like recurrent neural networks (RNNs) with causal inference techniques such as inverse probability weighting (IPW)~\cite{conference/nips2018/31Lim,conference/kdd2024/12Wu} or G-computation~\cite{conference/mlh2021/282Li, conference/mlh2024/252Hong} to better account for time-varying treatment effects. These methods extend traditional balancing strategies by leveraging sequential models to capture temporal patterns in observational data. Among these, state-of-the-art machine learning methods such as the counterfactual recurrent network (CRN)~\cite{conference/iclr2020/Bica}, causal transformer (CT)~\cite{conference/icml2022/Melnychuk}, and adversarial counterfactual temporal inference network (ACTIN)~\cite{conference/icml2024/wang} employ adversarial training to enhance balancing between treatment groups to mitigate bias. Specifically, the core idea of adversarial domain adaptation is to train a discriminator to distinguish between treatment groups while the feature extractor learns representations that make this discrimination difficult. This forces the learned representations to be treatment-invariant, effectively reducing the influence of confounding variables in treatment assignment.
These adversarial-based balancing strategies provide a flexible and data-driven approach to balancing, avoiding the need for explicit functional form assumptions required by traditional causal inference methods. 

Despite their ability to remove associations between patient history and treatment assignments, they are highly sensitive to distribution shifts~\cite{conference/nips2022/Moayeri}, meaning they may fail when applied to out-of-distribution scenarios. 
In addition, these methods often struggle with the trade-off between balancing and covariate information preservation. The aggressive removal of treatment-related signals can inadvertently lead to information loss, particularly under severe confounding bias, reducing the accuracy of the estimation~\cite{conference/icml2024/Huang}. This highlights the need for alternative approaches that achieve robust balancing while preserving effective information for counterfactual outcome estimation. 
To mitigate this issue, recent research proposed a covariance de-correlation-based mechanism to achieve a better trade-off between bias reduction and prediction accuracy~\cite{conference/iclr2025/wang}. However, this method is designed specifically for state-space models (SSMs), which do not generalize well to other settings. 
Moreover, none of these methods have investigated potential distribution shifts in real-world healthcare, limiting their practical applicability.

\section{Dataset}
\label{sec:datset}

This section provides supplementary cohort-level statistics for the datasets used in this study. Table~\ref{tab:patient_cohort_amsterdam} summaries the demographic and clinical characteristics of the AmsterdamUMCdb cohort, which served as a geographically independent validation set. Table~\ref{tab:disease_icd9_mapping} lists the mapping of primary-diagnosis ICD-9 codes to the four disease categories used for disease-specific subgroup analysis on MIMIC-III.
% , AmsterdamUMCdb cohort table (unchanged) ,
\begin{table}[htbp]
\centering
\caption{Demographic and clinical characteristics of the patient cohort with intensive care unit (ICU) stays between 30 and 60 hours. This cohort retains the diversity of the broader population while ensuring tractable training and analysis.}
\label{tab:patient_cohort_amsterdam}
\begin{tabular}{p{7cm}c}
\toprule
\textbf{Characteristic} & \textbf{All Patients (N=2597)} \\
\midrule
\textbf{Age group} & \\
\hspace{1em}18-39, n (\%) & 344 (13.2\%) \\
\hspace{1em}40-49, n (\%) & 218 (8.4\%) \\
\hspace{1em}50-59, n (\%) & 417 (16.1\%) \\
\hspace{1em}60-69, n (\%) & 616 (23.7\%) \\
\hspace{1em}70-79, n (\%) & 665 (25.6\%) \\
\hspace{1em}80+, n (\%) & 337 (13.0\%) \\
\addlinespace[0.3em]
\textbf{Gender} & \\
\hspace{1em}Male, n (\%) & 1614 (62.2\%) \\
\hspace{1em}Female, n (\%) & 983 (37.8\%) \\
\addlinespace[0.3em]
\textbf{Vitals}$^{\textit{a}}$ & \\
\hspace{1em}Heart rate (bpm), mean (SD) & 83.38 (15.11) \\
\hspace{1em}Red blood cells (M/$\mu$L), mean (SD) & 1.10 (0.63) \\
\hspace{1em}Sodium (mmol/L), mean (SD) & 128.48 (4.22) \\
\hspace{1em}Mean blood pressure (mmHg), mean (SD) & 76.48 (10.81) \\
\hspace{1em}Glucose (mmol/L), mean (SD) & 7.72 (38.63) \\
\hspace{1em}Chloride urine (mmol/L), mean (SD) & 0.46 (47.67) \\
\hspace{1em}Hematocrit (\%), mean (SD) & 0.31 (5.09) \\
\hspace{1em}PEEP (cmH$_2$O), mean (SD) & 4.32 (2.21) \\
\hspace{1em}Respiratory rate (bpm), mean (SD) & 17.02 (3.96) \\
\hspace{1em}Prothrombin time (s), mean (SD) & 0.65 (4.94) \\
\hspace{1em}Cholesterol (mmol/L), mean (SD) & 0.14 (47.35) \\
\hspace{1em}Hemoglobin (g/dL), mean (SD) & 6.85 (1.82) \\
\hspace{1em}Creatinine ($\mu$mol/L), mean (SD) & 107.45 (1.31) \\
\hspace{1em}Blood urea nitrogen (mmol/L), mean (SD) & 7.16 (19.47) \\
\hspace{1em}Bicarbonate (mmol/L), mean (SD) & 21.35 (3.98) \\
\hspace{1em}Calcium ionized (mmol/L), mean (SD) & 0.90 (0.57) \\
\hspace{1em}Partial pressure of CO$_2$ (mmHg), mean (SD) & 36.90 (8.78) \\
\hspace{1em}Magnesium (mmol/L), mean (SD) & 0.82 (0.32) \\
\hspace{1em}Anion gap (mEq/L), mean (SD) & 6.43 (3.16) \\
\hspace{1em}Phosphorous (mg/dL), mean (SD) & 1.00 (1.12) \\
\hspace{1em}Platelets (K/$\mu$L), mean (SD) & 195.40 (104.81) \\
\hspace{1em}Calcium urine (mmol/L), mean (SD) & 5.11 (8.88) \\
\hspace{1em}Diastolic blood pressure (mmHg), mean (SD) & 58.30 (10.41) \\
\addlinespace[0.3em]
\textbf{Treatments}$^{\textit{b}}$ & \\
\hspace{1em}Vasopressor (h), mean (SD) & 12.91 (15.76) \\
\hspace{1em}Mechanical ventilation (h), mean (SD) & 12.34 (15.40) \\
\bottomrule
\end{tabular}
\vspace{0.5cm}
\footnotesize
Abbreviations: SD, standard deviation; SVR, systemic vascular resistance; PEEP, positive end-expiratory pressure. (\textbf{\textit{a}}) For time-varying vital signs, mean values were computed over the first 24 hours following ICU admission. (\textbf{\textit{b}}) Treatment durations reflect the average number of hours continuous or intermittent interventions were administered, averaged across all patients.
\end{table}

% , Disease category ICD-9 mapping (NEW) ,
\begin{table}[htbp]
\centering
\caption{Mapping of primary diagnosis ICD-9 codes to disease categories used for disease-specific subgroup analysis.}
\label{tab:disease_icd9_mapping}
\begin{tabular}{p{0.9\textwidth}}
\toprule
\textbf{Cardiovascular \& circulatory disorders} \\
\midrule
Ventricular tachycardia, complete heart block, acute coronary syndrome, atrial fibrillation, bradycardia, cardiac arrest, aortic dissection, aortic stenosis, congestive heart failure, coronary artery disease, unstable angina, NSTEMI, STEMI, acute myocardial infarction, CABG, catheterisation, AVR, MVR, MAZE, stent \\
\addlinespace[0.5em]
\toprule
\textbf{Neurological disorders} \\
\midrule
Stroke, TIA, intracranial haemorrhage, subdural haematoma, subarachnoid haemorrhage, seizure, status epilepticus, head bleed, intraparenchymal haemorrhage, epidural haematoma, meningitis, altered mental status, carotid stenosis, spinal cord injury, CVA, brain tumor, Chiari malformation \\
\addlinespace[0.5em]
\toprule
\textbf{Infectious \& inflammatory diseases} \\
\midrule
Pneumonia, sepsis, septic shock, urosepsis, cellulitis, bacteraemia, endocarditis, cholangitis, pyelonephritis, meningitis, fever of unknown origin, neutropenic fever, hypoxia \\
\addlinespace[0.5em]
\toprule
\textbf{Gastrointestinal, hepatobiliary \& metabolic disorders} \\
\midrule
Gastrointestinal bleed (upper/lower), pancreatitis, cholangitis, cholecystitis, liver failure, hepatic encephalopathy, cirrhosis, diverticulitis, abdominal pain, diabetic ketoacidosis, ulcerative colitis, colon cancer, oesophageal cancer \\
\bottomrule
\end{tabular}
\end{table}

\subsection{Details on synthetic tumor growth dataset}
\label{subsec:details_synthetic_dataset}

The Tumor Growth (TG) simulator \cite{journal/scirep2017/7Geng} models the tumor volume $
\hat{\bm{Y}}_{t+1}$ at $t+1$ days post-diagnosis, where the outcome is one-dimensional (i.e., $d_y=1$). The model incorporates two binary treatment variables: (i) radiotherapy ($\bm{A}_{t}^{(r)}$) and (ii) chemotherapy ($\bm{A}_{t}^{(c)}$). Specifically, radiotherapy induces an immediate effect $d(t)$ on the subsequent outcome, whereas chemotherapy exerts an influence over multiple future time points through an exponentially decaying effect $C(t)$. And they are modeled as following equation:
\begin{equation}
    \bm{Y}_{t+1} = (1 + \rho\log\frac{K}{\bm{Y}_t}-\beta_{c}C_{t}-(\alpha_{r}d_{t} + \beta_{t}d_{t}^{2}) + \epsilon_{t})\bm{Y}_t,
\end{equation}
where $\rho, K, \beta_c, \alpha_r, \beta_r$ are parameters in the simulation and and where $\epsilon_t \sim N(0, 0.01^2)$ is the sampled noise. The parameters $\beta_c, \alpha_r, \beta_r$ characterize individual patient responses and are drawn from a mixture of truncated normal distributions with three components. For exact parameter values, refer to the code implementation. The mixture component indices are treated as static co-variates ($d_v=1$). Time-varying confounding is introduced through a biased treatment assignment, which remains identical for both treatment groups; i.e., 
\begin{equation}
    \bm{A}_{t}^{c}, \bm{A}_{t}^{r} \sim Bernoulli(\sigma(\frac{\gamma}{D_{max}}(\bar{D}_{15}\bar({\bm{Y}}_{t-1})-\frac{D_{max}}{2}))),
\end{equation}
where $\sigma$ is a sigmoid activation with an output between [0,1] as the probability parameter of the Bernoulli distribution, $D_{max}$ is the maximum tumor diameter, $\bar{D}_{15}\bar({\bm{Y}}_{t-1})$ is the average tumor diameter over the last 15 days, and $\gamma$ is a confounding parameter, controlling the "biasing effect" of tumor size on treatment assignment. The larger $\gamma$ is, the stronger the bias is. This is a mechanism that introduces confounding dynamically based on tumor growth, simulating a real-world scenario where physicians may adjust treatment strategies according to the tumor size.

\subsection{Experiment details on MIMIC-III dataset}
\label{subsec:details_mimiciii_dataset}
We utilized the MIMIC-extract dataset~\cite{conference/chil2020/222Wang}, which applies a standardized preprocessing pipeline to the MIMIC-III dataset~\cite{journal/scidata2016/Johnson}. MIMIC-extract offers intensive care unit (ICU) data aggregated on an hourly basis. To handle missing values, both forward and backward filling are employed, followed by standard normalization of all continuous time-varying features.
Our analysis includes 29 vital sign indicators, such as heart rate, respiratory rate, diastolic blood pressure, glucose, blood urea nitrogen, and 19 others. In addition, we consider 3 static attributes (e.g., age, gender, and ethnicity). Categorical features are represented using one-hot encoding. These variables, comprising both dynamic co-variates and invariant characteristics, are considered potential confounders. We examine two binary treatments: vasopressor administration and mechanical ventilation. The primary outcome of interest is (diastolic) blood pressure, which may either increase or decrease in response to these treatments. This variation is crucial for clinicians when assessing the anticipated progression of patient trajectories under such interventions.
From the full MIMIC-III cohort of 25,186 eligible patients (see Methods: Patient Cohort), each experiment randomly sampled 5,000 individuals who were admitted to the ICU for at least 30 hours, with a maximum stay capped at 60 hours. The dataset was split into training, validation, and testing sets in a 70\%/15\%/15\% ratio. The study's methodology was adapted based on the forecast horizon $\tau$. Specifically:
\begin{enumerate}
    \item For one-step-ahead predictions, the full test set trajectories were used.
    
    \item For multi-step prediction ($\tau \geq$ 2) the process involved defining $\tau_{max}\geq\tau$ as the longest projection horizon. Sub-trajectories of at least $\tau_{max}+1$ steps were then extracted using a rolling origin approach, while initial vital sign readings up to $\tau^{(i)}-\tau_{max} + 1$ were removed to eliminate any foresight bias in the prediction process.
\end{enumerate}

To evaluate the generalization of the model under distribution shifts, we designed two out-of-distribution (OOD) settings based on both patient demographics and admission diagnoses.

\textbf{Ethnicity-based OOD setting.} In the first OOD setting, we trained the model exclusively on White patients and evaluated its performance on non-White subpopulations, treating each ethnicity group as an independent OOD test set.
To ensure statistical reliability and sufficient sample size, we selected Asian (N=119), Black (N=383), and Hispanic (N=143) patients, as these groups are the most represented among non-White patients in the dataset. This setup allows us to assess the model's robustness across various ethnicity subpopulations, identifying potential biases in treatment outcome prediction.

\begin{small}   % 缩小字号
\setlength{\tabcolsep}{4pt}  % 缩小列间距
\begin{longtable}{p{3.8cm}cccc}

\caption{Demographic and clinical characteristics of the patient 
cohort in ethnicity-based distribution shift settings.}
\label{tab:table_ethnicity_ood} \\

\toprule
\textbf{Characteristic} & \textbf{White} & \textbf{Asian} 
& \textbf{Black} & \textbf{Hispanic} \\
& (N=3{,}560) & (N=119) & (N=383) & (N=143) \\
\midrule
\endfirsthead

% ,- 续表表头 ,-
\multicolumn{5}{l}{\small\itshape 
Table~\ref{tab:table_ethnicity_ood} continued} \\[4pt]
\toprule
\textbf{Characteristic} & \textbf{White} & \textbf{Asian} 
& \textbf{Black} & \textbf{Hispanic} \\
& (N=3{,}560) & (N=119) & (N=383) & (N=143) \\
\midrule
\endhead

% ,- 续表脚注 ,-
\midrule
\multicolumn{5}{r}{\small\itshape Continued on next page} \\
\endfoot

% ,- 最终脚注 ,-
\bottomrule
\multicolumn{5}{p{\linewidth}}{\footnotesize
Abbreviations: SD, standard deviation; SVR, systemic vascular 
resistance; GCS, Glasgow Coma Scale; PEEP, positive end-expiratory 
pressure. 
(\textbf{\textit{a}})~For time-varying vital signs, mean values 
were computed over the first 24\,h following ICU admission. 
(\textbf{\textit{b}})~Treatment durations reflect the average 
number of hours of continuous or intermittent interventions, 
averaged across all patients.} \\
\endlastfoot

% =============== 表格内容 ===============
\textbf{Age} & & & & \\
\quad Age ($\leq$89), mean (SD) 
  & 64.16 (16.32) & 59.18 (19.19) & 58.30 (17.93) & 53.77 (17.38) \\
\quad Age $>$89, n (\%) 
  & 209 (5.4\%) & 9 (7.56\%) & 18 (4.70\%) & 2 (1.40\%) \\
\addlinespace[3pt]

\textbf{Gender} & & & & \\
\quad Male, n (\%) 
  & 2{,}001 (56.2\%) & 62 (52.1\%) & 180 (47.4\%) & 93 (64.5\%) \\
\quad Female, n (\%) 
  & 1{,}559 (33.8\%) & 57 (47.9\%) & 203 (52.6\%) & 50 (35.5\%) \\
\addlinespace[3pt]

\textbf{Vitals}$^{\textit{a}}$ & & & & \\
\quad Heart rate (bpm) 
  & 84.76 (15.23) & 84.85 (16.11) & 87.31 (16.47) & 89.08 (16.27) \\
\quad Red blood cells (M/$\mu$L) 
  & 3.64 (0.60) & 3.65 (0.69) & 3.74 (0.69) & 3.80 (0.65) \\
\quad Sodium (mEq/L) 
  & 138.46 (4.29) & 139.26 (4.66) & 138.66 (4.88) & 139.13 (3.96) \\
\quad Mean BP (mmHg) 
  & 77.77 (10.45) & 79.47 (11.04) & 82.21 (11.24) & 82.92 (12.23) \\
\quad SVR (dyn$\cdot$s/cm$^5$) 
  & 1{,}499.76 (697.57) & 1{,}571.20 (694.78) & 1{,}575.48 (696.18) & 1{,}666.86 (656.79) \\
\quad Glucose (mg/dL) 
  & 137.09 (37.48) & 137.29 (34.15) & 145.40 (47.91) & 139.41 (43.12) \\
\quad Chloride urine (mEq/L) 
  & 67.14 (48.43) & 64.44 (50.21) & 65.22 (47.60) & 72.71 (49.99) \\
\quad GCS score 
  & 13.58 (2.56) & 13.40 (2.78) & 13.88 (2.22) & 13.77 (2.47) \\
\quad Hematocrit (\%) 
  & 32.43 (5.06) & 31.97 (5.21) & 32.47 (5.58) & 33.24 (5.77) \\
\quad PEEP (cmH$_2$O) 
  & 5.15 (2.21) & 4.79 (1.45) & 5.08 (2.33) & 4.95 (1.94) \\
\quad Respiratory rate (bpm) 
  & 18.50 (3.90) & 17.81 (3.87) & 19.39 (4.38) & 18.28 (4.38) \\
\quad Prothrombin time (sec) 
  & 15.18 (5.12) & 14.16 (2.64) & 15.00 (4.02) & 14.75 (3.90) \\
\quad Cholesterol (mg/dL) 
  & 162.94 (48.35) & 166.28 (48.51) & 160.52 (47.05) & 161.60 (47.20) \\
\quad Hemoglobin (g/dL) 
  & 11.04 (1.81) & 10.77 (1.80) & 10.85 (1.94) & 11.37 (2.02) \\
\quad Creatinine (mg/dL) 
  & 1.25 (1.13) & 1.28 (1.27) & 1.83 (2.17) & 1.29 (1.42) \\
\quad BUN (mg/dL) 
  & 23.30 (18.90) & 23.83 (18.49) & 26.56 (23.00) & 21.03 (16.46) \\
\quad Bicarbonate (mEq/L) 
  & 24.04 (4.04) & 23.55 (3.66) & 23.80 (4.30) & 23.54 (3.49) \\
\quad Calcium ionized (mmol/L) 
  & 1.56 (7.69) & 1.15 (0.17) & 1.68 (10.05) & 2.85 (13.03) \\
\quad pCO$_2$ (mmHg) 
  & 40.94 (8.75) & 40.19 (8.10) & 41.51 (9.96) & 39.41 (7.34) \\
\quad Magnesium (mg/dL) 
  & 2.01 (0.34) & 2.09 (0.55) & 1.99 (0.34) & 1.97 (0.29) \\
\quad Anion gap (mEq/L) 
  & 13.75 (3.13) & 13.64 (2.86) & 14.44 (3.42) & 13.84 (3.10) \\
\quad Phosphorous (mg/dL) 
  & 3.49 (1.12) & 3.41 (0.97) & 3.61 (1.26) & 3.60 (1.00) \\
\quad Venous PvO$_2$ (mmHg) 
  & 50.69 (13.38) & 52.61 (14.21) & 50.91 (13.36) & 52.11 (13.34) \\
\quad Platelets (K/$\mu$L) 
  & 217.89 (104.52) & 202.96 (102.38) & 228.30 (100.73) & 215.04 (89.34) \\
\quad Calcium urine (mg/dL) 
  & 5.52 (9.81) & 6.56 (11.59) & 5.16 (8.93) & 5.19 (9.86) \\
\quad Diastolic BP (mmHg) 
  & 60.02 (10.13) & 61.77 (9.86) & 65.14 (11.05) & 65.83 (11.21) \\
\addlinespace[3pt]

\textbf{Treatments}$^{\textit{b}}$ & & & & \\
\quad Vasopressor (h) 
  & 4.15 (0.38) & 3.97 (0.37) & 2.38 (0.30) & 2.19 (0.91) \\
\quad Ventilation (h) 
  & 5.52 (0.42) & 6.12 (0.44) & 5.12 (0.41) & 6.29 (2.62) \\

\end{longtable}
\end{small}

\textbf{Diagnosis-based OOD setting.} The second OOD setting introduces an additional level of domain shift by selecting specific broad disease categories from the OOD test set. We focus on the following major disease groups to ensure a sufficient sample size: cardiovascular diseases, neurological disorders, and infectious and inflammatory diseases. This selection strategy avoids the issue of data scarcity that would arise from choosing a single specific disease. By evaluating the model on these distinct diagnostic subgroups, we aim to investigate whether domain shifts in underlying medical conditions further impact model performance beyond demographic shifts alone.
The details of this OOD setting are presented in Table~\ref{tab:ethnicity_disease}.
\begin{table*}[htbp]
    \centering
    \resizebox{\textwidth}{!}{
    \begin{tabular}{lcccc}
        \toprule
        \textbf{Ethnicity} & \textbf{Cardiovascular diseases} & \textbf{Neurological disorders} & \textbf{Infectious and inflammatory diseases} &\textbf{Total} \\
        \midrule
        Asian             & 15 & 13 & 15 & 43 \\
        Black             & 43 & 57 & 57 & 157\\
        Hispanic          & 17 & 27 & 13 & 57\\
        Total             & 75 & 87 & 85 &  257\\
        \bottomrule
    \end{tabular}
    }
    \caption{Details of the OOD test set in the diagnosis-based OOD setting.}
    \label{tab:ethnicity_disease}
\end{table*}

\subsection{Experiment details on human-AI comparison and collaboration}

\label{subsec:details_ai_human_comparison}
This section provides supplementary details for the human-AI comparison and collaboration study described in Methods (Section~\ref{subsec:human_ai_comparison}).
\textbf{Patient cohort and task.}
The evaluation cohort comprised 205 ventilator weaning cases from MIMIC-III (see Section~\ref{subsec:study_design}). Patients were selected based on ICD-9 diagnosis codes associated with conditions frequently requiring mechanical ventilation: heart failure (428.x: 428, 4280, 4281, 42820-42843, 4289), and acute respiratory distress syndrome (518.82, 518.5). The prediction task was binary: whether the patient would require re-intubation within six hours of extubation (label $= 1$) or not (label $= 0$).

\begin{figure*}[t]
  \centering
  \includegraphics[width=\linewidth]{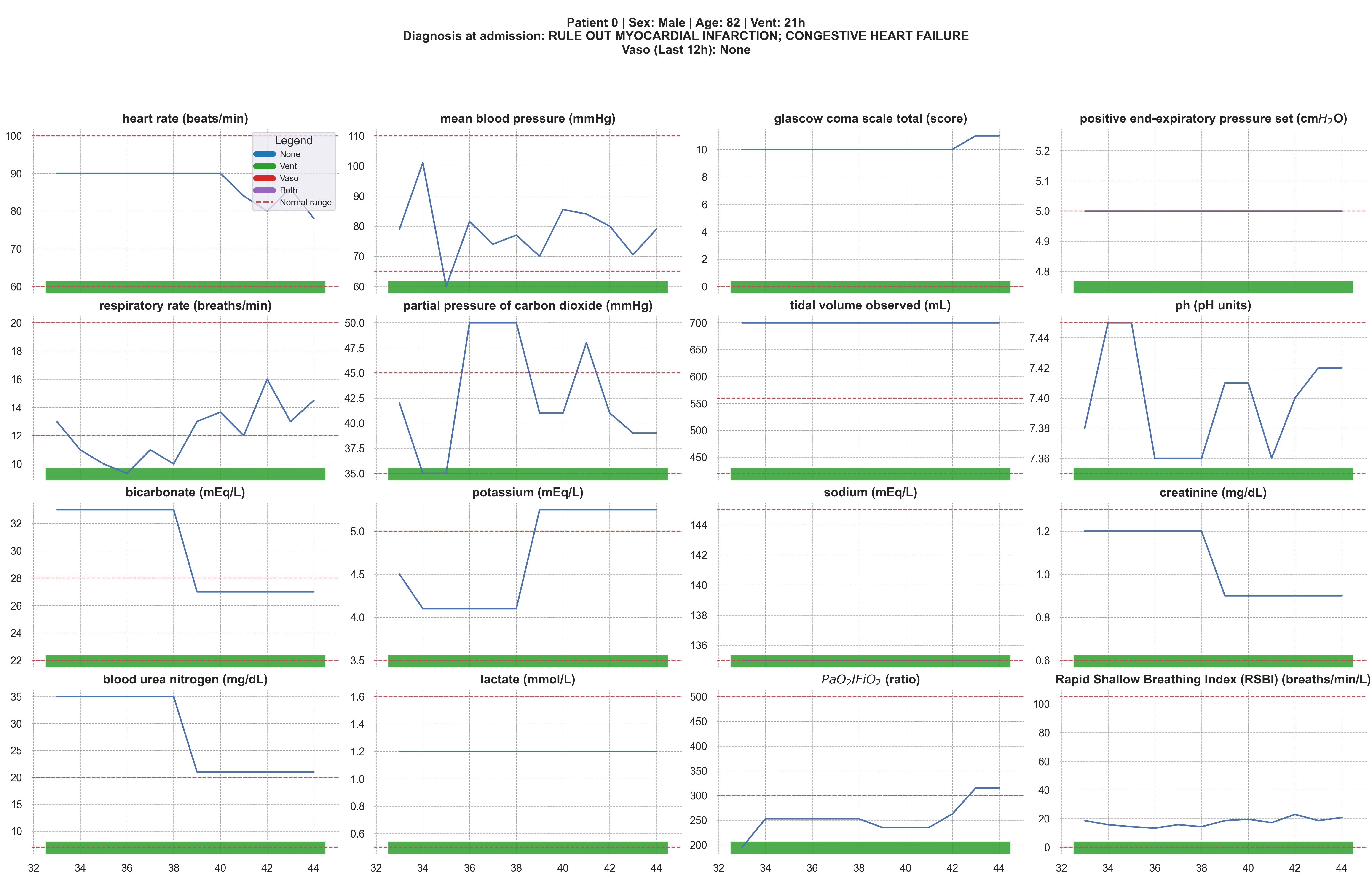}
  \caption{\textbf{Clinical dashboard - unassisted condition.}
  Clinicians received patient demographics (age, sex), primary 
  diagnosis at admission, cumulative duration of mechanical 
  ventilation, vasopressor administration history with dosage, 
  and 12-hour temporal trajectories of 16 physiological parameters. 
  See Figure~\ref{fig:patient_template_w_pred} for the 
  GITO-assisted condition.}
  \label{fig:patient_template_wo_pred}
\end{figure*}

\begin{figure*}[t]
  \centering
  \includegraphics[width=\linewidth]{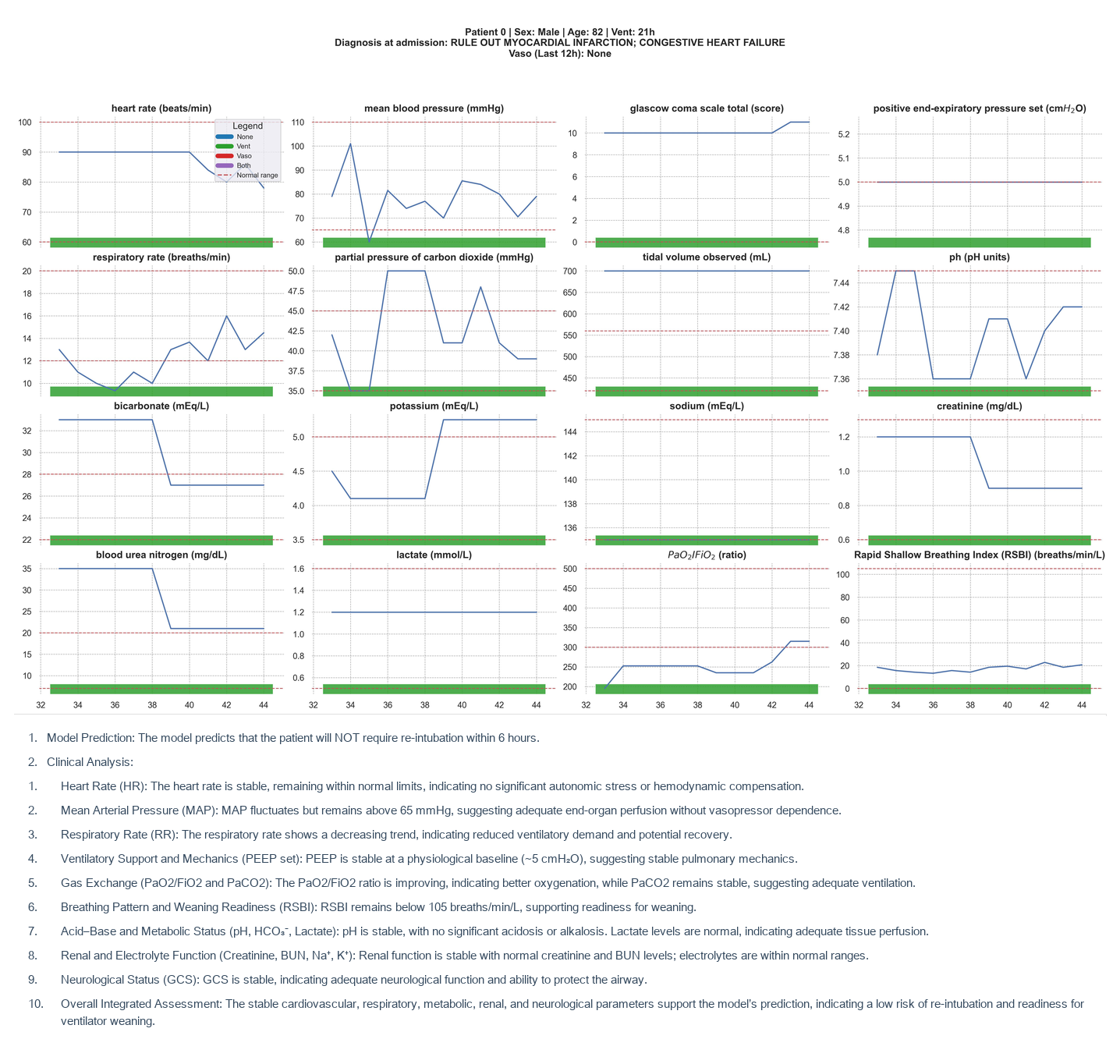}
  \caption{\textbf{Clinical dashboard - GITO-assisted condition.}
  In addition to the information in 
  Figure~\ref{fig:patient_template_wo_pred}, clinicians received 
  GITO's 6-hour predicted trajectories (orange curves) appended 
  to the observed history, together with a quantitative 
  re-intubation risk score and an attribution-based interpretable 
  explanation. Both dashboards were presented to $n = 4$ medical 
  students and $n = 3$ clinicians during the two-period crossover 
  experiment in counterbalanced order 
  (see Methods~\ref{subsec:human_ai_comparison}).}
  \label{fig:patient_template_w_pred}
\end{figure*}

\textbf{Data presentation to clinicians.}
For human participants, each patient's clinical information was presented via a structured dashboard. Vital signs were arranged by their average values and visualized in groups of five time-series panels, with the final panel displaying the prediction target and assigned treatment history. Demographic data (age, gender), admission diagnosis, and key vital signs from the preceding 12 hours were provided in all conditions. In the GITO-assisted condition, the dashboard was augmented with GITO's quantitative risk prediction and an attribution-based interpretable explanation. Figure~\ref{fig:patient_template_wo_pred} and Figure~\ref{fig:patient_template_w_pred}) present an example of the clinical visualization interface used in both the unassisted and GITO-assisted conditions.

\textbf{LLM prompt design.}
Four large language models, GPT-4o, GPT-5.1, Gemini-3, and Grok-4.1, were evaluated using a Structured Clinical Reasoning Pipeline.  
To ensure that foundation models operated at their peak potential 
in the human-AI comparison experiment (Section~\ref{subsec:human_ai_comparison}), 
we designed a structured clinical reasoning pipeline rather than 
relying on zero-shot inference. The complete system prompt and 
user prompt are presented in Box~\ref{box:llm_reintubation_prompt}. 
All models received identical prompts with temperature set to 0 
for deterministic output.
The pipeline enforces a three-stage reasoning process:
\begin{enumerate}
    \item \textbf{Stage~A (Data Extraction):} The model extracts 
    the most recent values for 14 physiological parameters from 
    the vitals trend image.
    \item \textbf{Stage~B (Clinical Scoring):} Extracted values 
    are evaluated against established weaning criteria, including 
    the Rapid Shallow Breathing Index (RSBI), oxygenation status 
    (PaO$_2$/FiO$_2$ ratio), acid-base balance, haemodynamic 
    stability, and neurological status, culminating in a composite 
    Spontaneous Breathing Trial (SBT) likelihood assessment.
    \item \textbf{Stage~C (Risk Prediction):} The model outputs 
    a probabilistic re-intubation risk estimate (0.0-1.0), a 
    categorical risk level, and a 3-6 sentence clinical rationale.
\end{enumerate}
All outputs were constrained to a structured JSON schema to 
enable automated parsing and comparison against ground-truth labels. 
The decision threshold was set at 0.5 (probability $\geq 0.5$ 
classified as requiring re-intubation).

\begin{tcolorbox}[
  breakable,
  colback=gray!5,
  colframe=gray!60,
  title={\textbf{Box~1: Structured prompting protocol for 
  foundation model re-intubation prediction}},
  label={box:llm_reintubation_prompt},
  fonttitle=\small,
  fontupper=\footnotesize,
  left=4pt, right=4pt, top=4pt, bottom=4pt
]

\textbf{System Prompt}
\vspace{0.3em}
\hrule
\vspace{0.5em}

\begin{verbatim}
You are an ICU clinical decision support system.

STAGE A - DATA EXTRACTION
Extract the most recent approximate values for:
- Heart rate (beats/min)
- Mean blood pressure MAP (mmHg)
- Respiratory rate RR (breaths/min)
- Tidal volume VT (mL)
- PaCO2 (mmHg), pH
- Bicarbonate (mEq/L), Potassium (mEq/L)
- Sodium (mEq/L), Creatinine (mg/dL)
- BUN (mg/dL), Lactate (mmol/L)
- Glasgow Coma Scale total score
- PaO2/FiO2 ratio (PF ratio)
Return null if unreadable.

STAGE B - CLINICAL SCORING
Compute:
- RSBI = RR / (VT / 1000)
  "good" if <80; "acceptable" if 80-105; "poor" if >105
- Oxygenation_ok: PF > 200
- Acid_base_ok: pH 7.35-7.45 and PaCO2 35-45
- Hemodynamics_ok: MAP >= 65
- Neurological_ok: GCS >= 8
- Respiratory_mechanics_ok: RR 8-30, VT adequate
- SBT_likelihood: high / moderate / low

STAGE C - 6-HOUR REINTUBATION RISK
Predict:
- probability (0.0-1.0)
- risk_level: "high_risk" (>=0.5) / "moderate_risk"
  (0.2-0.49) / "low_risk" (<0.2)
- explanation: 3-6 sentence clinical rationale

OUTPUT FORMAT: valid JSON only.
\end{verbatim}

\vspace{0.5em}
\hrule
\vspace{0.5em}
\textbf{User Prompt (per patient)}
\vspace{0.3em}
\hrule
\vspace{0.5em}

\begin{verbatim}
Below is a 12-hour vitals trend for an ICU patient.
Perform STAGE A, STAGE B, and STAGE C as defined.

Patient demographics:
- Gender: [Male/Female]
- Age: [age]
- Diagnosis: [primary diagnosis at admission]
- Ventilator duration: [hours]

[Attached: 12-hour vitals trend image]
\end{verbatim}

\end{tcolorbox}

\section{Baseline methods}
\label{sec:baseline_methods}

\subsection{Details about baseline models}
\begin{figure}[H]
    \centering
        \includegraphics[width=\textwidth]{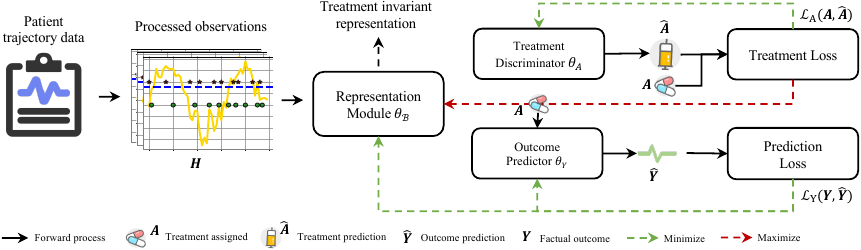}
    \caption{Graphical illustration of adversarial balancing strategies employed by baseline models.}
    \label{fig:adversarial_balancing}
\end{figure}
\textbf{Counterfactual Recurrent Network (CRN)}~\cite{conference/iclr2020/Bica}: A recurrent neural network (RNN)-based framework for counterfactual outcome estimation in longitudinal settings. The model employs a sequence-to-sequence architecture to capture patient history over time, enabling the prediction of treatment outcomes at future time points.
To address time-varying confounding bias, the model incorporates an adversarial gradient reversal (ADR) strategy. As illustrated in Figure~\ref{fig:adversarial_balancing}, patient trajectories are first encoded by an RNN-based representation module $\Theta_\mathcal{B}$ to obtain latent representations. These representations are then fed into two parallel branches:
\begin{enumerate}
    \item Outcome Prediction Branch (Green Path): The representations are combined with assigned treatments and passed to an outcome predictor, which generates treatment outcome estimates. The model minimizes a prediction loss defined in Equation~\ref{eq:prediction_loss}, encouraging the representations to be informative for outcome prediction.
    \item Adversarial Balancing Branch (Red Path): The same representations are simultaneously passed to a treatment discriminator $\Theta_{A}$, which attempts to predict the treatment assignments by minimizing a treatment loss defined in Equation~\ref{eq:treatment_loss}. Meanwhile, the representation module $\Theta_\mathcal{B}$ is trained adversarially via gradient reversal, aiming to maximize the discriminator's loss. This is designed to fool the discriminator, encouraging the learned representations to be invariant to treatment assignments and thereby mitigating confounding bias.
\end{enumerate}
Through this dual-objective design, the model learns representations that are both predictive of outcomes and balanced with respect to treatment groups, making it well-suited for counterfactual inference in time-varying clinical settings.
\begin{equation}
\mathcal{L}_{\Theta_A}  = - \sum_{j=1}^{d_a} \mathbb{I}(A_t = a_j) \log \Theta_A(\bm{\mathcal{B}}_t),
\label{eq:treatment_loss}
\end{equation}

\textbf{Causal Transformer (CT)}~\cite{conference/icml2022/Melnychuk}:
A transformer-based model designed for counterfactual outcome estimation in longitudinal healthcare settings. CT leverages both self-attention and cross-attention mechanisms to extract rich contextual representations from patient trajectories, capturing complex temporal dependencies.
To address treatment-related confounding, CT employs an adversarial balancing strategy. Unlike CRN, which uses gradient reversal to directly maximize the treatment prediction loss, CT introduces a \textit{causal domain confusion (CDC)} loss to achieve treatment-invariant representations. This strategy encourages the representation module $\Theta_{\bm{\mathcal{B}}}$ to generate embeddings that are indistinguishable across different treatment groups, thereby rendering treatment assignments uninformative with respect to the learned representations.
Specifically, during adversarial training, the treatment discriminator $\Theta_A$ is optimized to accurately classify treatment assignments using the standard treatment classification loss (Equation~\ref{eq:treatment_loss}). In contrast, the representation encoder $\Theta_{\bm{\mathcal{B}}}$ is trained to confuse the discriminator via the CDC loss (Equation~\ref{eq:CDC_loss}), which pushes the treatment prediction distribution toward uniformity, simulating random guessing:
\begin{equation}
\mathcal{L}_{\text{conf}} = - \sum_{j=1}^{d_a} \frac{1}{d_a} \log \Theta_A(\mathcal{B}_t),
\label{eq:CDC_loss}
\end{equation}
By explicitly enforcing treatment-invariant representations through CDC, CT effectively mitigates confounding bias while maintaining the temporal coherence of patient trajectories.

\textbf{Adversarial Counterfactual Temporal Inference Network (ACTIN)}~\cite{conference/icml2024/wang}: A temporal counterfactual inference framework that introduces a dual-module architecture to improve the estimation of treatment outcomes over time.
To address confounding bias, ACTIN adopts a generative adversarial network (GAN)-based strategy, which differs fundamentally from gradient reversal (as in CRN) and domain confusion (as in CT). In this approach, the treatment discriminator $\Theta_A$ is trained to distinguish between real treatment assignments $\bm{A}$ and synthetic (or ``fake'') treatments $\bm{A}_{\text{fake}}$, which are generated by randomly shuffling or sampling from the treatment distribution. These treatments are then paired with learned representations $\bm{\mathcal{B}}$ as input to the discriminator. The adversarial objective consists of two competing goals:
\begin{enumerate}
    \item The discriminator $\Theta_A$ is optimized to accurately identify whether a given treatment-representation pair is real or fake.
    \item Meanwhile, the representation module $\Theta_{\bm{\mathcal{B}}}$ is trained to fool the discriminator,encouraging it to produce representations that obscure treatment identity and thereby reduce the mutual information between treatments and representations.
\end{enumerate}
This adversarial alignment pushes $\bm{\mathcal{B}}$ toward a balanced latent space that is less predictive of treatment group, helping to mitigate confounding bias. In parallel, ACTIN also minimizes a standard prediction loss to ensure that representations remain informative for outcome estimation.
By decoupling treatment information from the learned representations via a GAN-based setup, ACTIN enables more robust and unbiased counterfactual outcome estimation across time-varying clinical data.

\subsection{Optimization properties and convergence discussion}
\label{appendix:optimization_properties_and_convergence_discussion}

The joint objective in Eq.~\ref{eq:objective_function} combines the factual prediction loss~$\mathcal{L}_{\Theta_Y}$ and the sampling-based MMD regularizer~$\mathcal{L}_{\mathcal{B}}$. Both components are bounded below, differentiable almost everywhere, and Lipschitz continuous on compact parameter domains, which ensures that the overall objective satisfies the standard conditions under which stochastic gradient descent (SGD)-type algorithms converge to first-order stationary points for non-convex problems. Although global optimality cannot be guaranteed, these regularity properties imply that the Optimization landscape is well-behaved in the sense required for contemporary deep learning systems.

In practice, however, jointly optimizing the predictive and balancing losses introduces non-trivial challenges. A large balancing weight~$\lambda$ applied too early in training may suppress physiologically meaningful variability in the learned representation, leading to underfitting or even representation collapse. To mitigate this effect, we employ a curriculum-style adaptive schedule for~$\lambda$ (Algorithm~\ref{alg:pseudocode}, line~5). Specifically, in training epoch $e$ of $E$ total epochs, the balancing coefficient is updated according to the sigmoidal progression:
\begin{equation}
    \lambda_e = \frac{2}{1 + \exp(-10 \cdot \frac{e}{E})} - 1.
\label{appendix_eq:lambda_schedule}
\end{equation}
This schedule begins near zero, gradually increases during mid-training, and asymptotically approaches one. Early training therefore prioritises minimizing $\mathcal{L}_{\Theta_Y}$, enabling the encoder to learn a stable embedding of physiological dynamics. As training progresses, the increasing~$\lambda_e$ progressively strengthens distributional alignment in the latent space. This progressive scheme improves Optimization stability and avoids premature over-regularization.

\subsection{Implementation details of the ventilator re-intubation classifier}

The ventilator re-intubation classifier was implemented in PyTorch.  
Each input sequence comprised 12 time steps of 14 features representing vital signs, augmented with time-invariant statistical descriptors (mean, standard deviation, and temporal slope) that were repeated along the temporal axis.  

The model architecture, referred to as \texttt{CNN1DAvg}, included two $3\times1$ convolutional layers with ReLU activations, followed by a residual block composed of two additional $3\times1$ convolutions and a skip connection.  
A global average pooling layer aggregated temporal information, and the resulting representation was passed through a dropout layer ($p=0.3$) and a linear classification head that produced the final scalar logit.  
Training was conducted with the AdamW optimizer (learning rate $10^{-3}$, weight decay $10^{-4}$), using binary cross-entropy loss with a positive class weight equal to the ratio of negative-to-positive samples.  
In selected runs, a focal loss variant ($\alpha = 0.25, \gamma = 2.0$) was adopted to emphasize difficult cases.  
A weighted random sampler ensured class balance during training.  
Learning rate scheduling followed cosine annealing with warm restarts ($T_0=5$, $T_{\text{mult}}=2$).  
Gradients were clipped to $\lVert\nabla\theta\rVert_2 < 1.0$ at each step to improve stability.  
Training was performed for 100 epochs with a batch size of 256 on an NVIDIA A40 GPU.  
The optimal classification threshold was determined on the validation set by maximizing the F1-score.  
All implementation details, including data augmentation, masking, and reproducibility controls, are available in the released code repository.

\subsection{Implementation details of the reconstruction decoder}
\label{subsec:reconstruction_decoder}

The reconstruction decoder is trained in a two-stage procedure to evaluate the information content of the learned balanced representations $\bm{\mathcal{B}}$. This two-stage design ensures that the decoder's reconstruction quality reflects the encoder's representation fidelity rather than being confounded by joint Optimization dynamics.

\textbf{Stage~1: Main model training (encoder + outcome head).}
In the first stage, the full GITO model, comprising the TCN-based encoder, the balanced representation module, and the outcome prediction head, is trained end-to-end for 400 epochs using the joint objective $\mathcal{L}_{\Theta_Y} + \lambda_e \mathcal{L}_{\mathcal{B}}$ with the adaptive $\lambda$ schedule described in Appendix (Optimization Properties). The Adam optimizer is used with a learning rate of $10^{-3}$ and weight decay of $10^{-4}$. At the end of Stage~1, the best model checkpoint is selected based on the validation loss.

\textbf{Stage~2: Reconstruction decoder training (frozen encoder).}
After Stage~1, \emph{all encoder parameters are frozen} (\texttt{requires\_grad=False}). A lightweight LSTM decoder and a linear projection layer are then trained to reconstruct the original input co-variates and static features from the balanced representations. The architectural details are:
\begin{itemize}
    \item \textbf{LSTM decoder:} A single-layer LSTM that takes $\bm{\mathcal{B}}_t \in \mathbb{R}^{48}$ (the balanced representation at each time step) as input, with a hidden size of 25.
    \item \textbf{Projection layer:} A linear layer mapping from the LSTM hidden state ($\mathbb{R}^{25}$) to the combined covariate, static feature space ($\mathbb{R}^{d_x + d_s}$, where $d_x$ is the number of time-varying co-variates and $d_s$ is the number of static features after one-hot encoding).
\end{itemize}
The reconstruction loss is mean squared error (MSE), masked by the active entries indicator to account for variable-length sequences:
\begin{equation}
    \mathcal{L}_{\text{recon}} = \frac{1}{NT} \sum_{i=1}^{N} \sum_{t=1}^{T} \lVert \hat{\bm{X}}_{i,t} - \bm{X}_{i,t} \rVert^2,
\end{equation}
where $\hat{\bm{X}}_{i,t}$ denotes the reconstructed covariate vector and $\bm{X}_{i,t} = [\bm{x}_{i,t}; \bm{s}_i]$ is the concatenation of the observed time-varying co-variates and the (time-expanded) static features.
Stage~2 training is conducted for 300 epochs using the Adam optimizer with a learning rate of $10^{-3}$ and a batch size of 64. Only the LSTM decoder and projection layer parameters are updated; the number of trainable parameters in this stage is approximately 2\% of the total model parameters. Validation loss is monitored at each epoch, and the best decoder checkpoint is retained.
After training, per-variable reconstruction quality is assessed on the validation set by computing variable-specific MSE and $R^2$ scores, separately for time-varying co-variates and static features. These metrics directly inform the $\Delta R^2$ analysis reported in the main text (Section~\ref{subsec:ood_mimiciii_results}).

\subsection{Generative AI configuration and prompt engineering}
\label{appendix:llm-full-output}

\textbf{Model configuration and input encoding}
The interpretability module utilizes a multimodal LLM (GPT-4o by default; the platform supports user-selectable alternatives) configured with a temperature of 0.7 to balance creativity with adherence to clinical facts. The maximum output token limit is set to 800 to accommodate the four-paragraph output structure. Inputs are constructed using a hybrid schema:
\begin{itemize}
    \item \textbf{Textual context:} Patient demographics (age, primary diagnosis); serialized MAP predictions for four treatment scenarios (None, Vaso, Vent, Both) over five time steps; and calculated statistics for the top-5 feature-attribution variables (latest value, moving average, linear trend direction).
    \item \textbf{Visual context:} Three high-resolution plots—(i)~Vital Signs Trend, (ii)~Prediction Trajectory, and (iii)~Patient History with treatment markers—are rendered using \texttt{matplotlib}, converted to Base64 strings, and injected into the model's vision context.
\end{itemize}

\textbf{Prompt structure and constraints}
The prompting strategy enforces an ``Extraction-to-Interpretation'' logic via three components:
\textbf{1.\ System persona and constraints.}
The system prompt defines the agent as an \textit{``AI-based clinical decision-support analyst''} and imposes explicit negative constraints: \textit{``You are NOT a treating physician,''} \textit{``You must NOT issue definitive medical advice or prescriptive treatment orders,''} and \textit{``Avoid prescriptive or guideline-based language.''} This ensures the tone remains analytical and descriptive rather than directive.
\textbf{2.\ Dynamic context injection (user prompt).}
The user prompt is dynamically assembled at runtime to include:
\begin{itemize}
    \item \textbf{Data anchoring:} A list of the top-5 key variables with their Integrated Gradients contribution scores, explicit latest values, and trend directions (e.g., \textit{``Respiratory rate (contribution 0.15): Latest 24/min, Trend: increasing''}). This serves as the Stage~I ground truth to prevent numerical hallucinations.
    \item \textbf{Comparison discipline instructions:} The model is instructed to (i)~use approximate deltas (e.g., \textit{``{\raise.17ex\hbox{$\scriptstyle\sim$}}3-5\% higher''}) rather than absolute precision; (ii)~explicitly state if scenarios are \textit{``clinically similar''} to avoid over-interpreting noise; and (iii)~follow a \emph{minimal-intervention rule}: if a less intensive strategy (especially None) is predicted to reach and remain within the diagnosis-appropriate target range, it should be treated as sufficient.
\end{itemize}
\textbf{3.\ Output formatting rules.}
The model is constrained to produce exactly four paragraphs without markdown headers:
\begin{itemize}
    \item \textbf{Paragraph~1 (Target measurement context):} summarize the displayed measurement's range, variability, and alignment with clinically relevant reference ranges over the observed period.
    \item \textbf{Paragraph~2 (Influential vital signs):} For each of the top-5 variables, report the attribution score, approximate current value, and clinical implication.
    \item \textbf{Paragraph~3 (Predicted trajectory interpretation):} Compare the four treatment strategies in terms of magnitude, trend, and stability of the predicted MAP trajectories, explicitly linking predictions to the patient's current physiological state and historical treatment responses.
    \item \textbf{Paragraph~4 (Model preference distribution):} Report a numerical preference score for each of the four treatment scenarios as approximate percentages summing to 100\%, reflecting a trade-off between (i)~sufficiency in achieving the diagnosis-specific target range, (ii)~trajectory stability, (iii)~consistency with historical responses, and (iv)~intervention intensity following a minimal necessary intervention principle.
\end{itemize}
An abbreviated example of the generated output structure:
\begin{small}
\begin{verbatim}
{
  "paragraph_1": "MAP has fluctuated between 58 and 72 mmHg ...",
  "paragraph_2": "Heart rate (contribution 0.23) is elevated ...",
  "paragraph_3": "The model predicts higher MAP under Vaso ...",
  "preference_scores": {
    "None": 15,
    "Vasopressors": 40,
    "Ventilation": 20,
    "Both": 25
  }
}
\end{verbatim}
\end{small}
\textbf{Fallback mechanism}
To ensure system robustness in clinical settings, a deterministic fallback mechanism is implemented. In the event of an API failure or a violation of the formatting constraints (detected via regex parsing), the system reverts to a template-based generator. This fallback concatenates the pre-computed variable importance rankings and statistical trends into a simplified text summary, ensuring that decision support remains available even without LLM generation.

\subsection{Hyperparameter tuning}

Following the methods used in ACTIN \cite{conference/icml2024/wang}, we conduct hyperparameter optimization for all baseline models using random searches. The ranges for the random searches for CRN, CRN-sMMD, CT, CT-MMD, ACTIN and ACTIN-MMD are provided in Tables \ref{tab:crn_parameters}, \ref{tab:ct_parameters}, \ref{tab:actin_parameters}, respectively. Following the original research, we conduct hyperparameter optimization for two distinct base models, TCN and LSTM for ACTIN. It is worth noting that all sub-models within ACTIN utilizes the same base model within our experiments. 

\begin{table}[htbp]
\centering
\caption{The ranges for hyperparameter tuning of CRN and CRN-sMMD for synthetic tumor growth and MIMIC-III datasets. The symbols $\Theta_{en}$ and $\Theta_{de}$ denote the Encoder and Decoder sub-models, respectively.}
\begin{tabular}{l|c|c}
\hline
\textbf{Hyperparameter} & \textbf{Range for Tumor-growth} & \textbf{Range MIMIC-III} \\
\hline
LSTM layers & 1 & 1, 2 \\
Learning rate & 0.01, 0.001, 0.0001 & 0.01, 0.001, 0.0001 \\
Minibatch size ($\Theta_{en}$) & 64, 128, 256 & 64, 128, 256 \\
Minibatch size ($\Theta_{de}$) & 256, 512, 1024 & 256, 512, 1024 \\
LSTM hidden units ($\Theta_{en}$) & 3, 6, 12, 18, 24 & 36, 72, 144 \\
LSTM hidden units ($\Theta_{de}$) & 3, 6, 12, 18, 24 & 36, 72, 144 \\
BR size $D^{en}$ ($\Theta_{en}$) & 3, 6, 12, 18, 24 & 36, 72, 144 \\
BR size $D^{de}$ ($\Theta_{de}$) & 3, 6, 12, 18, 24 & 47, 94, 188 \\
FC hidden units ($\Theta_{en}$) & $0.5D^{en}$, $1D^{en}$, $2D^{en}$, $3D^{en}$, $4D^{en}$ & $0.5D^{en}$, $1D^{en}$, $2D^{en}$ \\
FC hidden units ($\Theta_{de}$) & $0.5D^{de}$, $1D^{de}$, $2D^{de}$, $3D^{de}$, $4D^{de}$ & $0.5D^{de}$, $1D^{de}$, $2D^{de}$ \\
LSTM dropout rate & 0.1, 0.2, 0.3, 0.4, 0.5 & 0.1, 0.2, 0.3, 0.4, 0.5 \\
Random search iterations ($\Theta_{en}$) & 50 & 50 \\
Random search iterations ($\Theta_{de}$) & 30 & 30 \\
Number of epochs & 100, 200 & 200 \\
\hline
\end{tabular}
\label{tab:crn_parameters}
\end{table}

\begin{table}[htbp]
\centering
\caption{The ranges for hyperparameter tuning of CT and CT-sMMD for synthetic tumor growth and MIMIC-III datasets.}
\begin{tabular}{l|c|c}
\hline
\textbf{Hyperparameter} & \textbf{Range Tumor-growth} & \textbf{Range MIMIC-III} \\
\hline
Transformer blocks & 1 & 1, 2 \\
Learning rate & 0.01, 0.001, 0.0001 & 0.01, 0.001, 0.0001 \\
Minibatch size & 64, 128, 256 & 32, 64 \\
Attention heads & 2 & 2, 4 \\
Transformer units & 4, 8, 12, 16 & 24, 48, 64 \\
BR size $D$ & 2, 4, 8, 12, 16 & 22, 44, 88 \\
FC hidden units & $0.5D$, $1D$, $2D$, $3D$, $4D$ & $0.5D$, $1D$, $2D$ \\
Sequential dropout rate & 0.1, 0.2, 0.3, 0.4, 0.5 & 0.1, 0.2, 0.3, 0.4, 0.5 \\
Max positional encoding & 15 & 30 \\
Random search iterations & 50 & 50 \\
Number of epochs & 150 & 300 \\
\hline
\end{tabular}
\label{tab:ct_parameters}
\end{table}

\begin{table}[htbp]
\centering
\caption{The ranges for hyperparameter tuning of ACTIN for synthetic tumor growth and MIMIC-III datasets.}
\begin{tabular}{l|c|c}
\hline
\textbf{Hyperparameter} & \textbf{Range Tumor-growth} & \textbf{Range MIMIC-III} \\
\hline
Linear transformation size & 4, 8, 16 & 16, 32, 64 \\
Learning rate $l$ & 0.01, 0.002, 0.001 & 0.01, 0.002, 0.001 \\
Learning rate $l_{\mathcal{D}}$ & 0.001, 0.0002, 0.0001 & 0.001, 0.0002, 0.0001 \\
Minibatch size & 64, 128, 256 & 64, 128, 256 \\
BR size $D$ & 8, 12, 16, 24, 36 & 16, 32, 64 \\
$\lambda$ & 0.01 & 0.01 \\
\hline
\multicolumn{3}{l}{\textbf{TCN-based}} \\
\hline
\hspace{3mm} Kernel sizes & 2, 3 & 2, 3 \\
\hspace{3mm} Dilation factors & 2, 3 & 2, 3 \\
\hspace{3mm} Channel size $d_{c}$ & 4, 8, 12, 16, 24, 36 & 28, 32, 36, 64 \\
\hline
\multicolumn{3}{l}{\textbf{LSTM-based}} \\
\hline
\hspace{3mm} LSTM layers & 1 & 1, 2 \\
\hspace{3mm} LSTM hidden units & 4, 8, 12, 16 & 16, 32, 64 \\
\hline
FC hidden units & 16, 32, 64 & 16, 32, 64 \\
Dropout rate & 0.1, 0.2, 0.3 & 0.1, 0.2, 0.3 \\
Random search iterations & 50 & 50 \\
Number of epochs & 150 & 300 \\
\hline
\end{tabular}
\label{tab:actin_parameters}
\end{table}

\section{Experiment results}
\label{sec:experiment_results}

\subsection{Experiments reults on MIMIC-III dataset}
\label{subsec:ood_mimiciii_results}

In this section, we present additional experimental results on factual outcome estimation under the scenario of diagnosis-based distribution shift, involving three baseline models and their sMMD-enhanced versions. The three major disease categories are presented in Table~\ref{tab:cvd_results} (cardiovascular diseases), Table~\ref{tab:nd_results} (neurological disorders), and Table~\ref{tab:iaid_results} (infectious and inflammatory diseases), respectively.
These models are trained on White patients and evaluated on three non-White ethnicity subgroups across three disease categories.

In our experiments, both CRN and ACTIN models, when incorporated with the sMMD strategy, exhibited significant performance improvements under out-of-distribution (OOD) settings across all three disease cohorts. Specifically, the models demonstrated consistently enhanced generalisation to unseen patient populations. This indicates the effectiveness of sMMD in improving robustness. In contrast, CT-sMMD only showed performance gains in neurological disorders, while in other scenarios the improvements were marginal or negligible.
These findings suggest that sMMD is effective across different treatment response prediction tasks, whereas the benefits of CT-sMMD may be limited to certain patient groups or clinical conditions.

\begin{table*}[htbp]
\centering
\caption{Multi-step-ahead prediction results on the MIMIC-III dataset in patients within cardiovascular disease. Shown: RMSE as mean $\pm$ standard deviation over ten runs.}
\label{tab:cvd_results}
\resizebox{\textwidth}{!}{
\begin{tabular}{@{}llcccccc@{}}
\toprule
      &       & $\tau = 1$ & $\tau = 2$ & $\tau = 3$ & $\tau = 4$ & $\tau = 5$ & $\tau = 6$ \\ \midrule
\multirow{7}{*}{Asian} 
      & CRN       & 4.68$\pm$0.55  & 9.93$\pm$0.78  & 10.74$\pm$1.10  &11.02$\pm$1.02 & 11.06$\pm$0.80 & 11.01$\pm$0.64 \\
      & CRN-sMMD   & 4.64$\pm$0.33  & 8.03$\pm$0.74\sigstar\sigstar  &8.53$\pm$0.71\sigstar\sigstar  & 8.87$\pm$0.76\sigstar\sigstar & 9.11$\pm$0.89\sigstar\sigstar & 9.38$\pm$1.00\sigstar\sigstar \\
      & CT        & 4.19$\pm$0.44  & 7.75$\pm$0.77  &8.22$\pm$0.74  & 8.54$\pm$0.77  & 8.74$\pm$0.82 &8.95$\pm$0.85 \\
      & CT-sMMD    & 4.15$\pm$0.47  & 7.75$\pm$0.79  &8.22$\pm$0.74  & 8.54$\pm$0.77  & 8.74$\pm$0.82 &8.95$\pm$0.84 \\
      & ACTIN     & 4.45$\pm$0.35  & 5.10$\pm$0.45  & 5.39$\pm$0.48  &5.66$\pm$0.60 & 5.89$\pm$0.75 & 6.11$\pm$0.89 \\
    & ACTIN-sMMD & 4.39$\pm$0.38  & 4.84$\pm$0.45  & 5.00$\pm$0.33\sigstar  &5.14$\pm$0.32\sigstar & 5.28$\pm$0.32\sigstar & 5.38$\pm$0.31\sigstar \\ \midrule
\multirow{7}{*}{African} 
      & CRN       &5.28$\pm$0.51  &11.06$\pm$1.86   & 11.89$\pm$1.96   &12.25$\pm$1.81 &12.31$\pm$1.47  &12.28$\pm$1.16 \\
      & CRN-sMMD   & 5.25$\pm$0.49  &9.47$\pm$0.46\sigstar  & 10.00$\pm$0.46\sigstar\sigstar  & 10.43$\pm$0.42\sigstar\sigstar &10.71$\pm$0.45\sigstar\sigstar & 10.96$\pm$0.49\sigstar\sigstar \\
      & CT        & 4.84$\pm$0.54  & 9.24$\pm$0.46  &9.75$\pm$0.46  & 10.12$\pm$0.39  & 10.36$\pm$0.38 &10.56$\pm$0.41 \\
      & CT-sMMD    & 4.71$\pm$0.58  & 8.98$\pm$0.76  &9.47$\pm$0.84  & 9.86$\pm$0.86  & 10.09$\pm$0.91 &10.29$\pm$0.94 \\
      & ACTIN     & 5.17$\pm$0.51  & 5.58$\pm$0.59  & 5.76$\pm$0.59  & 6.04$\pm$0.66  & 6.24$\pm$0.76  &6.51$\pm$0.86 \\
      & ACTIN-sMMD & 4.97$\pm$0.61  & 5.28$\pm$0.67\sigstar  & 5.43$\pm$0.58\sigstar  & 5.54$\pm$0.58\sigstar  & 5.62$\pm$0.56\sigstar  &5.79$\pm$0.58\sigstar \\ \midrule
\multirow{7}{*}{Latino} 
    & CRN       &4.30$\pm$0.45   &10.73$\pm$0.77   &11.67$\pm$0.74   &11.82$\pm$0.66 &11.56$\pm$0.63  &11.24$\pm$0.69  \\
    & CRN-sMMD   &4.11$\pm$0.35 &8.05$\pm$0.85\sigstar   &8.60$\pm$0.86\sigstar\sigstar   &8.95$\pm$0.78\sigstar\sigstar &9.27$\pm$0.78\sigstar\sigstar &9.56$\pm$0.78\sigstar\sigstar    \\
    & CT       &3.70$\pm$0.33   &7.96$\pm$0.85   &8.52$\pm$0.86   &8.86$\pm$0.79 &9.15$\pm$0.80 &9.36$\pm$0.80 \\
    & CT-sMMD    &3.66$\pm$0.31   &7.94$\pm$0.85   &8.51$\pm$0.85   &8.85$\pm$0.76 &9.14$\pm$0.75  &9.36$\pm$0.74  \\
    & ACTIN     & 4.14$\pm$0.35  & 4.62$\pm$0.50  & 4.96$\pm$0.61  &5.26$\pm$0.75 & 5.54$\pm$0.90 & 5.83$\pm$1.05 \\
    & ACTIN-sMMD & 3.82$\pm$0.48\sigstar  & 4.11$\pm$0.56\sigstar\sigstar  & 4.27$\pm$0.55\sigstar\sigstar  &4.43$\pm$0.54\sigstar\sigstar & 4.56$\pm$0.54\sigstar\sigstar & 4.71$\pm$0.55\sigstar\sigstar \\
\bottomrule
\end{tabular}
}
\end{table*}

\begin{table*}[htbp]
\centering
\caption{Multi-step-ahead prediction results on the RW dataset in out of distribution settings (patients within neurological disorders ). Shown: RMSE as mean $\pm$ standard deviation over ten runs.}
\label{tab:nd_results}
\resizebox{\textwidth}{!}{
\begin{tabular}{@{}llcccccc@{}}
\toprule
      &       & $\tau = 1$ & $\tau = 2$ & $\tau = 3$ & $\tau = 4$ & $\tau = 5$ & $\tau = 6$ \\ \midrule
\multirow{7}{*}{Asian} 
      & CRN       & 5.60$\pm$0.98  & 11.20$\pm$2.23  & 11.92$\pm$2.25  & 12.19$\pm$1.97  & 12.34$\pm$1.67  &12.48$\pm$1.48 \\
      & CRN-sMMD   & 5.37$\pm$0.87  & 9.69$\pm$1.01\sigstar  & 10.38$\pm$1.04\sigstar  & 10.81$\pm$1.01\sigstar  & 11.33$\pm$1.04\sigstar &11.90$\pm$1.05\sigstar \\
      & CT       &5.06$\pm$1.02   &9.15$\pm$0.97   &9.62$\pm$0.97   &9.95$\pm$1.01\sigstar &10.32$\pm$1.02 &10.77$\pm$0.98 \\
      & CT-sMMD    &4.69$\pm$0.96\sigstar    &8.53$\pm$1.25\sigstar    &9.04$\pm$1.20\sigstar    &9.38$\pm$1.26\sigstar   &9.69$\pm$1.34\sigstar   &10.01$\pm$1.40\sigstar   \\
      & ACTIN     & 5.53$\pm$1.01  & 6.23$\pm$1.09  & 6.52$\pm$1.07  & 6.69$\pm$1.12  & 7.02$\pm$1.27  &7.30$\pm$1.31 \\
      & ACTIN-sMMD & 5.37$\pm$1.17 & 5.91$\pm$1.21\sigstar\sigstar  &6.10$\pm$1.19\sigstar\sigstar  & 6.15$\pm$1.17\sigstar\sigstar  & 6.40$\pm$1.19\sigstar\sigstar  &6.64$\pm$1.17\sigstar\sigstar \\ \midrule
\multirow{7}{*}{African} 
      & CRN        &5.41$\pm$0.46   &11.47$\pm$2.05  &12.33$\pm$2.24   &12.56$\pm$2.07 &12.57$\pm$1.78  &12.53$\pm$1.46   \\
      & CRN-sMMD   &5.43$\pm$0.45   &9.88$\pm$0.54\sigstar  &10.62$\pm$0.54\sigstar   &10.98$\pm$0.53\sigstar &11.27$\pm$0.56\sigstar  &11.55$\pm$0.60\sigstar   \\
      & CT        & 4.99$\pm$0.54  &9.63$\pm$0.49  & 10.32$\pm$0.54  & 10.66$\pm$0.53 & 10.90$\pm$0.54 &11.14$\pm$0.57 \\
      & CT-sMMD   & 4.91$\pm$0.48  &9.46$\pm$0.52\sigstar   & 10.01$\pm$0.58\sigstar   & 10.35$\pm$0.51\sigstar  &10.58$\pm$0.52\sigstar  & 10.82$\pm$0.57\sigstar  \\
      & ACTIN     & 5.37$\pm$0.51  &5.94$\pm$0.67   &6.34$\pm$0.80  &6.59$\pm$0.86    &6.84$\pm$0.93    &7.10$\pm$1.05\\
      & ACTIN-sMMD & 5.10$\pm$0.54\sigstar  & 5.48$\pm$0.65\sigstar\sigstar  & 5.72$\pm$0.74\sigstar\sigstar  & 5.83$\pm$0.76\sigstar\sigstar  & 5.95$\pm$0.78\sigstar\sigstar  &6.07$\pm$0.80\sigstar\sigstar \\ \midrule
\multirow{7}{*}{Latino} 
    & CRN       &5.27$\pm$0.68   &11.44$\pm$1.95  &12.50$\pm$2.16   &12.80$\pm$2.01 &12.98$\pm$1.72  &13.10$\pm$1.45   \\
    & CRN-sMMD   &5.26$\pm$0.72   &9.91$\pm$1.02\sigstar  &10.79$\pm$1.16\sigstar  &11.21$\pm$1.15\sigstar &11.68$\pm$1.12\sigstar &12.10$\pm$1.12\sigstar   \\
    & CT       &4.59$\pm$0.80   &9.50$\pm$1.30   &10.32$\pm$1.49   &10.72$\pm$1.52 &11.12$\pm$1.57 &11.49$\pm$1.63 \\
    & CT-sMMD    &4.57$\pm$0.79   &9.50$\pm$1.29   &10.32$\pm$1.48   &10.72$\pm$1.52 &11.12$\pm$1.57  &11.49$\pm$1.64  \\
    & ACTIN     &5.23$\pm$0.70   &5.68$\pm$0.98   &6.00$\pm$1.03  &6.32$\pm$1.06 &6.66$\pm$1.09   &7.06$\pm$1.21   \\
    & ACTIN-sMMD & 5.01$\pm$0.60\sigstar  & 5.35$\pm$0.95\sigstar  & 5.54$\pm$0.89\sigstar  &5.74$\pm$0.82\sigstar & 5.96$\pm$0.78\sigstar &6.22$\pm$0.78\sigstar \\
\bottomrule
\end{tabular}
}
\end{table*}

\begin{table*}[htbp]
\centering
\caption{Multi-step-ahead prediction results on the MIMIC-III dataset in out of distribution settings (patients with infectious and inflammatory diseases). Shown: RMSE as mean $\pm$ standard deviation over ten runs.}
\label{tab:iaid_results}
\resizebox{\textwidth}{!}{
\begin{tabular}{@{}llcccccc@{}}
\toprule
      &       & $\tau = 1$ & $\tau = 2$ & $\tau = 3$ & $\tau = 4$ & $\tau = 5$ & $\tau = 6$ \\ \midrule
\multirow{7}{*}{Asian} 
      & CRN       & 4.88$\pm$0.48  & 10.52$\pm$1.15  & 11.43$\pm$1.16  & 11.89$\pm$1.01  & 12.21$\pm$0.80  &12.36$\pm$0.60 \\
      & CRN-sMMD   & 4.86$\pm$0.50  & 9.47$\pm$0.68\sigstar  & 10.28$\pm$0.74\sigstar  & 10.92$\pm$0.79\sigstar  & 11.46$\pm$0.82\sigstar &11.89$\pm$0.91\sigstar \\
      & CT     & 4.52$\pm$0.48  & 9.17$\pm$0.68  & 9.90$\pm$0.86  & 10.43$\pm$0.93  & 10.91$\pm$0.96  &11.26$\pm$1.03 \\
      & CT-sMMD &4.37$\pm$0.62 &8.40$\pm$1.25\sigstar   &9.12$\pm$1.49\sigstar  & 9.54$\pm$1.64\sigstar  & 9.86$\pm$1.79\sigstar  &10.15$\pm$1.85\sigstar \\
      & ACTIN     & 4.82$\pm$0.50  & 5.43$\pm$0.53  & 5.76$\pm$0.60  & 6.13$\pm$0.72  & 6.39$\pm$0.85  &6.62$\pm$0.96 \\
      & ACTIN-sMMD & 4.60$\pm$0.42\sigstar & 5.02$\pm$0.40\sigstar  &5.20$\pm$0.39\sigstar  & 5.43$\pm$0.36\sigstar  & 5.54$\pm$0.35\sigstar  &5.65$\pm$0.32\sigstar \\ \midrule
\multirow{7}{*}{African} 
      & CRN        &5.16$\pm$0.69   &11.06$\pm$1.94  &11.96$\pm$2.17   &12.30$\pm$2.04 &12.39$\pm$1.79  &12.33$\pm$1.55   \\
      & CRN-sMMD   &5.28$\pm$0.61   &9.66$\pm$0.58\sigstar  &10.42$\pm$0.61\sigstar   &10.89$\pm$0.64\sigstar &11.27$\pm$0.64\sigstar  &11.54$\pm$0.62\sigstar   \\
      & CT        & 4.93$\pm$0.65  &9.42$\pm$0.64  & 10.13$\pm$0.66& 10.53$\pm$0.70 & 10.85$\pm$0.71 &11.01$\pm$0.70 \\
      & CT-sMMD   & 4.92$\pm$0.65  &9.40$\pm$0.62  & 10.12$\pm$0.66  & 10.51$\pm$0.70 &10.82$\pm$0.71 & 10.98$\pm$0.70 \\
      & ACTIN     & 5.30$\pm$0.57  &5.67$\pm$0.62   &6.07$\pm$0.73  &6.35$\pm$0.82    &6.57$\pm$0.90    &6.75$\pm$1.02\\
      & ACTIN-sMMD & 5.02$\pm$0.66\sigstar  & 5.29$\pm$0.67\sigstar  & 5.53$\pm$0.81\sigstar  & 5.70$\pm$0.87\sigstar  & 5.81$\pm$0.88\sigstar  & 5.84$\pm$0.89\sigstar \\ \midrule
\multirow{7}{*}{Latino} 
    & CRN       &5.31$\pm$0.48   &10.93$\pm$1.32  &11.56$\pm$1.44   &11.77$\pm$1.28 &11.75$\pm$1.05  &11.69$\pm$0.86   \\
    & CRN-sMMD   &5.43$\pm$0.46   &9.83$\pm$1.04\sigstar  &10.31$\pm$0.92\sigstar   &10.65$\pm$0.79\sigstar &10.90$\pm$0.67\sigstar  &11.15$\pm$0.63\sigstar   \\
    & CT       &4.88$\pm$0.80   &9.69$\pm$0.91   &10.17$\pm$0.86   &10.49$\pm$0.86 &10.71$\pm$0.86 &10.89$\pm$0.82 \\
    & CT-sMMD    &4.90$\pm$0.75   &9.69$\pm$0.93   &10.16$\pm$0.89   &10.49$\pm$0.90 &10.71$\pm$0.89  &10.89$\pm$0.87  \\
    & ACTIN     &5.08$\pm$0.54   &5.67$\pm$0.67   &5.91$\pm$0.78  &6.12$\pm$0.83 &6.40$\pm$1.05   &6.57$\pm$1.19   \\
    & ACTIN-sMMD & 5.21$\pm$0.65  & 5.66$\pm$0.67  & 5.80$\pm$0.70  &5.99$\pm$0.71 & 6.16$\pm$0.74\sigstar &6.20$\pm$0.76\sigstar \\
\bottomrule
\end{tabular}
}
\end{table*}

\begin{table*}
\caption{Multi-step prediction results on the fully-synthetic tumor-growth dataset ($\gamma = 10$, lower values are better, with the best highlighted in bold), where woBMR represents model without balancing strategy. Shown: RMSE as mean $\pm$ standard deviation over ten runs.}
\resizebox{\textwidth}{!}{
\begin{tabular}{@{}lcccccccc@{}}
\toprule
      & $\tau = 1$ & $\tau = 2$ & $\tau = 3$ & $\tau = 4$ & $\tau = 5$ & $\tau = 6$ & Average\\ \midrule
      ACTIN &\textbf{3.57$\pm$0.51} &3.33$\pm$1.46 &4.11$\pm$1.68 &4.56$\pm$1.70 &4.71$\pm$1.60 &4.61$\pm$1.47 &4.14$\pm$0.58\\
      ACTIN-woBRM &4.52$\pm$1.14 &1.94$\pm$0.65 &\textbf{2.58$\pm$0.92} &\textbf{3.08$\pm$1.15} &3.53$\pm$1.38 & 3.92$\pm$1.60 &3.26$\pm$0.93\\
      ACTIN-sMMD & 4.38$\pm$0.97 & \textbf{1.94$\pm$0.63} &2.59$\pm$0.86 &3.10$\pm$1.04 &\textbf{3.51$\pm$1.15} &\textbf{3.87$\pm$1.25} &\textbf{3.24$\pm$0.88}\\
\bottomrule
\end{tabular}
\label{tab:tg_multi_step_prediction}
}
\end{table*}

\section{Case study: patient selection and interpretability workflow}
\label{appendix:case_studies}
This section details the patient selection criteria and the end-to-end workflow used to generate the interpretability analysis presented in the main text (Section~\ref{subsec:case_study}).
\subsection{Patient selection}
The case study patient was selected from the MIMIC-III ventilator weaning subcohort ($N = 205$) according to the following criteria:
\begin{enumerate}
    \item \textbf{Diagnosis:} The patient's primary admission diagnosis was septic shock (ICD-9 785.52), a condition in which vasopressor therapy decisions are clinically impactful and where the trade-off between treatment escalation and conservative management is well-characterised.
    \item \textbf{Treatment diversity:} The patient's ICU trajectory included periods of both vasopressor administration and mechanical ventilation, as well as intervals without active treatment. This diversity ensured that all four counterfactual scenarios (None, Vaso, Vent, Both) were clinically plausible given the patient's history.
    \item \textbf{Non-trivial prediction:} GITO's predicted re-intubation risk for this patient fell within an intermediate probability range (neither near 0 nor near 1), representing a clinically ambiguous case where decision-support tools provide the greatest added value.
    \item \textbf{Outcome availability:} The patient's subsequent clinical trajectory (successful recovery without vasopressor escalation) was documented, enabling retrospective validation of the model's preference distribution.
\end{enumerate}
\subsection{End-to-end interpretability workflow}
The following steps describe the complete pipeline from raw patient data to the final LLM-generated explanation shown in Box~\ref{appendix:full_LLMreasoning}:
\begin{enumerate}
    \item \textbf{Data ingestion.} The patient's hourly time-series data (25 vital signs, 3 static attributes, 2 binary treatments) were loaded from the MIMIC-extract preprocessed dataset and normalized using the cohort-level Z-score parameters (see Methods: Data Preprocessing).
    \item \textbf{Multi-step outcome prediction.} GITO's encoder-decoder architecture generated MAP predictions over a 5-step horizon ($\tau = 5$, corresponding to 5 hours) under each of the four treatment scenarios. At each rollout step, the predicted outcome was fed back as input for the next step (autoregressive inference with teacher forcing disabled).
    \item \textbf{Integrated Gradients attribution.} For each prediction step $j \in \{1, \dots, 5\}$, Integrated Gradients was applied with the cohort-mean baseline to compute per-variable attribution scores $\phi^{(j)}_{i}$. Scores were averaged across the prediction horizon to obtain $\omega_i^{\text{raw}}$ and normalized via softmax to produce $\omega_i$ (see Methods: Interpretability Pipeline, Eqs.~\ref{eq:avg_ig}-\ref{eq:softmax_ig}). The top-$k$ ($k=5$) variables were selected for downstream reporting.
    \item \textbf{Visualisation rendering.} Three charts were generated using \texttt{matplotlib}:
    \begin{itemize}
        \item \textit{Vital Signs Trend:} Strip plots of the top-5 variables with normal-range shading.
        \item \textit{Prediction Trajectory:} Historical MAP with four counterfactual projection lines.
        \item \textit{Patient History:} Full MAP trajectory with treatment markers (shape-coded: circle = None, triangle = Vaso, square = Vent, star = Both) and baseline/follow-up separation.
    \end{itemize}
    All plots were converted to Base64-encoded PNG images for injection into the LLM's vision context.
    \item \textbf{Statistics extraction.} For each top-5 variable, the latest value, moving average, and linear trend direction were computed from the raw time series and serialized as structured text (Stage~I grounding data).
    \item \textbf{LLM inference.} The system prompt, dynamic user prompt (with data anchoring and comparison discipline instructions), and three visual inputs were assembled and sent to GPT-4o (temperature = 0.7, max tokens = 800). The model produced a four-paragraph response including the preference distribution (see Appendix~\ref{appendix:llm-full-output} for the full prompt structure).
    \item \textbf{Output validation.} The returned text was parsed to verify structural compliance (four paragraphs, preference scores summing to 100\%). Had the response failed validation, the deterministic fallback generator would have been invoked automatically.
\end{enumerate}
The total inference time for steps 2-7 was approximately 3.2 seconds on CPU (Intel Xeon 8452Y) plus LLM API latency, consistent with the real-time decision-support requirements reported in the main text.

\begin{tcolorbox}[
    title=\small Box: LLM-generated explanation,
    fonttitle=\bfseries,
    colback=gray!5!white,
    colframe=gray!75!black,
    width=\columnwidth,
    left=4pt,right=4pt,top=4pt,bottom=4pt,
    boxsep=1.5pt
]
\label{appendix:full_LLMreasoning}
\small
The patient's Mean Arterial Pressure (MAP) has been fluctuating below the standard therapeutic target range of 65-85 mmHg throughout the observed period. Historical data indicate significant variability, with several transient dips likely reflective of underlying septic shock. Treatment markers indicate the use of vasopressors, aligning temporally with periods of lower MAP, although no direct causal inference is made. The trend does not demonstrate a clear monotonic recovery, highlighting ongoing instability in blood pressure regulation.

Among the top 5 influential vital signs, tidal volume (contribution 0.20) is stable at 650.00 mL, suggesting consistent ventilation settings. Respiratory rate (contribution 0.25) is currently 17/min and increasing, which may indicate evolving respiratory demands. Systemic vascular resistance (contribution 0.18) remains stable at 645.91 dyn·s/cm$^5$, suggesting peripheral vasodilation. Heart rate (contribution 0.22) is 68/min, lower than average, with an increasing trend, potentially indicating compensatory responses. Oxygen saturation (contribution 0.15) is 98\%, showing improvement and suggesting adequate oxygenation.

The predicted MAP trajectories under different treatment strategies show distinct patterns. The 'None' strategy projects a gradual increase in MAP, eventually reaching the target range but at a slower pace. 'Vaso' predicts a rapid increase, achieving and exceeding the target range quickly, likely due to the patient's previous responsiveness to vasopressors. 'Vent' also improves MAP, but with less magnitude compared to 'Vaso.' The 'Both' strategy predicts the highest MAP, potentially overshooting the target range. These predictions are influenced by the patient's stable tidal volume and increasing respiratory rate, reinforcing the benefit of vasopressors in this context.

\textbf{Treatment suggestion:} The model's preference distribution reflects these insights. 'Vaso' is preferred at 40\% due to its rapid attainment of MAP within the target range, consistent with historical treatment responses. 'None' is assigned 30\%, acknowledging its eventual sufficiency but slower response. 'Vent' is preferred at 20\% for moderate improvement without overshooting. 'Both' receives 10\%, as it may provide unnecessary elevation in MAP.
\end{tcolorbox}

\end{appendices}

%%===========================================================================================%%
%% If you are submitting to one of the Nature Portfolio journals, using the eJP submission   %%
%% system, please include the references within the manuscript file itself. You may do this  %%
%% by copying the reference list from your .bbl file, paste it into the main manuscript .tex %%
%% file, and delete the associated \verb+\bibliography+ commands.                            %%
%%===========================================================================================%%

\bibliography{sn-bibliography}% common bib file
%% if required, the content of .bbl file can be included here once bbl is generated
%%\input sn-article.bbl

\end{document}